\def\opre{}

\documentclass[mnsc,nonblindrev]{informs3}

\RequirePackage{tgtermes}
\RequirePackage{newtxtext}
\RequirePackage{newtxmath}
\RequirePackage{endnotes}
\usepackage{changepage}

\OneAndAHalfSpacedXI 

\usepackage{tikz}
\usepackage{algorithm}
\usepackage{algorithmic}

\usepackage{natbib}
 \bibpunct[, ]{(}{)}{,}{a}{}{,}%
 \def\bibfont{\small}%
\usepackage{amsmath,amsfonts,cancel}
\usepackage{thmtools}
\usepackage{adjustbox}
\usepackage[shortlabels]{enumitem}
\usepackage[table]{xcolor}
\definecolor{edits}{rgb}{0,0,0}
\usepackage{mathtools}
\usepackage{dsfont}
\usepackage{tcolorbox}
\usepackage{booktabs}
\usepackage{bbm}
\usepackage{xspace}
\usepackage{subcaption}
\usepackage{tensor}
\usepackage{bm}
\usepackage{tikz}
\usetikzlibrary{patterns}
\usepackage{natbib}
\usepackage[mathlines]{lineno}
\usepackage{changepage}
\usepackage{dsfont}

\usepackage{graphicx}
\usepackage{subcaption}

\usepackage{thm-restate}

\usepackage{placeins}
\usepackage[colorlinks=true,breaklinks=true,bookmarks=true,urlcolor=magenta,citecolor=magenta,linkcolor=magenta,bookmarksopen=false,draft=false]{hyperref}
\usepackage[capitalize]{cleveref}
\usepackage{tabularx}
\newenvironment{rproofof}[1]{%
    \ifdefined\opre 
        {\it Proof of #1.}%
    \else 
        \begin{rproof}[Proof of #1]%
    \fi
}{%
    \ifdefined\opre
        \hfill\Halmos
    \else
        \end{rproof}%
    \fi
}

\newenvironment{rproof}{%
    \ifdefined\opre 
        \proof{\it Proof.}%
    \else 
        \begin{rproof}%
    \fi
}{%
    \ifdefined\opre
        \hfill\Halmos\endproof  
    \else
        \end{rproof}
    \fi
}

\usepackage[suppress]{color-edits}
\addauthor{srs}{orange}
\addauthor{zz}{violet}

\newcommand{\R}{\mathbb{R}}
\renewcommand{\S}{\mathcal{S}}
\newcommand{\Regret}{\textsf{Regret}}
\newcommand{\A}{\mathcal{A}}
\newcommand{\B}{\mathcal{B}}

\newcommand{\E}{\mathcal{E}}
\newcommand{\Ex}{\mathbb{E}}

\newcommand{\ALG}{\textsf{O}-\textsf{O} \textsf{UCRL}-\textsf{VTR}\xspace}
\newcommand{\UCRL}{\textsf{UCRL}\xspace}
\newcommand{\UCBVI}{\textsf{UCBVI}\xspace}
\newcommand{\DPLSVI}{\textsf{DP-LSVI}\xspace}
\newcommand{\COMPLETE}{\textsf{COMPLETE}\xspace}

\newcommand{\F}{\mathcal{F}}
\renewcommand{\Pr}{\mathbb{P}}

\newcommand{\diag}{\operatorname{diag}}
\newcommand{\Dgap}{\Delta^{\off}_{\text{gap}}}
\newcommand{\off}{\mathrm{off}}
\newcommand{\on}{\mathrm{on}}
\newcommand{\all}{\mathrm{all}}

\newcommand{\C}{\mathcal{C}}

\newcommand{\Qpi}{Q^{\pi}}
\newcommand{\Vpi}{V^{\pi}}

\newcommand{\Qpik}{Q^{\pi_k}}
\newcommand{\Vpik}{V^{\pi_k}}
\newcommand{\Qoptimal}{Q^{\pi^*}}
\newcommand{\Voptimal}{V^{\pi^*}}
\newcommand{\Qhat}{\widehat{Q}}
\newcommand{\Vhat}{\widehat{V}}
\newcommand{\TildeV}{\widetilde V}
\newcommand{\thetaon}{\widehat{\theta}^{\on}}
\newcommand{\thetaall}{\widehat{\theta}^{\all}}

\usepackage{multirow}

\crefname{assumption}{assumption}{assumptions}
\crefname{algocf}{algorithm}{algorithm}

\usepackage{mathtools}

\DeclarePairedDelimiter{\abs}{\lvert}{\rvert}

\mathchardef\mhyphen="2D 
\ifdefined \argmin
\else \DeclareMathOperator*{\argmin}{argmin}
\fi
\ifdefined \argmax
\else \DeclareMathOperator*{\argmax}{argmax}
\fi
\DeclarePairedDelimiter{\norm}{\lVert}{\rVert}

\let\originalleft\left
\let\originalright\right
\renewcommand{\left}{\mathopen{}\mathclose\bgroup\originalleft}
\renewcommand{\right}{\aftergroup\egroup\originalright}

\renewcommand{\paragraph}[1]{\noindent\textbf{#1}.}

\renewcommand{\paragraph}[1]{\subsubsection*{#1}}

\EquationsNumberedThrough    

\TheoremsNumberedThrough     
\ECRepeatTheorems  %

\MANUSCRIPTNO{MNSC-0001-2024.00}

\begin{document}


\RUNAUTHOR{Zhang and Sinclair}

\RUNTITLE{Offline--Online Liner Mixture MDPs under Environment Shift}

\TITLE{Offline--Online Reinforcement Learning for Linear Mixture MDPs}

\ARTICLEAUTHORS{%

\AUTHOR{Zhongjun Zhang}
\AFF{Department of Industrial Engineering and Management Sciences,
Northwestern University, \EMAIL{zhongjun@u.northwestern.edu}}

\AUTHOR{Sean R. Sinclair}
\AFF{Department of Industrial Engineering and Management Sciences,
Northwestern University, \EMAIL{sean.sinclair@northwestern.edu}}

} 

\ABSTRACT{%
We study offline--online reinforcement learning in linear mixture Markov decision processes (MDPs) under environment shift.  In the offline phase, data are collected by an unknown behavior policy and may come from a mismatched environment, while in the online phase the learner interacts with the target environment.  We propose an algorithm that adaptively leverages offline data. When the offline data are informative, either due to sufficient coverage or small environment shift, the algorithm provably improves over purely online learning. When the offline data are uninformative, it safely ignores them and matches the online-only performance.  We establish regret upper bounds that explicitly characterize when offline data are beneficial, together with nearly matching lower bounds.  Numerical experiments further corroborate our theoretical findings.
}%





\maketitle


\section{Introduction}\label{sec_intro}

Recent progress in reinforcement learning (RL) has demonstrated strong performance in complex domains such as large-scale video games~\citep{silver2016mastering,mnih2015human,vinyals2017starcraft}.  However, these successes rely heavily on massive amounts of online interaction, often enabled by high-fidelity simulators. In many real-world applications, such interaction is limited or simulators are unavailable. For example, in inventory planning for new products, decisions must be made under uncertain demand without sufficient historical data to accurately simulate future outcomes, making it difficult to optimize policies through online experimentation alone. A natural approach is to leverage pre-collected data (i.e., offline data) from related environments. In retail, for instance, historical data from existing products can inform decisions for new products with limited data~\citep{amazon_scot_2022}. Such transfer has proven effective in practice, particularly in cold-start settings where direct data collection is costly or infeasible~\citep{aws_coldstart_2021}.

Despite its promise, leveraging offline data is challenging due to two fundamental issues. First, offline data may be generated under a similar but not identical environment, leading to bias under environment shift~\citep{mendonca2020meta}.  For instance, historical sales from existing products may be similar but not identical to the demand distribution of a new product.  Second, the offline data is collected under a behavior policy, which shapes the observed offline data that may not align with the target objective~\citep{jin2021pessimism}. 

While both purely online and offline RL are well understood for many canonical models, such as bandits and tabular MDPs, their combination (i.e., offline--online learning) remains less explored. In particular, existing work on offline--online learning often relies on simplifying assumptions on either environment shift or the offline behavior policy. Some works assume no environment shift~\citep{hao2023leveraging, wagenmaker2023leveraging}, while others impose restrictive assumptions on the offline behavior policy, such as fixed sampling in bandits~\citep{cheung2024leveraging} or uniform visitation in tabular MDPs~\citep{chen2022data}. These assumptions obscure the joint role of environment shift and the behavior policy, and do not extend naturally to large or continuous state–action spaces. Linear mixture MDPs provide a natural framework to study this gap, capturing rich function approximation while retaining tractable structure~\citep{chen2021near,zhou2021nearly}. 
Notably, the inventory example above can be modeled as a special case of the linear mixture MDP studied in this work~\citep{wan2024exploiting}.

Addressing these challenges is nontrivial. Naively combining offline and online data without accounting for environment shift can introduce bias and lead to linear regret~\citep{song2022hybrid}.  This motivates the need for algorithms that {\em safely} incorporate offline data --- achieving improved regret when the offline data is informative, while matching purely online guarantees otherwise.  At the same time, the presence of environment shift complicates the coverage conditions and pessimism-based estimation techniques commonly used in offline RL~\citep{jin2021pessimism,yin2022near,nguyen2023instance}, suggesting that these principles must be reconsidered in the offline--online setting.

Motivated by these challenges, we investigate the following questions in the offline--online linear mixture MDP setting:
\smallskip

\begin{center}
   {\em How can offline data be safely leveraged in online learning under environment shift without degrading regret guarantees? How do environment shift and the offline behavior policy jointly govern the informativeness of offline data?}
\end{center}

\subsection{Our Contributions}

\paragraph{Offline--Online Learning Algorithm to Safely Leverage Offline Data.}
Our first contribution is an algorithm,~\ALG (\Cref{alg_main}), for offline--online linear mixture MDPs under environment shift and uncontrollable offline behavior policy. The backbone of~\ALG follows the \textsf{UCRL} with value-targeted regression (\textsf{UCRL}-\textsf{VTR}\xspace) algorithm in~\citet{jia2020model}. The key difference is that, at each stage of the online phase, we maintain two estimators of the true parameter $\theta^*$ that governs the transition dynamics of the online target environment: an {\em online-only} estimator based solely on online interactions, and an {\em offline--online} estimator that combines both offline data and online interactions. 
Our algorithm adaptively relies on the more reliable estimator, thereby leveraging offline data only when it is informative. A detailed description of~\ALG is provided in~\Cref{sec_algo}.

We establish a regret upper bound for~\ALG in~\Cref{thm1}, which depends on a quantity $\tau$ that measures the quality of the offline behavior policy. $\tau$ is defined by the minimum eigenvalue of the offline design matrix (see~\Cref{assumption_eigen}), and is closely related to uniform coverage conditions in offline RL literature~\citep{yin2022near}.
In line with prior approaches in offline RL~\citep{yin2022near, nguyen2023instance}, the value of $\tau$ is not required by the implementation of~\ALG.
The regret bound in~\Cref{thm1} exhibits a ``min-of-two'' structure: it is always safeguarded by the regret of the purely online algorithm~\citep{jia2020model}, and can be strictly smaller when the offline data is informative, due to either small environment shift or large coverage $\tau$.

\paragraph{Characterization of Offline Data Informativeness and Coverage $\tau$.}
In contrast to the nearly minimax regret in the purely online setting, which scales as $\sqrt{K}$ with $K$ denoting the online interaction budget~\citep{jia2020model}, we provide, in~\Cref{sec_when_offlinedata_informative}, sufficient conditions under which offline data is informative, that is, yields $o(\sqrt{K})$ regret (see~\Cref{def_informative}) in the online learning process.

We first consider the ``sufficient coverage'' case $\tau = \Theta(1)$. In this regime, offline data is informative if: (i) the offline sample size $M^\off$ is sufficiently large relative to the online interaction budget $K$; and (ii) the environment shift $\Delta$ is below a threshold determined by $K$, where $\Delta$ is measured in the $\|\cdot\|_2$ norm (see~\Cref{assump_theta_gap}). 
This aligns with the intuition that more offline data and smaller environment shift, when properly leveraged as in~\Cref{alg_main}, lead to greater offline data informativeness.
We further show in~\Cref{sec_comparison} that $\tau = \Theta(1)$ holds under the conventional uniform coverage condition in the offline RL literature~\citep{nguyen2023instance}, as formalized in~\Cref{assumption_data_coverage_main}, under appropriate design of offline value function estimates (see~\Cref{cor_uniformdata_coverage} and~\Cref{cor_uniformdata_coverage2}). Notably, the offline value function estimates we propose in~\Cref{sec_comparison} differ from conventional pessimism-based principles in the offline RL literature, as the optimal policies in the offline and online environments can differ substantially due to environment shift.

We next consider the complementary case $\tau = o(1)$. In this case, offline data is informative if: (i) the coverage $\tau$ exceeds a threshold determined by $K$ and $M^\off$; and (ii) the ratio $\Delta/\tau$ is below a threshold determined by $K$. Condition (ii) implies a {\em linear} relationship between $\Delta$ and $\tau$, so that larger $\tau$ permits proportionally larger environment shift while preserving offline data informativeness. 
This regime captures a less explored setting in the offline RL literature, where the offline behavior policy is {\em adaptive}~\citep{nguyen2023instance}. We provide a bandit example in~\Cref{instance_adaptive_policy} showing that offline data can remain informative when the behavior policy follows \textsf{UCB1}~\citep{auer2002finite}. In this case, uniform coverage (\Cref{assumption_data_coverage_main}) is violated, yet our algorithm achieves $o(\sqrt{K})$ regret under suitable regimes of $K$, $M^\off$, and $\Delta$.

\paragraph{Nearly Matching Regret Lower Bound.}
We also provide a nearly matching lower bound in~\Cref{sec_lower_bound}. 
The lower bound considers the case when $\tau = \Theta(1)$ and exhibits a similar ``min-of-two'' structure, and nearly match the upper bound in~\Cref{thm1} in terms of $d, M^\off, K$ and $\Delta$, in both regimes where offline data is informative and where it is not.
A discussion of the regret lower bound for the complementary case $\tau = o(1)$ is also provided in~\Cref{sec_lower_bound}.

\paragraph{Empirical Simulations.}
We conduct empirical simulations in a synthetic setting, comparing~\ALG with several baseline algorithms for linear mixture MDPs in~\Cref{sec_simulations}. The simulation results corroborate our analysis: (i) larger values of $\tau$ and smaller environment shift $\Delta$ lead to greater informativeness of the offline data; (ii) \ALG is robust, performing no worse than the purely online baseline~\UCRL when $\Delta$ is large, and significantly better when $\Delta$ is small; (iii) when $\Delta$ scales linearly with $\tau$, offline data remains informative as $\Delta$ increases, whereas the benefit of offline data rapidly diminishes when $\Delta$ grows superlinearly than $\tau$. 
We also compare~\ALG with more baselines designed for tabular MDPs; additional details are provided in~\Cref{appendix_experiment}.

\paragraph{Paper Organization.}
We review the related literature in the remainder of this section. In~\Cref{sec:preliminary}, we introduce the offline--online linear mixture MDP model, followed by a detailed description of our algorithm~\ALG in~\Cref{sec_algo}. 
Our main algorithmic results are presented in~\Cref{sec_main_results}. \Cref{sec_upper_bound} establishes the regret upper bound, and in~\Cref{sec_when_offlinedata_informative}, we analyze how the informativeness of offline data depends on the environment shift $\Delta$ and coverage $\tau$. In~\Cref{sec_comparison}, we further study how $\tau$ is shaped by the feature mapping of linear mixture MDP model, the offline behavior policy, and the design of offline value function estimates.
We complement these results with nearly matching lower bounds in~\Cref{sec_lower_bound}. Finally, we demonstrate the strong empirical performance of our algorithm in~\Cref{sec_simulations} and conclude in~\Cref{sec_conclusion}.

\subsection{Related Literature}
Our work is highly related to three threads of research in the literature: online RL, offline RL, and offline--online RL under either environment or distribution shift. For additional background, we refer readers to \citet{sutton1998reinforcement, bertsekas2019reinforcement} for online RL, and to Chapter~4 of \citet{szepesvari2022algorithms} as well as \citet{jiang2025offline} for offline RL.

\paragraph{Online RL.} A substantial body of work in reinforcement learning focuses on the online setting. The classical formulation is tabular RL, where both the state and action spaces are finite. Two main algorithmic paradigms dominate this literature: model-based methods~\citep{brafman2002r, azar2017minimax, auer2008near, dann2017unifying, simchowitz2019non} and $Q$-learning-based approaches~\citep{jin2018q}. Beyond these frameworks, alternative perspectives have also been explored, including posterior sampling~\citep{agrawal2017optimistic} and policy-level optimization under structural assumptions on the policy class, such as convexity or counterfactual estimability~\citep{agrawal2019learning, zhang2025reinforcement}. 
Beyond the tabular regime, increasing attention has been devoted to online RL with function approximation in large or continuous state and action spaces. Under linear realizability assumptions, much of the recent progress concerns linear MDPs~\citep{yang2019sample, jin2020provably, zanette2020frequentist}, linear mixture MDPs~\citep{modi2020sample, jia2020model, zhou2021nearly, ayoub2020model}, and analyses based on controlling Bellman errors~\citep{jiang2017contextual, zanette2020learning}. There is also a growing literature addressing function approximation beyond linear models, including bilinear and more general representations~\citep{du2021bilinear, foster2021statistical, zhu2023provably}. 

This work builds on the framework of online RL for linear mixture MDPs, with an explicit incorporation of an offline data phase, in which both environment shift and offline data coverage play essential roles. In the absence of offline data, our regret bound reduces to $\tilde{O}(dH^2\sqrt{K})$, matching the guarantees in \citet{jia2020model, ayoub2020model}. A tighter bound of $\tilde{O}(dH^{3/2}\sqrt{K})$ is obtained by \citet{zhou2021nearly} via a Bernstein-style concentration argument, at the cost of a substantially more involved analysis.  While similar techniques could potentially be used to remove the extra $\sqrt{H}$ factor in our bound, our focus here is on isolating the roles of environment shift and coverage in the offline--online setting. We leave such refinements to future work.

\paragraph{Offline RL.} Unlike online RL, in offline RL the learner cannot interact with the environment and must rely entirely on a fixed, pre-collected dataset. As a result, coverage plays a central role in determining the statistical feasibility of offline learning.  Early work focused on \emph{uniform} coverage, requiring the logged data to adequately explore the entire state--action space, including \citet{munos2008finite} under finite pseudo-dimension assumptions and \citet{chen2019information} in the tabular setting. More recent studies shift to \emph{partial} coverage, examining whether the behavior policy sufficiently covers the optimal policy’s state--action distribution. Representative works include \citet{jin2021pessimism} and \citet{xie2021bellman}, which develop general function approximation frameworks with linear instantiations; \citet{nguyen2023instance}, which study linear MDPs and linear mixture MDPs; and \citet{jiang2025offline}. In structured models such as linear MDPs and linear mixture MDPs, coverage is often characterized in the feature space via eigenvalue conditions on the design matrix, as in \citet{yin2022near} for linear MDPs and \citet{uehara2021pessimistic}, which provides a general framework instantiated in tabular, linear mixture, and low-rank MDPs.

Among the works discussed above, those most closely related to our setting, namely linear mixture MDPs with feature-space coverage conditions, are \citet{yin2022near, uehara2021pessimistic} and~\citet{nguyen2023instance}. \citet{yin2022near} study linear MDPs under a uniform coverage assumption in the feature space, formalized via a uniformly positive minimum eigenvalue of the design matrix. This condition is closely related to our assumption (see~\Cref{assumption_eigen}), though their setting focuses on linear MDPs rather than the linear mixture MDPs considered here. \citet{uehara2021pessimistic} and \citet{nguyen2023instance} consider offline RL for linear mixture MDPs under partial coverage, with the former focusing on data-space coverage and the latter on feature-space coverage. A detailed comparison between~\Cref{assumption_eigen} and the coverage condition in~\citet{nguyen2023instance} is given in~\Cref{discussion_offlineRL} and~\Cref{discussion_coverage_2}. All of these works achieve an suboptimality gap of order $\tilde{O}(1/\sqrt{M^{\off}})$, where $M^{\off}$ denotes the offline sample size. This rate is recovered as a special case of our results when $K=1$ and the environment shift is negligible (see~\Cref{sec_comparison}).

\paragraph{Offline--Online RL.}
Building on the extensive literature on online and offline RL, offline--online RL has recently emerged as an active research direction. The central question is whether and how logged offline data can be leveraged to improve subsequent online learning. Beyond data coverage, which is critical in the offline phase, environment shift also becomes a key factor. When there is no environment shift and the offline and online environments coincide, it is relatively well understood in both RL and bandits that offline data can be leveraged to reduce both the bias and the uncertainty in value and model estimation~\citep{hao2023leveraging, wagenmaker2023leveraging, shivaswamy2012multi, banerjee2022artificial, agrawal2023optimal, sentenac2025balancing, song2022hybrid}.

When environment shift is present, however, the offline and online environments may differ, and this is precisely the regime studied in this work. In the bandit setting,~\citet{cheung2024leveraging, yang2025best} analyze offline--online learning under environment shift, but assume fixed numbers of offline pulls for each arm, so the offline behavior policy and coverage do not play an explicit role. In tabular RL,~\citet{chen2022data} study the offline--online setting, but their algorithm can perform worse than a purely online method when the environment shift $\Delta$ is large, and neither the offline behavior policy nor coverage is explicitly characterized. In contrast, our work extends to linear mixture MDPs, which subsume both bandits and tabular MDPs (see~\Cref{dis_comparison_new}), and proposes an algorithm (\Cref{alg_main}) whose performance is always safeguarded by a purely online baseline. Moreover, we provide a detailed characterization of how environment shift and coverage interact to determine when offline data are informative (see~\Cref{sec_when_offlinedata_informative}).

Related lines of work also include settings where the offline behavior policy can be queried during the online phase rather than relying on logged data~\citep{ross2012agnostic, xie2021policy}, as well as non-stationary RL and change-point detection~\citep{xie2025online, auer2019adaptively}. There is also a growing body of empirical studies on offline--online RL~\citep{ball2023efficient, nakamoto2023cal, zheng2023adaptive}. Specifically, \citet{zheng2023adaptive} study an offline--online RL setting and propose an algorithm that combines pessimism-based estimation in the offline phase with optimistic exploration in the online phase. In contrast, our analysis in~\Cref{sec_main_results}, together with the simulations in~\Cref{sec_simulations}, suggests that in our setting, offline data can be utilized more effectively by prioritizing statistical efficiency in the offline phase, rather than enforcing pessimism.

\section{Preliminary}\label{sec:preliminary}

\paragraph{Technical Notation.}
Let $[K] = \{1,\ldots,K\}$. For a vector $x \in \mathbb{R}^d$, $\|x\|_p$ denotes its $p$-norm. For a positive definite matrix $A \in \mathbb{R}^{d \times d}$, define the weighted Euclidean norm $\|x\|_A = \sqrt{x^\top A x}$. The smallest and largest eigenvalues of $A$ are denoted by $\lambda_{\min}(A)$ and $\lambda_{\max}(A)$, respectively.
The space $\mathbb{R}^2$ is equipped with the lexicographic order: $(a_1,b_1) \le (a_2,b_2)$ if $a_1 < a_2$, or if $a_1 = a_2$ and $b_1 \le b_2$; the strict order $(a_1,b_1) < (a_2,b_2)$ is defined analogously. The vector $\mathbf 1_d \in \mathbb{R}^d$ denotes the all-ones vector. $B(p)$ denotes the Bernoulli distribution with parameter $p \in [0,1]$.

\paragraph{Offline--Online MDP.}
We study a reinforcement learning problem consisting of two phases: an offline phase and an online phase. Each phase is modeled as an episodic Markov decision process (MDP) with a finite horizon. Specifically, the offline MDP is given by the tuple $(\mathcal S, \mathcal A, P^{\off}, r^{\off}, H, s^{\off}_1)$, where $\mathcal S$ is the state space, $\mathcal A$ is the action space, $P^{\off}$ is the transition kernel, $r^{\off}:\mathcal S \times \mathcal A \to [0,1]$ is a deterministic reward function, $H$ is the episode length, and $s^{\off}_1$ is the initial state. The online MDP is defined analogously by $(\mathcal S, \mathcal A, P, r, H, s_1)$.
For simplicity, we assume that the state space $\mathcal S$, action space $\mathcal A$, and horizon $H$ are identical in the offline and online MDPs. For notational simplicity in what follows, we refer to the pair $\{(\mathcal S, \mathcal A, P^{\off}, r^{\off}, H, s^{\off}_1), (\mathcal S, \mathcal A, P, r, H, s_1)\}$ as the corresponding \emph{offline--online MDP}.

Throughout the paper, we restrict attention to homogeneous linear mixture MDPs~\citep{jia2020model, zhou2021nearly, nguyen2023instance}, defined as follows.
\begin{definition}\label{def_mdp_1}
There exist unknown parameter vectors $\theta^{\off,*}, \theta^* \in \mathbb R^d$ such that, for all $s,s' \in \mathcal S$ and $a \in \mathcal A$,
\begin{align}
\label{def_MDP}
P^{\off}(s' \mid s,a) = \langle \theta^{\off,*}, \phi(s' \mid s,a) \rangle, 
\qquad
P(s' \mid s,a) = \langle \theta^*, \phi(s' \mid s,a) \rangle,
\end{align}
where $\phi:\mathcal S \times \mathcal A \times \mathcal S \to \mathbb R^d$ is a known feature mapping.
\end{definition}
We assume that the feature mapping satisfies $\sum_{s'} \lvert \phi_j(s' \mid s,a) \rvert \le 1,$ for all $(s,a)$ and $j\in[d]$, where $\phi(s'\mid s,a) = [\phi_1(s'\mid s,a),\ldots,\phi_d(s'\mid s,a)]^\top$.  Moreover, we assume that $\max\{\|\theta^{\off,*}\|_2,\|\theta^*\|_2\}\le B,$ for a known constant $B>0$~\citep{jia2020model}. For future use, we define $\phi(\cdot \mid s, a) = [\phi(s' \mid s, a)]_{s' \in \S} \in \mathbb{R}^{d \times |\S|}$ and, for each $j \in [d]$, $\phi_j(\cdot \mid s, a) = [\phi_j(s' \mid s, a)]_{s' \in \S} \in \mathbb{R}^{|\S|}$.

For clarity of exposition,\footnote{The extension to settings with distinct offline and online feature mappings, as well as unknown reward functions, follows the same analysis as in~\Cref{alg_main} and leads to nearly identical results to those in~\Cref{sec_main_results}; see~\citet{jia2020model}.} we additionally assume that the reward functions $r^{\off}$ and $r$ are deterministic and known~\citep{jia2020model,zhou2021nearly}, and that the feature mapping $\phi$ is shared between the offline and online MDPs. We emphasize that we do not require each $\phi_j(\cdot\mid s,a)$ to define a valid probability transition model for all $(s,a)$; in particular, $\phi_j(\cdot\mid s,a)$ need not be nonnegative nor sum to one.

\begin{remark}\label{remark_tabularMDP_equivalence}
Note that any tabular MDP can be expressed as a linear mixture MDP with dimension $d = |\mathcal S|^2 |\mathcal A|$. 
To see this, we let $\phi(s' \mid s,a) = e_{(s',s,a)}$ be a one-hot encoding of the tuple $(s',s,a)$.  Then if $\theta_{(s',s,a)} = P(s' \mid s,a)$ we trivially have that $P(s' \mid s,a) = \theta^\top \phi(s' \mid s,a)$.
This construction shows that the linear mixture MDP model strictly generalizes the tabular setting. We later discuss how the results in~\Cref{sec_main_results} go beyond existing guarantees for tabular MDPs~\citep{chen2022data}.
\end{remark}

Finally, we assume that the discrepancy between the offline and online transition parameters is {\em known} and bounded.
\begin{assumption}\label{assump_theta_gap}
$\|\theta^* - \theta^{\off,*}\|_2 \le \Delta$, where $\Delta\ge0$ is a known constant.
\end{assumption}

This assumption is standard in offline--online learning settings~\citep{crammer2008learning,chen2022data,besbes2022beyond,cheung2024leveraging}. A detailed discussion of \Cref{assump_theta_gap}, including comparisons with environment shift assumptions in the existing literature, is deferred to~\Cref{dis_comparison_new}.

\paragraph{Offline Data and Behavior Policy.}
Learning begins with an offline phase, during which the learner does not interact with the environment. 
In this phase, a total of $M^{\off}$ trajectories are collected from the offline environment. 
The trajectories are generated sequentially under a sequence of behavior policies 
$\{\pi^{\off}_m\}_{m=1}^{M^\off}$.
For each $m\in[M^\off]$, the behavior policy $\pi^{\off}_m=\{\pi^{\off}_{m,h}\}_{h=1}^H$ 
consists of stage-wise mappings $\pi^{\off}_{m,h}:\S \to \A$. 
Under the behavior policy $\pi^{\off}_m$, the environment generates a length-$H$ trajectory $\mathcal H^{\off}_m$ according to the offline transition dynamics $P^{\off}$, starting from the initial state $s^{\off}_1$:
\[
\mathcal H^{\mathrm{off}}_m =
\left\{
(s^{\mathrm{off}}_{m,1},a^{\mathrm{off}}_{m,1}, r^{\mathrm{off}}(s^{\mathrm{off}}_{m,1},a^{\mathrm{off}}_{m,1})),
\ldots, 
(s^{\mathrm{off}}_{m,H},a^{\mathrm{off}}_{m,H},
r^{\mathrm{off}}(s^{\mathrm{off}}_{m,H},a^{\mathrm{off}}_{m,H}))
\right\},
\]
where at stage $h$, upon observing state $s^{\off}_{m,h}$, the action is $a^{\off}_{m,h}=\pi^{\off}_{m,h}(s^{\off}_{m,h})$, the received reward is $r^{\off}(s^{\off}_{m,h},a^{\off}_{m,h})$, and the next state satisfies $s^{\off}_{m,h+1}\sim P^{\off}(\cdot\mid s^{\off}_{m,h},a^{\off}_{m,h})$. The resulting offline dataset is
\[
\mathcal D_{\off} = \bigl\{ \mathcal H^{\off}_m \bigr\}_{m=1}^{M^{\off}}.
\]

The index $m$ in behavior policies $\{\pi^{\off}_m\}_{m=1}^{M^\off}$ indicates that the sequence of behavior policies may be adaptive during the offline phase~\citep{nguyen2023instance}; that is, the policy $\pi^{\off}_m$ may vary across $m$. 
No additional assumptions are imposed on the behavior policies $\{\pi^{\off}_m\}_{m = 1}^{M^\off}$, such as being Markovian or stationary~\citep{laroche2023occupancy}. Its impact on learning performance is reflected implicitly through the distribution and geometric properties of the offline dataset $\mathcal D^{\off}$ induced by the offline MDP, and is examined further in~\Cref{sec_upper_bound}.

\paragraph{Online Interaction and Regret.}
After observing the offline dataset $\mathcal D_{\off}$, the learner interacts with the online MDP for $K$ episodes, each of horizon $H$, starting from the designated initial state $s_1$. 
At the beginning of episode $k$, the learner selects a deterministic policy $\pi_k$ based on $\mathcal D_{\off}$ and all past online observations, and executes $\pi_k$ throughout the episode. 
Similar to the offline behavior policies, the policy $\pi_k=\{\pi_{k,h}\}_{h=1}^H$ consists of stage-wise mappings $\pi_{k,h}:\mathcal S\to\mathcal A$.
At stage $h$, upon observing state $s^{\on}_{k,h}$, the learner takes action $a^{\on}_{k,h}=\pi_{k,h}(s^{\on}_{k,h})$, receives reward $r(s^{\on}_{k,h},a^{\on}_{k,h})$, and transitions to $s^{\on}_{k,h+1}\sim P(\cdot\mid s^{\on}_{k,h},a^{\on}_{k,h})$.
At the end of episode $k$, the realized trajectory is
\[
\left\{ (s^{\on}_{k,1},a^{\on}_{k,1},r(s^{\on}_{k,1},a^{\on}_{k,1})),
\ldots, (s^{\on}_{k,H},a^{\on}_{k,H},r(s^{\on}_{k,H},a^{\on}_{k,H})) \right\}.
\]

For any policy $\pi=\{\pi_h\}_{h=1}^H$, the $Q$ functions and value functions in the online phase are defined as
\begin{align}
\Qpi_h(s,a) = \mathbb E_\pi \left[ \sum_{i=h}^H r(s_i,a_i) \Bigm| s_h=s, a_h=a \right], 
\qquad
\Vpi_h(s) = \Qpi_h\left(s,\pi_h(s)\right), \label{eq_vdefinition}
\end{align}
with the convention $\Vpi_{H+1}(s)=0$ for all $s\in\mathcal S$.
Let $\pi^*$ denote an optimal policy for the online MDP. The performance metric is the (expected) regret over $K$ episodes, defined as
\begin{equation}
\Regret(K) = \sum_{k=1}^K \mathbb E\left[ \Voptimal_1(s_1) - \Vpik_1(s_1) \right], 
\label{eq:regret_definition}
\end{equation}
where the expectation is taken over the randomness of both the offline data generation and the online interaction, since the offline dataset $\mathcal D_{\off}$ influences the deployed policies $\{\pi_k\}_{k=1}^K$ as well.

We emphasize that the offline environment may differ from the online environment. As a result, offline trajectories are, in general, biased with respect to the online dynamics. A central goal of this work is to understand when and how such biased offline data can still be exploited for online learning, and to quantify the respective effects of \emph{environment shift} (i.e., $P^{\off} \neq P$ so $\Delta > 0$) and the \emph{coverage} induced by the offline behavior policy~\citep{jiang2025offline} on the resulting performance guarantees.

\section{Algorithm Design}
\label{sec_algo}

\begin{algorithm}[h!]
\caption{\ALG}
\label{alg_main}
\begin{algorithmic}[1]
\REQUIRE Online MDP, dimension $d$, number of online episodes $K$, offline dataset $\mathcal{D}_{\off}$, auxiliary value functions $\{\TildeV_{m,h}\}$, radii $\{\gamma_k\}_{k\le K}$ and $\{\beta_k\}_{k\le K}$, ridge regression coefficient $\lambda$.
\STATE Initialize online-only design matrix and offline--online design matrix:
$
M^{\on}_1 \leftarrow \lambda I, w^{\on}_1 \leftarrow 0\in\R^d, M^{\all}_1 \leftarrow \lambda I, w^{\all}_1 \leftarrow 0\in\R^d.
$ 
\STATE \textit{\# Offline phase}
\FOR{$m=1 : M_{\off}$} 
    \FOR{$h=1 : H$}
        \STATE Form the regression pair 
        $
        x^{\off}_{m,h} \leftarrow \phi(\cdot \mid  s^{\off}_{m,h},a^{\off}_{m,h})\TildeV_{m,h+1}\in\R^d,
        y^{\off}_{m,h} \leftarrow \TildeV_{m,h+1}\bigl(s^{\off}_{m,h+1}\bigr)
        $. 
        \STATE Update statistics
        $
        M^{\all}_1 \leftarrow M^{\all}_1 + x^{\off}_{m,h}(x^{\off}_{m,h})^\top,
        w^{\all}_1 \leftarrow w^{\all}_1 + x^{\off}_{m,h} y^{\off}_{m,h}.
        $
    \ENDFOR
\ENDFOR
\STATE \textit{\# Online phase}
\FOR{$k=1:K$}
    \STATE Compute estimators 
    $
    \thetaall_k \leftarrow (M^{\all}_k)^{-1} w^{\all}_k,
    \thetaon_k \leftarrow (M^{\on}_k)^{-1} w^{\on}_k.
    $
    \STATE Define confidence sets
    $
    \B_k \leftarrow \left\{\theta\in\R^d:\|\theta-\thetaall_k\|_{M^{\all}_k}\le \gamma_k\right\},
    \C_k \leftarrow \left\{\theta\in\R^d:\|\theta-\thetaon_k\|_{M^{\on}_k}\le \beta_k\right\}.
    $
    \STATE Set $\Qhat_{k, H+1}(\cdot,\cdot)\leftarrow 0$, and for $h=H,\dots,1$ define \hfill // \textit{Compute optimistic estimates $\Qhat$ and $\Vhat$}
    \[
    \Vhat_{k,h}(s)\leftarrow \max_{a\in\A} \Qhat_{k,h}(s,a),
    \qquad
    \Qhat_{k,h}(s,a)\leftarrow r(s,a)+\max_{\theta\in\B_k\cap\C_k}\sum_{j=1}^d \theta_j \phi_j(\cdot\mid s,a)^\top \Vhat_{k,h+1}.
    \]
    \STATE Set $s^{\on}_{k,1} =  s_1$.
    \FOR{$h=1:H$}
        \STATE Choose $a^{\on}_{k,h}\in\arg\max_{a\in\A} \Qhat_{k,h}(s^{\on}_{k,h},a)$ and observe $s^{\on}_{k,h+1}\sim P(\cdot\mid s^{\on}_{k,h},a^{\on}_{k,h})$.
        \STATE Form the online regression pair
        $
        x^{\on}_{k,h} \leftarrow \phi(\cdot \mid s^{\on}_{k,h},a^{\on}_{k,h})\Vhat_{k,h+1},
        y^{\on}_{k,h} \leftarrow \Vhat_{k,h+1}\bigl(s^{\on}_{k,h+1}\bigr).
        $
        \STATE Update online-only statistics
        $
        M^{\on}_{k} \leftarrow M^{\on}_k + x^{\on}_{k,h}(x^{\on}_{k,h})^\top,
        w^{\on}_{k} \leftarrow w^{\on}_k + x^{\on}_{k,h} y^{\on}_{k,h}.
        $
        \STATE Update offline--online statistics
        $
        M^{\all}_{k} \leftarrow M^{\all}_k + x^{\on}_{k,h}(x^{\on}_{k,h})^\top,
        w^{\all}_{k} \leftarrow w^{\all}_k + x^{\on}_{k,h} y^{\on}_{k,h}.
        $
        \ENDFOR
        \STATE $
        M^{\on}_{k+1} \leftarrow M^{\on}_{k},
        M^{\all}_{k+1} \leftarrow M^{\all}_{k},
        w^{\on}_{k+1} \leftarrow w^{\on}_{k},
        w^{\all}_{k+1} \leftarrow w^{\all}_{k}.
        $
\ENDFOR
\end{algorithmic}
\end{algorithm}

This section presents the details of~\Cref{alg_main}, which addresses offline--online linear mixture MDPs where the offline environment may differ from the online one. 
At a high level, the algorithm consists of two components: estimating the transition parameter $\theta^*$, and using this estimate to compute an optimistic approximation of the optimal value function $\Voptimal_1$ and the corresponding policy $\pi^*$. A central feature of the offline--online setting is that $\theta^*$ can be estimated in two distinct ways. The first is an \emph{online-only} estimator that uses data collected from online interaction alone, while the second is an \emph{offline--online} estimator that aggregates both offline and online data.
By appropriately leveraging these two estimators, the algorithm effectively uses the more reliable estimate at each stage, enabling it to benefit from informative offline data while remaining robust when such data are uninformative. 

\paragraph{Estimation of $\theta^*$.}
We estimate $\theta^*$ via \emph{value-targeted regression}~\citep{jia2020model}.  We begin with the \emph{online-only} estimator (line~11 of~\Cref{alg_main}). At episode $k$ and stage $h$, define
\begin{align}
x^{\on}_{k,h} = \phi(\cdot \mid s^{\on}_{k,h}, a^{\on}_{k,h})\Vhat_{k,h+1},
\qquad
y^{\on}_{k,h} = \Vhat_{k,h+1}(s^{\on}_{k,h+1}), \label{eq_defof_xon_yon}
\end{align}
where $\Vhat_{k,h+1} \in \mathbb{R}^{|\S|}$ is an estimate of the optimal value function $\Voptimal_{h+1}$ (specified later in~\cref{eq_vhat}).  Under the linear mixture model, we have
\begin{equation}
\theta^{*\top} x^{\on}_{k,h} = \mathbb E\left[\Vhat_{k,h+1}(s^{\on}_{k,h+1}) \mid s^{\on}_{k,h}, a^{\on}_{k,h}\right],
\label{eq_regression}
\end{equation}
and $y^{\on}_{k,h}$ is a corresponding realization.
This yields a standard linear regression problem, allowing $\theta^*$ to be estimated via ridge regression:
\begin{align}
\thetaon_k = (M^{\on}_k)^{-1} w^{\on}_k, \label{eq_regression_online}
\end{align}
where
\begin{align*}
M^{\on}_k = \lambda I + \sum_{k'=1}^{k-1} \sum_{h=1}^{H} x^{\on}_{k',h}\bigl(x^{\on}_{k',h}\bigr)^{\top}, 
\qquad
w^{\on}_k = \sum_{k'=1}^{k-1} \sum_{h=1}^{H} x^{\on}_{k',h} y^{\on}_{k',h}.
\end{align*}
Here $\lambda>0$ is a ridge regularization parameter to be specified later.

Importantly, the regression in \cref{eq_regression} remains valid for any choice of $\Vhat_{k,h+1}$ that is non-anticipatory, i.e. depends on the data observed prior to stage $h$ in episode $k$ (and all earlier episodes), but is held fixed when sampling $s^\on_{k,h+1}$ given $s^\on_{k,h},a^\on_{k,h}$.  However, the regret guarantee hinges on a specific data-adaptive optimistic construction of $\Vhat_{k,h+1}$ (we show later in~\cref{eq_vhat}), as it controls how parameter estimation error propagates through planning to value function error, and ultimately determines the regret bound.

In addition to the online-only estimator, we also construct an \emph{offline--online} estimator that additionally incorporates offline data. This is enabled by \Cref{assump_theta_gap}, which bounds the discrepancy between the offline and online transition parameters. Intuitively, when $\Delta$ is small, accurate estimation of $\theta^{\off,*}$ can be leveraged to improve estimation of $\theta^*$.
Specifically, for offline samples we define
\begin{align}
x^{\off}_{m,h} = \phi(\cdot \mid s^{\off}_{m,h}, a^{\off}_{m,h})\TildeV_{m,h+1}, 
\qquad
y^{\off}_{m,h} = \TildeV_{m,h+1}(s^{\off}_{m,h+1}), 
\label{eq_def_xoff_yoff}
\end{align}
where $\left\{\TildeV_{m,h} \in \mathbb{R}^{|\S|}\right\}$ are auxiliary value functions chosen by the learner.
The offline--online estimator is then
\begin{align}
\thetaall_k = (M^{\all}_k)^{-1} w^{\all}_k,
\label{eq_regression_offline_online}
\end{align}
where we have
\begin{align*}
M^{\all}_k = M^{\on}_k + \sum_{m=1}^{M^{\off}} \sum_{h=1}^{H} x^{\off}_{m,h}\bigl(x^{\off}_{m,h}\bigr)^{\top}, 
\qquad
w^{\all}_k = w^{\on}_k + \sum_{m=1}^{M^{\off}} \sum_{h=1}^{H} x^{\off}_{m,h} y^{\off}_{m,h}.
\end{align*}

Unlike the online phase, the offline phase does not require $\{\TildeV_{m,h}\}$ to be chosen adaptively for regret minimization. This provides additional flexibility in designing $\{\TildeV_{m,h}\}$. Both the choice of $\{\TildeV_{m,h}\}$ and the offline behavior policy affect the geometry of the offline design matrix and, consequently, the convergence rate of $\|\thetaall_k-\theta^*\|$. We defer a detailed discussion of these effects, as well as principled choices of $\{\TildeV_{m,h}\}$, to~\Cref{sec_main_results}.  

\paragraph{Optimistic Planning under Offline--Online Confidence Sets.}
In each episode $k$, given the two estimators $\thetaon_k$ and $\thetaall_k$, the algorithm follows a standard optimistic planning approach based on confidence set elimination. Specifically, combining the ridge regression estimators with Assumption~\ref{assump_theta_gap}, we construct two confidence sets for $\theta^*$:
\begin{align}
\C_k = \left\{ \theta \in \mathbb R^d : \|\theta-\thetaon_k\|_{M^{\on}_k} \leq \beta_k \right\},
\qquad
\B_k = \left\{ \theta \in \mathbb R^d : \|\theta-\thetaall_k\|_{M^{\all}_k} \leq \gamma_k \right\},
\label{eq_def_ellipsoid}
\end{align}
where $\beta_k$ and $\gamma_k$ are appropriately chosen confidence radii. The value function estimates $\Vhat_{k,h}$ are then computed via optimistic planning over the intersection $\C_k \cap \B_k$.
For all $h \leq H$, we define\footnote{{Both $\B_k$ and $\C_k$ are ellipsoids for each $k$. Since the objective is linear in $\theta$, the resulting optimization is a quadratically constrained quadratic program (QCQP)~\citep{bao2011semidefinite}, which is tractable.}}
\begin{align}
\Qhat_{k,h}(s,a) = r(s,a) + \max_{\theta\in\C_k\cap\B_k} \sum_{j=1}^d \theta_j \phi_j(\cdot\mid s,a)^\top \Vhat_{k,h+1},
\qquad
\Vhat_{k,h}(s) = \max_{a\in\mathcal A} \Qhat_{k,h}(s,a),
\label{eq_vhat}
\end{align}
with terminal condition $\Qhat_{k,H+1}\equiv 0$.

Optimizing over the intersection $\C_k\cap\B_k$ is central to the algorithm design. When the offline data are informative, the set $\B_k$ can substantially tighten the confidence region and improve learning efficiency. Conversely, when the offline data are biased or uninformative, the online-only confidence set $\C_k$ serves as a safeguard, ensuring that the regret is never worse than that of an algorithm that ignores offline data entirely. This ``intersection-of-two'' principle has been adopted in several offline--online learning frameworks~\citep{cheung2024leveraging, zhang2025contextual}.

In episode $k$, the optimistic value functions $\{\Vhat_{k,h}\}_{h=1}^H$ induce a deterministic policy $\pi_k$ via greedy action selection: at each stage $h$ and state $s$, the action $\pi_{k,h}(s)$ is chosen to maximize $\Qhat_{k,h}(s,a)$ (line~16 of~\Cref{alg_main}). $\{\Vhat_{k,h}\}_{h=1}^H$ are also used to form the value-targeted regression variables from the data collected in episode $k$, which are incorporated into the parameter estimation step for episode $k+1$ (~\cref{eq_regression}).
\section{Regret Upper Bound and Characterization of Offline Data Informativeness}\label{sec_main_results}

In this section, we first derive the regret upper bound, \Cref{thm1}, for offline–online linear mixture MDP in~\Cref{sec_upper_bound}. 
Building on the regret bound, we characterize regimes in which offline data is informative in~\Cref{sec_when_offlinedata_informative}, and analyze how the regret depends on the feature mapping $\phi$ of the offline MDP, the offline behavior policy, and the design of $\{\TildeV_{m,h}\}$ in~\Cref{sec_comparison}.
We also compare~\Cref{thm1} with prior regret bounds in the literature in~\Cref{dis_comparison_new}.

\subsection{Regret Upper Bound}\label{sec_upper_bound}
Before presenting the regret upper bound for~\Cref{alg_main}, we introduce a parameter
$\tau$ that quantifies the effective coverage of the offline dataset. Define the offline design matrix
\begin{align}
    G_{\off} = \sum_{m=1}^{M^{\off}} \sum_{h=1}^{H} x^{\off}_{m,h}\bigl(x^{\off}_{m,h}\bigr)^{\top}.
    \label{eq_def_goff}
\end{align}
Intuitively, $\lambda_{\min}(\lambda I + G_{\off})$ characterizes the amount of information contributed by the offline data in the least informative direction of the feature space $\mathbb{R}^d$, and therefore governs the worst-direction uncertainty induced by offline samples. We formalize this notion by defining the effective coverage of the offline data as follows.

\begin{definition}\label{assumption_eigen}
For a fixed $\lambda > 0$, we define the {\bf $\delta$-effective coverage} as:
\begin{align}
\tau(\delta) := \sup\left\{ z\ge 0 \Bigm| \Pr\left( \frac{1}{M^{\off}}\lambda_{\min}(\lambda I + G_{\off}) \ge z \right)\geq  1 - \delta \right\},
\qquad 
\delta\in(0,1).
\label{eq_def_of_tau}
\end{align}
\end{definition}
Note that for any $\delta \in (0,1)$, the quantity $\tau(\delta)$ depends on the choice of $\{\TildeV_{m,h}\}$ and the offline behavior policy; we elaborate on this dependence in~\Cref{sec_comparison}.
Throughout the paper, we write $\tau = \tau\left(\frac{1}{2KM^{\off}}\right)$ (so $\delta = 1 / (2KM^\off)$) for notational convenience.

Regardless of the offline data, one always has $\lambda_{\min}(\lambda I + G_\off) \ge \lambda$, and hence the trivial lower bound $\tau \ge \lambda / M^\off$ holds.  Moreover, since the offline regression vectors $x^{\off}_{m,h}$ are uniformly bounded, $\tau$ is upper bounded by a constant, regardless of the behavior policy or the value of $M^{\off}$.
\Cref{assumption_eigen} is closely related to standard uniform coverage assumptions in offline reinforcement learning~\citep{nguyen2023instance,duan2020minimax,yin2022near}, and a more detailed discussion is provided in~\Cref{sec_comparison} and~\Cref{discussion_coverage}. 
Importantly, in~\Cref{discussion_coverage_2} we provide a counterexample showing that convential partial coverage conditions are insufficient in the offline--online setting. Under environment shift, the offline- and online-optimal policies may differ substantially, so learning an offline-optimal policy from offline data can be uninformative.

We are now ready to state our main regret upper bound.

\begin{restatable}{theorem}{RegretBound}
\label{thm1}
Under Assumptions~\ref{assump_theta_gap} and~\Cref{assumption_eigen}, let $\lambda = H^2 d$ and define 
\begin{align*}
\gamma_k = H \sqrt{ 2 \log \left( \frac{2\det(M^{\all}_{k})^{1/2}K}{\det(\lambda I)^{1/2}} \right) } +  \Delta \frac{\lambda_{\max}(G_{\off})}{\sqrt{\lambda_{\min}(\lambda I + G_{\off})}} + \sqrt{\lambda}B,
\quad
\beta_k = H \sqrt{ 2 \log \left( \frac{2\det(M^{\on}_{k})^{1/2}K}{\det(\lambda I)^{1/2}}\right)} + \sqrt{\lambda}B.
\end{align*}
Then, when running~\Cref{alg_main}, the regret satisfies
\begin{align}
\Regret(K) = \tilde{O}\left( \min \left\{
\left[
    BH^2d \sqrt{K} + H^4d^{\frac{3}{2}}\Delta\sqrt{K} \sqrt{\frac{M^{\off}}{\tau}}
\right] 
\times
\sqrt{\log\left( 1 + \frac{H^3K}{\tau M^{\off}} \right)}
, 
BH^2d \sqrt{K}
\right\}
\right), 
\label{eq_regret}
\end{align}
where $\tilde O(\cdot)$ hides logarithmic factors in $K$ and $M^{\off}$.
\end{restatable}

The regret bound has a ``min-of-two'' form. The first term corresponds to the offline–online estimator (term~$(A)$), while the second term (term~$(B)$) corresponds to the online-only estimator. The proof is provided in~\Cref{appendix_proofs}.

\subsection{When is Offline Data Informative}
\label{sec_when_offlinedata_informative}
\begin{figure}
    \centering
    \includegraphics[width=0.7\linewidth]{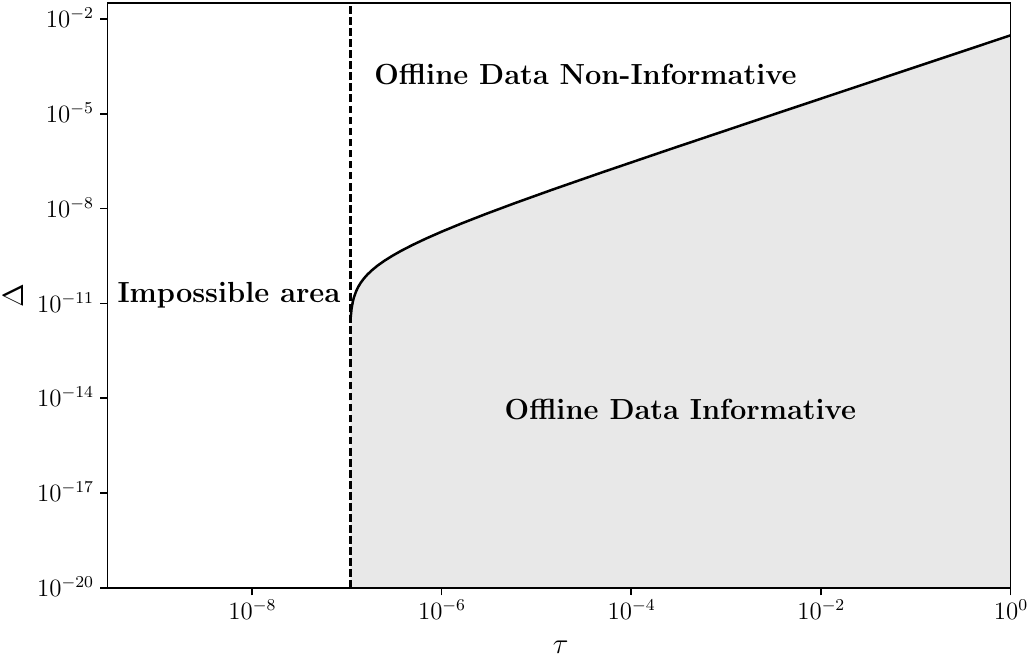}\caption{
    Graphical illustration of the sufficient conditions over the informativeness of offline data for $K = 10^3$, $M^{\off} = 1.5\times 10^{11}$, $\epsilon = 10^{-3}$, $H = 3$, $d = 5$, and $B = 10$. The admissible region is computed directly from~\cref{eq_regret}.
    When $\tau \lesssim 10^{-7}$, no value of $\Delta$ yields informative offline data. Once $\tau$ exceeds this threshold, larger $\tau$ admits a larger environment shift $\Delta$, with an approximately linear dependence. The deviation from linearity near $\tau \approx 10^{-7}$ is mainly caused by the logarithmic confidence term in~\cref{eq_regret}.
    }
    \label{fig_admissible}
\end{figure}

The regret bound in~\Cref{thm1} is always upper bounded by term~$(B)$, $\tilde O(B H^2 d\sqrt{K})$, which matches the online-only regret guarantee for linear mixture MDPs~\citep{jia2020model}.  This highlights the robustness of our algorithm.  When the offline data are uninformative (due to either poor coverage $\tau$ or a large environment shift $\Delta$), the algorithm provably performs no worse than ignoring offline data altogether. 

We next characterize regimes in which offline data is informative and yields a strict improvement over the online-only baseline.

\begin{definition}
\label{def_informative}
Given $\epsilon  \in (0, \frac{1}{2}]$, we say that the offline data is \textbf{informative} with respect to $\epsilon$ if the regret satisfies
\(
\Regret(K) = \tilde O\left(BH^2dK^{\frac{1}{2}-\epsilon}\right).
\)
\end{definition}

Any such improvement must be realized through term~$(A)$. The contribution of offline data improves monotonically as the environment shift $\Delta$ decreases and the coverage parameter $\tau$ increases.  If no nontrivial lower bound on $\tau$ can be established, namely when $\tau \leq \lambda / M^\off$, then offline data becomes uninformative once $K$ is sufficiently large.  This shows that nontrivial coverage is a necessary condition for offline data to be informative.
The following corollary gives a sufficient condition, stated in terms of $K$, $M^{\off}$, $\tau$, and $\Delta$, under which offline data is informative.

\begin{restatable}{corollary}{Maincorollary}
\label{cor_obervation_upper_bound}
If there exists $\epsilon \in (0, \frac{1}{2}]$ such that
\begin{align}
\tau = \Omega\left(H^3\frac{K^{1 + 2\epsilon}}{M^{\off}}\right)
\quad\text{and}\quad
\frac{\Delta}{\tau} = O\left(\frac{B}{H^{7/2} d^{1/2}K^{\frac{1}{2} + \epsilon}}\right),
\end{align}
then the offline data is informative with respect to $\epsilon$.
\end{restatable}

The proof is deferred to~\Cref{proof_observation_upper_bound}. We next discuss how the coverage parameter $\tau$ and the environment shift $\Delta$ affect the regret bound, treating $K$ and $M^{\off}$ as fixed.

\paragraph{Case 1: $\tau = \Theta(1)$, or equivalently, $\lambda_{\min}(G_\off) = \Omega(M^\off)$ with high probability.}
This corresponds to the ideal regime in which the offline data exhibit uniform coverage over the entire state--action space.
In this case, \Cref{cor_obervation_upper_bound} simplifies to the conditions $M^{\off} \gtrsim K^{1+2\epsilon}$ and $\Delta \lesssim K^{-1/2-\epsilon}$, up to constant factors. That is, when the amount of offline data is sufficiently large relative to $K$, and the environment shift $\Delta$ is suitably small, the offline data become informative.

A noteworthy implication is that offline data need not be excessively large to yield substantial gains. For any $\epsilon \in (0, \frac{1}{2}]$, when $M^{\off} = \Theta(K^2)$, whether offline data is informative depends solely on $\Delta$. In particular, under a sufficiently small shift, e.g., $\Delta \lesssim K^{-1}$, the regret reduces to $O(\log (K))$. This improvement indicates that, at an appropriate scale, informative offline data can substantially reduce uncertainty about the online environment before interaction begins.

\paragraph{Case 2: Decaying Coverage $\tau$ as $M^{\off}$ Increases.}
Unlike Case~1, where the coverage parameter $\tau$ is assumed to be bounded away from $0$, we now consider regimes in which $\tau$ may decay as $M^{\off}$ increases. Such behavior can arise when the offline behavior policy is \emph{adaptive}, meaning that it depends on previously collected data. As $M^{\off}$ grows, the policy may concentrate more heavily on certain state--action pairs and reduce visits to others, thereby decreasing the effective coverage parameter $\tau$. See~\citet{nguyen2023instance, jin2020provably} for further discussion on adaptivity in offline RL. We next analyze how the scaling of $\tau$ with $M^{\off}$ affects the informativeness of the offline data.

The first condition in~\Cref{cor_obervation_upper_bound}, $\tau = \Omega(H^3 K^{1+2\epsilon} / M^{\off})$ imposes a minimum coverage requirement. Given $K$ and $M^{\off}$, the effective coverage $\tau$ must exceed a threshold for the offline data to be informative. The second condition,
$\Delta / \tau = O(B \left(H^{7/2} d^{1/2} K^{1/2+\epsilon}\right)^{-1})$, shows that the admissible magnitude of the environment shift $\Delta$ scales {\em linearly} with $\tau$.
Together, these sufficient conditions indicate that stronger effective coverage (i.e., a slower decay of $\tau$ with respect to $M^{\off}$) linearly enlarges the range of environment shifts under which offline data is informative. A similar argument suggests an upper bound on the offline data size $M^{\off}$ that is sufficient to attain a $\tilde O(\log (K))$ regret, analogous to Case~1. We omit the details for brevity.

The discussion above is summarized visually in~\Cref{fig_admissible}.

\subsection{Dependence of $\tau$ on the Feature Mapping $\phi$, the Offline Behavior Policy, and the Design of $\{\TildeV_{m,h}\}$}
\label{sec_comparison}

The guarantee in \Cref{thm1} critically depends on the coverage parameter $\tau$. In this subsection, we study how $\tau$ is shaped by three components: the feature mapping $\phi$, the offline behavior policy, and the design of $\{\TildeV_{m,h}\}$.  The feature mapping $\phi$ determines whether a nontrivial lower bound on $\tau$ is possible, the offline behavior policy determines which state--action pairs are visited in the offline data, and the design of $\{\TildeV_{m,h}\}$ determines how these visits translate into the offline design matrix $G_{\off}$.

We first study a necessary condition on $\phi$.  We then consider the canonical uniform data coverage assumption on the offline behavior policy, under which we discuss the designs of $\{\TildeV_{m,h}\}$ and derive the corresponding lower bounds on $\tau$.  Finally, we discuss adaptive offline behavior policies, under which $\tau$ may decay with $M^{\off}$.  For clarity of exposition, we assume throughout this subsection that the state and action spaces are finite.  The extension to continuous spaces is similar and thus omitted.

\paragraph{The Feature Mapping $\phi$.}
We begin by isolating a structural property of the feature mapping $\phi$ that is necessary for $\tau$ to admit a nontrivial lower bound. Recall that the feature mapping $\phi$ of the offline MDP is assumed to be known to the learner. We say that the feature mapping of the offline MDP is \emph{uniformly learnable} if the following condition holds. The term ``uniformly'' refers to the fact that, in~\Cref{assumption_eigen}, the informativeness of the offline data is measured via a minimum eigenvalue condition over the entire feature space.

\begin{definition}
\label{assumption_offlineRL_span_main}
The feature mapping $\phi$ of the offline MDP is {\bf uniformly learnable} if there exist $L$ probe vectors $\{V_1,\ldots,V_L\}$ with
$V_l \in [-1,1]^{|\S|}$ for all $l \in [L]$, and state--action pairs $\{(s_i,a_i)\}_{i=1}^{I}$, such that
\begin{align}
\lambda_{\min}\left( \sum_{i=1}^{I}\sum_{l=1}^L \phi(\cdot \mid s_i,a_i) V_l \left(\phi(\cdot \mid s_i,a_i) V_l\right)^{\top}
\right) \ge \kappa,
\label{eq_uniform_learnability}
\end{align}
for some constant $\kappa>0$, where $I$ and $L$ are fixed positive integers.
\end{definition}

Uniform learnability is a property of the feature mapping of the offline MDP alone and does not depend on the behavior policy or the choice of $\{\TildeV_{m,h}\}$.   It formalizes the requirement that the feature mapping $\phi$ admits sufficiently diverse directions under suitable probing. The following result shows that this condition is necessary for the offline data to be informative.

\begin{restatable}{proposition}{PropositionLearnability}
\label{proposition_uniform_learnability}
For a given offline–online linear mixture MDP, if the feature mapping of the offline MDP is not uniformly learnable, then for any behavior policy and any design of $\{\TildeV_{m,h}\}$, we have $\tau(\delta) = \frac{\lambda}{M^{\off}}$ for any $\delta \in (0,1)$. In this case, the offline data is uninformative.
\end{restatable}
The proof of~\Cref{proposition_uniform_learnability} is deferred to~\Cref{appendix_proof_3}. 

\begin{remark}
\label{remark_learnability}
The one-hot feature mapping for any offline--online tabular MDP (see~\Cref{remark_tabularMDP_equivalence}) satisfies uniform learnability with $I = |\S||\A|$, $L = |\S|$, $V_{s'} = e_{s'}$ for all $s' \in \S$, and $\kappa = 1$, where $e_{s'} \in \mathbb{R}^{|\S|}$ denotes the canonical basis vector corresponding to state $s'$ in the one-hot representation.
In this case, for any $(s,a)$ we have $\phi(\cdot \mid s,a) V_{s'} V_{s'}^\top \phi(\cdot \mid s,a)^\top = e_{s'}e_{s'}^\top$.
Furthermore, since we assume finite state and action spaces in this subsection, any offline--online linear mixture MDP admits a tabular reformulation, under which the one-hot feature mapping of the offline MDP is uniformly learnable.
However, uniform learnability is defined with respect to the given linear mixture representation with dimension $d$, rather than the reformulated tabular one.
\end{remark}

\paragraph{Offline Behavior Policy.}
A coverage condition on the offline behavior policy is also required for the offline data to be informative.
Intuitively, even if the feature mapping of the offline MDP is uniformly learnable, the offline behavior policy must visit the specific $(s,a)$ pairs in~\Cref{assumption_offlineRL_span_main} in order to obtain a nontrivial lower bound on $\tau$ implied by uniform learnability.
Inspired by the offline RL literature~\citep{nguyen2023instance, jin2021pessimism, yin2022near, uehara2021pessimistic}, we focus on the uniform data coverage assumption~\citep{nguyen2021sample, nguyen2023instance}, which is a canonical assumption in offline RL.
The discussion of other coverage assumptions, including partial data coverage~\citep{nguyen2023instance} or feature-space coverage~\citep{yin2022near}, is deferred to~\Cref{discussion_coverage}. 

Formally, for any behavior policy $\pi^{\off}$, define the marginal state--action probability measure at stage $h$ in the offline MDP as
\[
    d_h^{\pi^{\off}}(s,a) := \Pr(s_h = s, a_h = a),
\]
where the probability is taken with respect to the trajectory distribution induced by policy $\pi^{\off}$ and the offline environment dynamics. The uniform data coverage assumption is stated as follows.

\begin{assumption}
\label{assumption_data_coverage_main}
In the offline phase, the behavior policy is non-adaptive, so the same policy is used in every episode. That is, $\pi_m^{\off} = \pi^{\off}$ for all $m \in [M^{\off}]$. Moreover, the probability of visiting any $(h,s,a)$ triple is lower bounded by $p_0 > 0$, i.e.,
\[
    \inf_{h,s,a} d_h^{\pi^{\off}}(s,a) \ge p_0.
\]
\end{assumption}

Under this assumption, we next propose two constructions of $\{\TildeV_{m,h}\}$ that yield nontrivial lower bounds on $\tau$.

\paragraph{Design of $\{\TildeV_{m,h}\}$.}
Even under uniform learnability and offline data coverage, however, a well-designed choice of $\{\TildeV_{m,h}\}$ is \emph{necessary} to achieve a nontrivial value of $\tau$. We illustrate this via a simple counterexample.

Consider an offline–online tabular MDP embedded into an offline–online linear mixture MDP via~\Cref{remark_tabularMDP_equivalence}. In this case, the offline MDP is uniformly learnable with probe vectors chosen as unit vectors. Moreover, under an appropriate design of $\{\TildeV_{m,h}\}$ and offline behavior policy, one can obtain $\tau = \Theta(1)$, as in~\Cref{remark_learnability}. However, if one sets $\TildeV_{m,h} = \mathbf{1}_{|\S|}$ for all $m,h$, the resulting offline design matrix $G_\off$ is not full rank, yielding $\tau = \frac{\lambda}{M^\off}$ regardless of the offline behavior policy. This shows that a well-designed $\{\TildeV_{m,h}\}$ is necessary.

To maximize $\tau$, a natural principle is to select $\{\TildeV_{m,h}\}$ so as to increase the minimum eigenvalue of the offline design matrix $G_\off$. Formally, one may attempt to choose $\TildeV_{m,h+1}$ sequentially and greedily according to
\begin{align}
\TildeV_{m,h+1} \in \arg\max_{V \in \mathbb [-1,1]^{|\S|}} \lambda_{\min}\Biggl(\sum_{(m',h') < (m,h)} x^{\off}_{m',h'}(\TildeV_{m',h'+1})\bigl(x^{\off}_{m',h'}(\TildeV_{m',h'+1})\bigr)^{\top} + x^{\off}_{m,h}(V) \bigl(x^{\off}_{m,h}(V)\bigr)^{\top}
\Biggr)
\label{eq_principle_tildev}
\end{align}
for each realized sample path, where $x^{\off}_{m,h}(V) = \phi(\cdot \mid s^{\off}_{m,h}, a^{\off}_{m,h}) V$ and $\TildeV_{m,H+1} = 0$ for all $m$. The set $[-1,1]^{|\S|}$ can be replaced by any bounded subset of $\mathbb{R}^{|\S|}$. 
However, the objective in~\cref{eq_principle_tildev} is in general neither convex nor concave in $V$; see~\Cref{discussion_concavity_tau} for a counterexample. Accordingly, we propose two tractable alternatives to solving \cref{eq_principle_tildev}.

The first approximation adopts a stochastic design of $\{\TildeV_{m,h}\}$. 
At a high level, to maximize the minimum eigenvalue of $G_\off$, one should probe the feature space in an “even” manner. 
\begin{definition}
\label{def_construction_tildeV}
For each offline index $(m,h)$, let  $\xi_{m,h}=(\xi_{m,h,1},\ldots,\xi_{m,h,L})$ be a vector of i.i.d. Rademacher random variables, i.e., $\Pr(\xi_{m,h,l} = 1) = \Pr(\xi_{m,h,l} = -1) = \frac{1}{2}$ for all $l \in [L]$, independent of the offline data. We define the stochastic design via
\[
\TildeV_{m,h} = \frac{1}{L}\sum_{l=1}^L \xi_{m,h,l}  V_l .
\]
By construction, $\TildeV_{m,h} \in [-1,1]^{|\S|}$ for all $(m,h)$.
\end{definition}

Under~\Cref{assumption_offlineRL_span_main}, the $L$ probe vectors span the entire feature space. 
Thus, for any visited $(s,a)$ pair in the offline data, we construct $\TildeV_{m,h}$ by randomly combining these probe vectors, thereby implementing the idea of probing the feature space in an “even” manner.
This choice of $\{\TildeV_{m,h}\}$ ensures that all directions in the feature space are probed, leading to adequate coverage.  Indeed, we can show:
\begin{restatable}{corollary}{CorollaryTauUniformCoverage}
\label{cor_uniformdata_coverage}
Suppose the feature mapping of the offline MDP is uniformly learnable (\Cref{assumption_offlineRL_span_main}) and the offline behavior policy satisfies uniform data coverage (\Cref{assumption_data_coverage_main}).  Then under the stochastic design of $\{\TildeV_{m,h}\}$ (\Cref{def_construction_tildeV}), for any $\delta>0$, the effective coverage $\tau(\delta)$ satisfies
\begin{align}
\tau(\delta) \ge \frac{p_0 \kappa}{L^2} - \sqrt{\frac{2 d p_0 \kappa}{L^2 M^{\off}}\log\left(\frac{d}{\delta}\right)}.
\label{eq_bound_stochastic_design_under_delta}
\end{align}
In particular, when $M^\off \ge \frac{8dL^2}{p_0\kappa}\log(2dKM^\off)$,we have
\begin{align}
\tau = \tau\left(\frac{1}{2KM^{\off}}\right) \ge \frac{p_0 \kappa}{2L^2}.
\label{eq_bound_stochastic_design}
\end{align}
\end{restatable}
The proof of~\Cref{cor_uniformdata_coverage} is provided in~\Cref{appendix_proof_3}. 

Next up we consider the offline--online tabular setting under the natural embedding in \Cref{remark_tabularMDP_equivalence}.  In this case, we can obtain a simpler approximation of~\cref{eq_principle_tildev} and provide a deterministic design of $\{\TildeV_{m,h}\}$ as follows.

\begin{definition}
\label{def_construction_det_tildeV}
In the offline--online tabular MDP setting, under the embedding into the offline–online linear mixture MDP (\Cref{remark_tabularMDP_equivalence}), we define the deterministic design of $\{\TildeV_{m,h}\}$ as
\begin{align}
\TildeV_{m,h} = e_{s'}, \quad \text{for } 
s' \in \arg\min_{x\in\mathcal S} N_{m,h}(x,s,a),
\label{eq_principle_tildev_deterministic}
\end{align}
where $N_{m,h}(x,s,a)$ counts how many times the triple $(x,s,a)$ has been selected in previous offline samples until $(m,h)$.
\end{definition}

Note that in the tabular setting,~\cref{eq_principle_tildev_deterministic} coincides with~\cref{eq_principle_tildev}, except that the optimization is restricted to $V \in \{e_1,\ldots,e_{|\S|}\}$ rather than $V \in [-1,1]^{|\S|}$ as in~\cref{eq_principle_tildev}. Consequently, the value of $\tau$ obtained via~\cref{eq_principle_tildev_deterministic} provides a lower bound on that achieved by~\cref{eq_principle_tildev}. Moreover, similar to~\Cref{cor_uniformdata_coverage}, the deterministic design also implies that $\tau = \Theta(1)$, as stated in the following corollary.

\begin{restatable}{corollary}{CorollaryTauDeterministicUniformCoverage}
\label{cor_uniformdata_coverage2}
In the offline--online tabular MDP setting, under the deterministic design of $\{\TildeV_{m,h}\}$ (\Cref{def_construction_det_tildeV}) and~\Cref{assumption_data_coverage_main}, for any $\delta > 0$ the effective coverage $\tau(\delta)$ satisifes
\begin{align}
    \tau(\delta) \ge \frac{p_0}{|\S|} - \frac{1}{|\S|}\sqrt{\frac{1}{2M^\off}\log\left(\frac{|\S||\A|}{\delta}\right)}.
    \label{eq_bound_det_design_under_delta}
\end{align}
In particular, when $M^\off \ge \frac{2}{p_0^2}\log(2|\S||\A|KM^\off)$, we have that
\begin{align}
\tau = \tau\left(\frac{1}{2KM^\off}\right) \ge \frac{p_0}{2|\S|}.
\label{eq_bound_det_design}
\end{align}
\end{restatable}
In the offline–online tabular MDP setting, the feature mapping of the offline MDP is uniformly learnable (see~\Cref{remark_learnability}); therefore, the assumption of uniform learnability is omitted in~\Cref{cor_uniformdata_coverage2}. The proof of~\Cref{cor_uniformdata_coverage2} is also deferred to~\Cref{appendix_proof_3}. 

Note that both~\Cref{cor_uniformdata_coverage} and~\Cref{cor_uniformdata_coverage2} imply that $\tau = \Theta(1)$. Thus Case~1 in~\Cref{sec_when_offlinedata_informative} applies and under appropriate values of $M^\off$, $K$, and $\Delta$, the offline data could be informative.
We conclude by noting that our results recover several existing results in the literature, while explicitly characterizing how both the design of $\{\TildeV_{m,h}\}$ and the offline behavior policy shape the coverage parameter $\tau$.
A detailed discussion of these connections is provided in~\Cref{dis_comparison_new}, including offline–online tabular and linear bandits~\citep{cheung2024leveraging, zhang2025contextual}, offline–online tabular MDPs~\citep{chen2022data}, and offline linear mixture MDPs~\citep{nguyen2023instance}.

\paragraph{Adaptive Offline Behavior Policy and Decaying Coverage $\tau$.}
Although the analysis above mainly focuses on the regime $\tau = \Theta(1)$ under the uniform data coverage assumption (\Cref{assumption_data_coverage_main}), this assumption can be overly idealized, as it essentially requires the offline behavior policies $\{\pi^{\off}_m\}_{m=1}^{M^\off}$ to be designed for exploration.
In practice, the offline data may instead be generated by learning algorithms that aim to minimize regret in the offline environment.
In this case the behavior policy becomes adaptive, and the visitation distribution gradually concentrates around the optimal policy of the offline MDP rather than exploring the state--action space uniformly. 

We show that even in this setting the offline data can remain informative under our algorithm~\ALG. This corresponds to Case~2 of \Cref{sec_when_offlinedata_informative}, where the coverage parameter $\tau$ may decay with $M^\off$. 
To illustrate this phenomenon, we consider an offline--online tabular bandit instance with $A$ arms, where the offline behavior policies follow the \textsf{UCB1} algorithm \citep{auer2002finite}. 
In this instance, let $\mu^{\off}(i)$ and $\mu(i)$ denote the expected rewards of arm $i \in [A]$ in the offline and online environments, respectively. In the offline environment, the reward obtained at $m \in [M^\off]$ by pulling arm $i \in [A]$ follows $\mu^{\off}(i) + \xi_m$, where $\{\xi_m\}_{m \in [M^{\off}]}$ are i.i.d.\ $N(0,1)$ noises. The reward model in the online environment is defined analogously.
Let $\mu^{\off,*}$ and $\mu^*$ denote the expected rewards of the optimal arms in the offline and online environments, respectively; the optimal arms in the two environments may differ. Define
\[
    \Dgap = \sup_{i \in [A]} \mu^{\off,*} - \mu^{\off}(i)
\]
as the maximum suboptimality gap in the offline environment.
Such an instance admits an equivalent offline--online linear mixture MDP reformulation (see~\Cref{remark_tabularMDP_equivalence}). For this reformulation, we assume that \Cref{assump_theta_gap} holds, namely that the environment shift is upper bounded by $\Delta$ and is known.
The following corollary characterizes the scaling of $\tau$ and shows that the offline data can remain informative under appropriate values of $M^\off$, $K$, and $\Delta$, whose admissible values depend on $\Dgap$.

\begin{restatable}{proposition}{PropositionAdaptivePoilicy}
\label{instance_adaptive_policy}
Consider the aforementioned offline--online tabular bandit setting, where the offline behavior policies follow the \textsf{UCB1} algorithm with exploration rate $\gamma_{M^\off} = \sqrt{2\log (M^\off)}$ (See Algorithm~1 of~\citet{han2024ucb}).
If $K = o((\log (M^{\off}))^{49})$\footnote{This condition arises from the high-probability bound on the number of pulls in a bandit model provided by Proposition~3.7 of~\citet{han2024ucb}. 
The exponent $49$ is not intrinsic to our framework and results from matching the failure probability with the confidence level in the definition of $\tau$. 
A sharper analysis could potentially relax this requirement.}, and there exists $p \in (0, \frac{1}{2}]$ such that
\(
    \Dgap = O((M^{\off})^{-p}),
\)
then under the deterministic design of $\{\TildeV_{m,h}\}$ (\Cref{def_construction_det_tildeV}), running~\ALG yields that 
\(
\tau = \Omega((M^{\off})^{2p-1}).
\)
Furthermore, if there exists $\epsilon \in (0, \frac{1}{2}]$ such that
\begin{align}
    (M^{\off})^{2p} = \Omega(K^{1 + 2\epsilon}), 
    \qquad 
    \Delta = O\left(\frac{(M^{\off})^{2p-1}}{K^{\frac{1}{2} + \epsilon}}\right),
\label{eq_bandit_adaptive}
\end{align}
then the offline data is informative.
\end{restatable}

The proof of~\Cref{instance_adaptive_policy} is provided in~\Cref{appendix_proof_3}. When $\Dgap$ decreases, meaning that the expected rewards of different arms in the offline bandit problem become closer, \textsf{UCB1} explores suboptimal arms more frequently, which leads to a larger effective coverage parameter $\tau$. As a result, a larger value of $\Delta$ can be tolerated while the offline data remains informative.
This illustrates the relation between $\tau$ and $\Delta$ in \Cref{fig_admissible} when the regime $\tau = \Theta(1)$ no longer holds. 
Note that $\Dgap$ affects the online phase in two ways. A smaller $\Dgap$ (together with a small $\Delta$) makes the online bandit problem harder due to smaller suboptimality gaps, while it also induces more exploration in the offline phase under \textsf{UCB1} and leads to a larger $\tau$, making the offline data more informative. 
Intuitively, this reveals a somewhat counterintuitive phenomenon: harder online instances may allow the offline data to provide greater benefit.

\section{Regret Lower Bound}\label{sec_lower_bound}

We complement our upper bound in~\Cref{thm1} with a nearly matching lower bound when $\tau = \Theta(1)$.

\begin{restatable}{theorem}{Lowerbound}
\label{thm2}
Under environment shift (\Cref{assump_theta_gap}) and coverage condition (\Cref{assumption_eigen}), for any $\Delta, H \ge 2$, and $d \ge 2$, for any sufficiently large $K$ and $M^{\off}$~\footnote{The requirement that $K$ and $M^\off$ are sufficiently large in~\Cref{thm2} arises from asymptotic notations, such as $\tau = \Theta(1)$, and is not intrinsic to the result. In particular, it suffices that $K = \Omega(d^2)$ ad $M^\off = \Omega(d^2)$.}, and any offline--online algorithm, there exists an offline--online linear mixture MDP instance such that, whenever the offline data induces $\tau = \Theta(1)$, any algorithm has regret at least 
\begin{align}
\Omega\left(\min\left\{dH\sqrt{K}, dH\sqrt{K}\left(\sqrt{\log\left(1+\frac{K}{\lambda_{\min}(\Ex[G_\off])}\right)}(1 + \frac{\Delta}{\sqrt{d}}\sqrt{\lambda_{\min}(\Ex[G_\off])})\right)\right\}\right).
\label{eq_lowerbound_main_2}
\end{align}
\end{restatable}


The proof of~\Cref{thm2} is deferred to~\Cref{proof_lower_bound}. The lower bound in~\Cref{eq_lowerbound_main_2} exhibits a ``min-of-two'' structure, mirroring the upper bound in~\Cref{thm1}. 
The lower bound in~\cref{eq_lowerbound_main_2} matches the upper bound in~\cref{eq_regret} in the dependence on $d$ and $K$ when the online-only term $dH\sqrt{K}$ dominates, corresponding to the regime where the offline data is not informative. 
The gap in the dependence on $H$ is consistent with the linear mixture MDP literature~\citep{zhou2021nearly, jia2020model}, and can be reduced via variance-aware analysis as in~\citet{zhou2021nearly}.
When the offline--online term dominates, corresponding to the regime where the offline data is informative, the condition $\tau = \Theta(1)$ ensures that $\lambda_{\min}(\Ex[G_\off]) = \Theta(M^\off)$ (see~\cref{eq_lowerbound_3} in Appendix), under which the lower bound matches~\cref{eq_regret} in the dependence on $K$, $M^\off$, and $\Delta$.
The dependence on $d$ is also nearly matched, up to a gap between $d^{\frac{3}{2}}$ and $\sqrt{d}$ in the $\Delta$-dependent term. This gap, together with the gap in the dependence on $H$, stems from the use of self-normalized concentration (\Cref{lemma_concentration}) in the upper bound and from measuring offline data through the worst-direction eigenvalue in~\Cref{assumption_eigen}, which is consistent with the offline RL literature (see, e.g., Table~1 of~\citet{nguyen2023instance}).

We next discuss the lower bound in the regime $\tau = o(1)$ and show that this case leads to a different regret lower bound, as the dependence on $d$ in~\Cref{thm2} no longer applies.
To illustrate, consider an offline--online bandit problem with $A$ arms (see~\Cref{discussion_tabular_bandit}) with no environment shift (so $d = \Theta(A)$). Suppose that all but one arm are well-covered in the offline data, while the remaining arm is never explored in the offline phase. In this case, we have $\tau = O\left(\frac{1}{M^\off}\right)$ and $\lambda_{\min}(\Ex[G_\off]) = O(1)$, as discussed in~\Cref{assumption_eigen}. 
Substituting these quantities and $\Delta = 0$ into~\cref{eq_lowerbound_main_2} would yield a lower bound of order $dH\sqrt{K}$ from~\Cref{thm2}, which is incorrect; In this instance, the problem reduces to identifying a single poorly explored arm in the online phase, and the regret need not scale with $d$.
This example highlights that, when $\tau = o(1)$, the lower bound is governed more by the geometry of the offline data, as captured by $\lambda_{\min}(\Ex[G_\off])$, rather than by the dimension $d$. However, the relation $\lambda_{\min}(\Ex[G_\off]) = \Theta(M^\off)$ no longer holds in this regime. 
In principle, one can derive lower bounds for $\tau = o(1)$ following the same framework as in the proof of~\Cref{thm2}, under additional conditions on the offline behavior policy that characterize $\lambda_{\min}(\Ex[G_\off])$. We do not pursue this direction here, as it requires a detailed, instance-dependent characterization of the behavior policy, which is beyond the scope of this paper.
\section{Numerical Simulations}
\label{sec_simulations}

In this section, we further support the results in~\Cref{sec_main_results} with numerical experiments on a synthetic tabular MDP.  In particular, we study the regret dependence on $M^\off$, the coverage parameter $\tau$, the environment shift $\Delta$, and the choice of $\{\TildeV_{m,h}\}$.  We also investigate when historical data becomes informative, as characterized by the trade-off between the environment shift $\Delta$ and the coverage parameter $\tau$. See \url{https://github.com/Zhongjun24/Offline_Online_RL} for the code base.

\subsection{Baselines and Simulation Setting}

We implement \ALG under the following two designs for $\{\TildeV_{m,h}\}$:
\begin{itemize}
    \item \ALG(\textsf{Optimal}): $\{\TildeV_{m,h}\}$ is constructed according to the deterministic design in \Cref{sec_comparison}, noting that the underlying setting is a tabular MDP.
    \item \ALG(\textsf{Pessimistic})~\citep{nguyen2023instance}: Motivated by prior work that applies pessimism-based estimation in the offline phase of offline--online RL~\citep{zheng2023adaptive}, we also incorporate a pessimism-based variant for constructing $\{\TildeV_{m,h}\}$ by adapting Algorithm~2 of \citet{nguyen2023instance} in the offline phase. The online phase remains identical to \ALG.
\end{itemize}
We additionally compare \ALG against the following baselines:
\begin{itemize}
    \item \UCRL~\citep{jia2020model}: An algorithm specifically designed for online linear mixture MDPs. When no offline data are available, \ALG reduces to a standard online algorithm and coincides with \UCRL.
    \item \COMPLETE (Algorithm~\ref{alg_complete} in~\Cref{appendix_experiment}): This baseline closely follows the \COMPLETE approach in~\citet{chen2022data} and \textsf{HUCB1} for bandits~\citep{shivaswamy2012multi}, which assumes $\Delta=0$ and directly merges offline data into online learning.
\end{itemize}

We further compare~\ALG with additional baselines for tabular MDPs, including~\DPLSVI (Algorithm~2 in~\citet{chen2022data}) and~\UCBVI~\citep{azar2017minimax} in~\Cref{appendix_experiment}.
All experiments are conducted in a synthetic tabular MDP with $|\mathcal S|=5$, $|\mathcal A|=10$, and horizon $H=3$. For simplicity, we redefine $\Delta$ through the tabular transition discrepancy $\|P(\cdot \mid s,a) - P^{\off}(\cdot\mid s,a)\|_1 \leq \Delta.$
As discussed in~\Cref{discusson_delta_in_tabular_MDP}, this definition is consistent with~\Cref{assump_theta_gap} up to norm equivalence. Each experiment is repeated over 50 independent runs. Further details on the synthetic tabular MDP, behavior policy, baseline implementations, and experimental settings are provided in~\Cref{appendix_experiment}.

\begin{figure}[t!]
    \centering
    \begin{subfigure}[t]{0.48\textwidth}
        \centering
        \includegraphics[width=\linewidth]{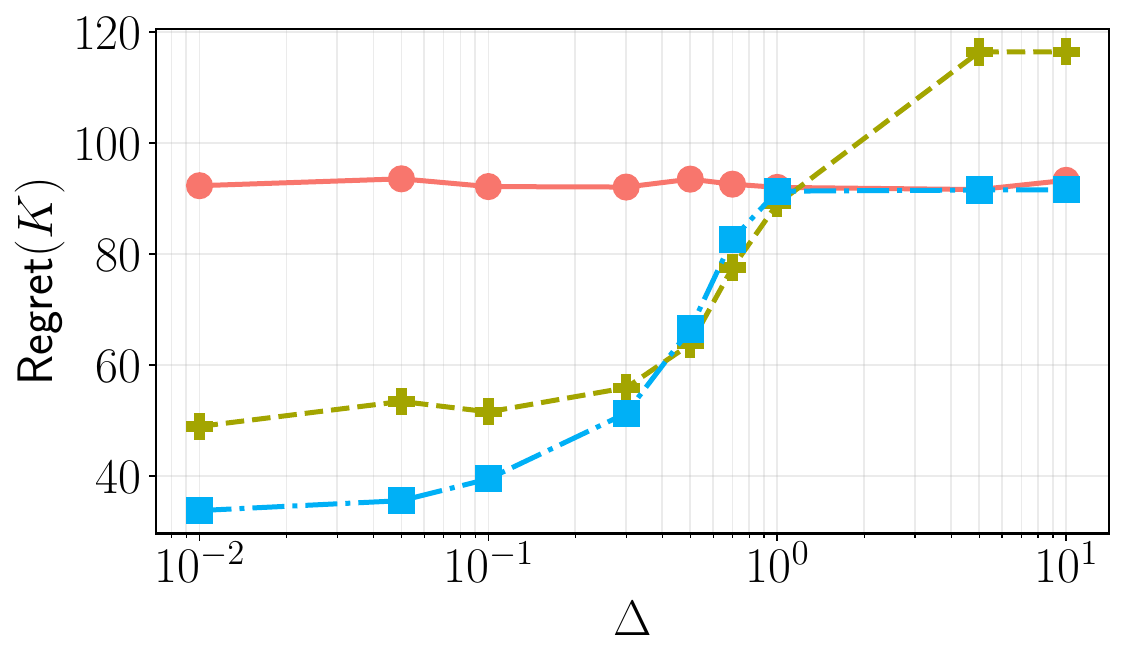}
        \caption{$\Regret(K)$ versus $\Delta$.}
        \label{fig2a}
    \end{subfigure}
    \hfill
    \begin{subfigure}[t]{0.48\textwidth}
        \centering
        \includegraphics[width=\linewidth]{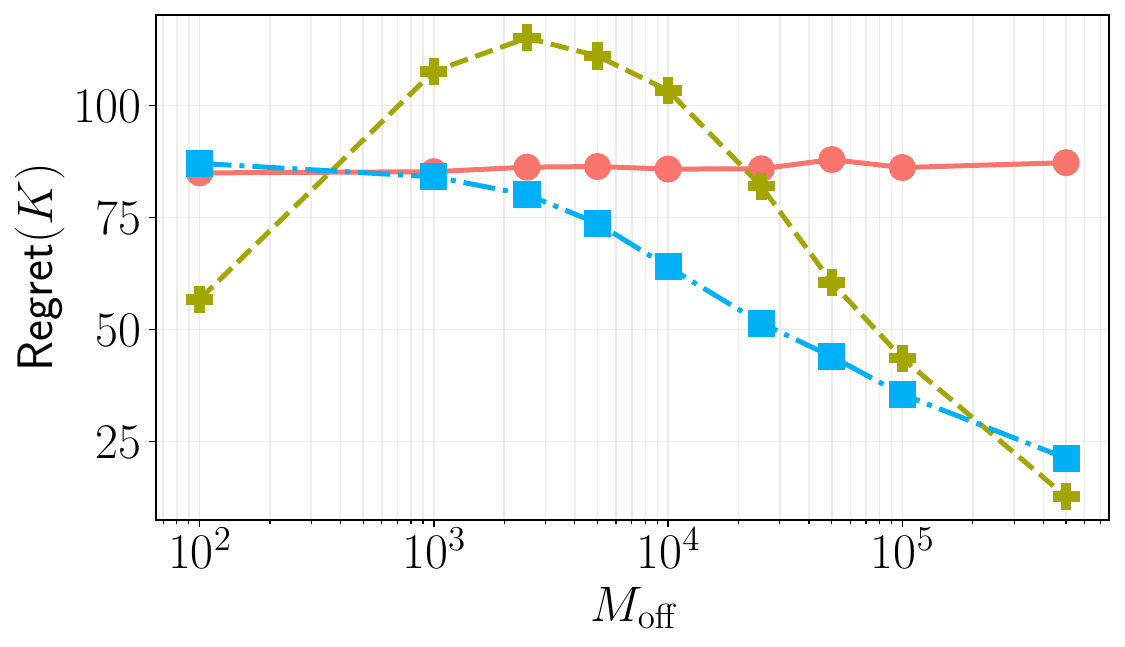}
        \caption{$\Regret(K)$ versus $M^\off$.}
        \label{fig2b}
    \end{subfigure}
    \hfill
    \begin{subfigure}[t]{0.48\textwidth}
        \centering
        \includegraphics[width=\linewidth]{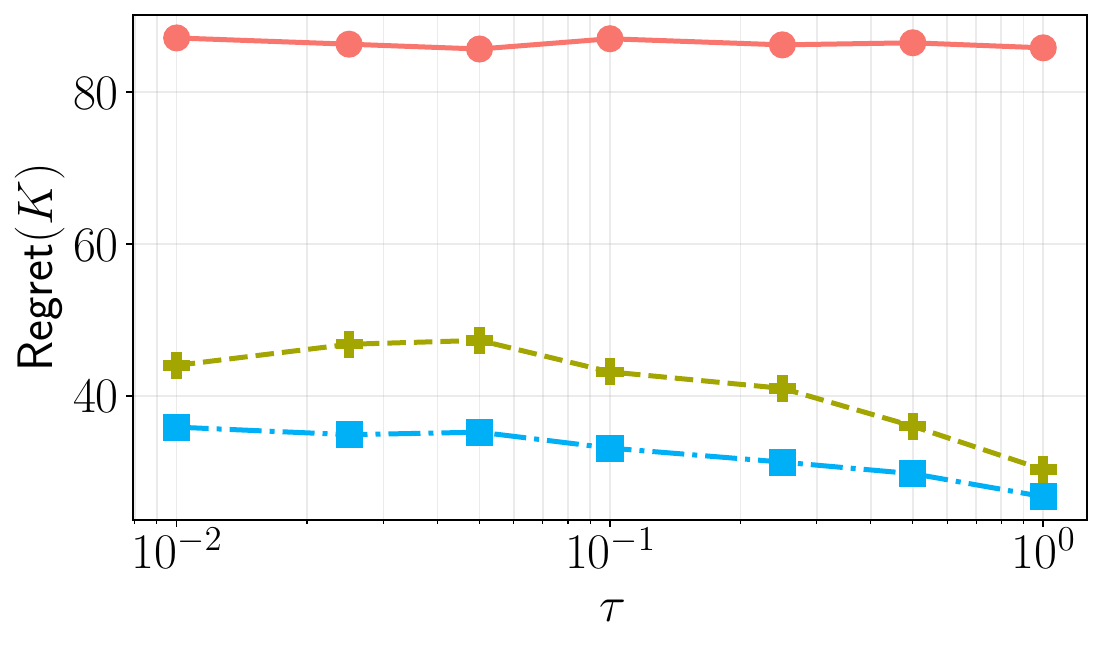}
        \caption{$\Regret(K)$ versus $\tau$.}
        \label{fig2c}
    \end{subfigure}
    \hfill
    \begin{subfigure}[t]{0.48\textwidth}
        \centering
        \includegraphics[width=\linewidth]{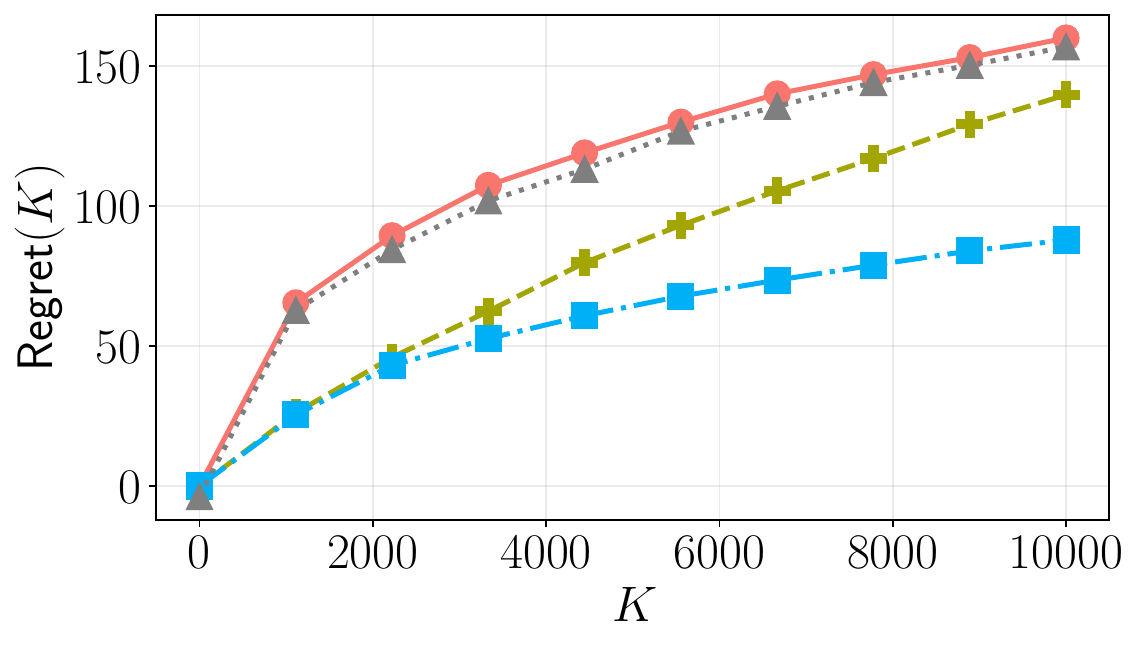}
        \caption{Cumulative regret over $K$.}
        \label{fig2d}
    \end{subfigure}
    \hfill  
    \vspace{0.4em}
    \begin{subfigure}[t]{1\textwidth}
        \centering
        \includegraphics[width=\linewidth]{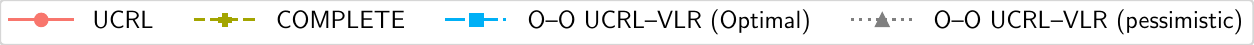}
    \end{subfigure}
    \caption{
    Comparison of \UCRL, \COMPLETE, \ALG(\textsf{Optimal}),
    and \ALG(\textsf{Pessimistic}) under different environment shifts $\Delta$, coverage $\tau$, $K$, and $M^\off$.
    }
    \label{fig_main}
\end{figure}

\subsection{Simulation Results}

\paragraph{Robustness to Environment Shift $\Delta$.} We first examine how performance varies with the environment shift $\Delta$. \cref{fig2a} plots the cumulative regret as $\Delta$ increases, with $\tau = \Theta(1)$, $K = 4\times 10^3$, and $M^\off = 10^5$ fixed.

As expected, \UCRL remains insensitive to $\Delta$ since it does not use any historical data. When $\Delta$ is small, we see that \ALG significantly outperforms \UCRL, demonstrating how our algorithm is able to effectively leverage offline data.  
As $\Delta$ grows, the regret of \ALG increases smoothly and approaches that of \UCRL, demonstrating the safeguarding property in~\Cref{thm1}. When offline data is uninformative, the algorithm gracefully falls back to purely online learning.

In contrast, methods that directly merge offline and online data without explicit robustness mechanisms, such as \COMPLETE, exhibit substantial performance degradation when $\Delta$ becomes large. This phenomenon underscores the robustness of~\ALG.

\paragraph{Impact of Offline Sample Size $M^\off$.} We next vary the offline sample size $M^\off$ while fixing $\Delta = 0.05$, $\tau = \Theta(1)$, and $K = 4\times 10^3$. \cref{fig2b} shows that once $M^\off \gtrsim 10^3$, \ALG begins to outperform \UCRL, indicating that the offline data have become informative relative to the online horizon $K$. As $M^\off$ increases further, both offline--online approaches improve. 
This behavior aligns with~\Cref{cor_obervation_upper_bound}, which predicts that the offline data become informative once their effective information content is comparable to that of online samples.

An interesting observation is that for small $M^\off$, the regret of \COMPLETE first increases and then decreases as $M^\off$ grows, unlike \ALG. Since \COMPLETE ignores the environment shift, when $M^\off$ is small, increasing $M^\off$ makes the estimator rely more heavily on the offline phase and become increasingly confident in a potentially misspecified model, thereby amplifying bias and increasing regret at $K$. As $M^\off$ becomes large, \COMPLETE effectively commits to the offline model and performs little to no exploration in the online phase. Given that the environment shift $\Delta = 0.05$ is small, this lack of exploration does not incur substantial loss for $K = 4\times10^3$, leading to a decrease in regret.

\paragraph{Role of Coverage $\tau$.} We next investigate the effect of the coverage parameter $\tau$, fixing $\Delta = 0.05$, $K = 4\times 10^3$, and $M^\off = 2 \times 10^5$. As shown in~\cref{fig2c}, the advantage of \ALG over \UCRL increases as $\tau$ grows. This observation is consistent with the analysis in \Cref{sec_main_results}.
When $\tau$ is small, \ALG does not coincide with \UCRL.  This is also consistent with the theory in~\Cref{sec_main_results}, since the regret bound depends on the worst-direction coverage, whereas in finite samples, certain well-covered directions may still provide empirical gains. 

\paragraph{Design of the Offline Value Function Estimates $\{\TildeV_{m,h}\}$.} \cref{fig2d} compares two  constructions of $\{\TildeV_{m,h}\}$ under $\Delta = 0.05$, $\tau = \Theta(1)$, and $M^{\off} = 10^5$.  Both \ALG(\textsf{Optimal}) and \ALG(\textsf{Pessimistic}) outperform the purely online baseline \UCRL. However, \ALG(\textsf{Optimal}) consistently attains substantially lower regret than \ALG(\textsf{Pessimistic}). 

This suggests that directly improving the statistical geometry of the offline Gram matrix $G_{\off}$ can be more effective than adopting a pessimism-driven construction in this setting. The result empirically validates the design principle proposed in \Cref{sec_comparison}.

\begin{figure}[t!]
    \centering
    \includegraphics[width=0.7\linewidth]{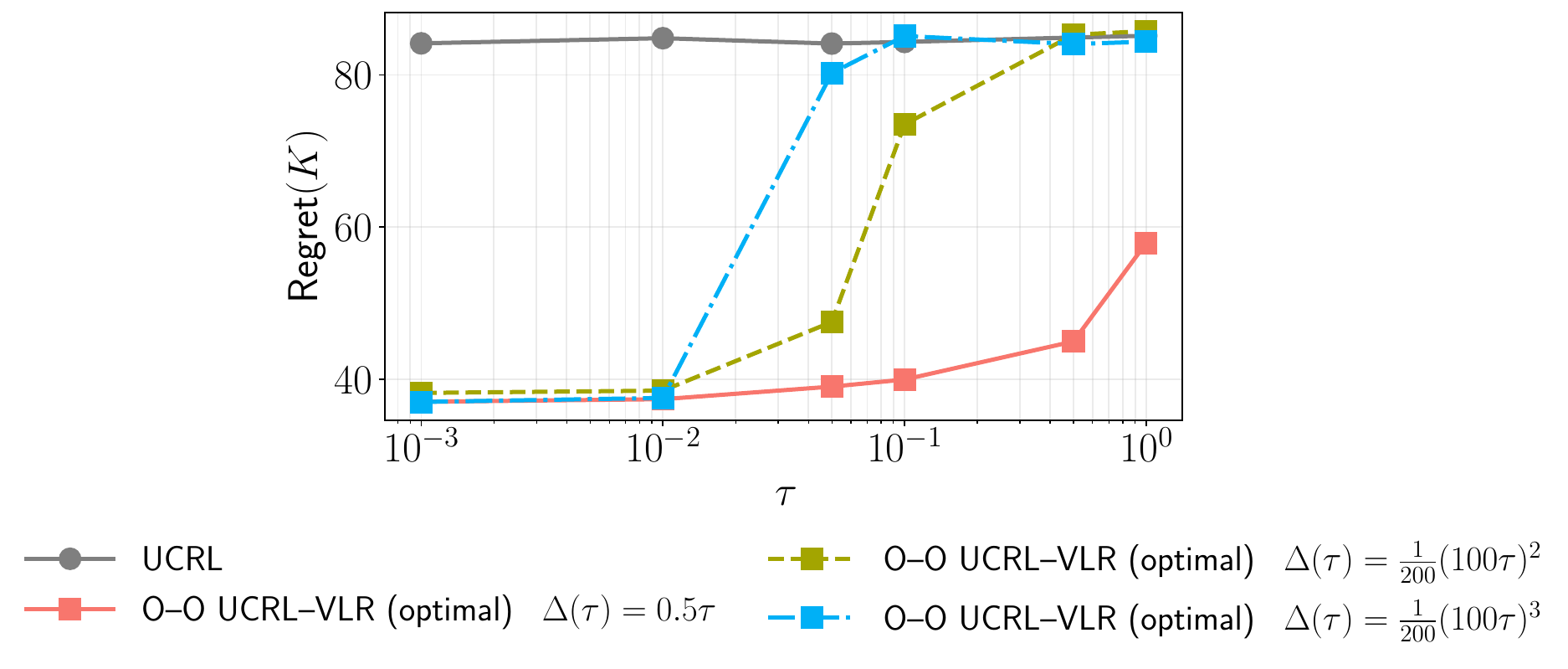}\caption{
Cumulative regret versus $\tau$ for different $\Delta(\tau)$, with $K = 4\times 10^3$ and $M^{\off} = 10^5$.}
    \label{fig_3}
\end{figure}

\paragraph{Interaction Between Environment Shift and Coverage.}
Finally, we examine how $\Delta$ and $\tau$ interact. \cref{fig_3} plots regret along trajectories $\Delta(\tau)$, with $K = 4\times 10^3$ and $M^{\off} = 10^5$ fixed, and all curves calibrated to share the same value of $\Delta$ at $\tau = 0.01$.

When $\Delta$ increases linearly with $\tau$, offline data remain informative even as $\Delta$ grows. In contrast, when $\Delta$ scales quadratically or cubically with $\tau$, the benefit of offline data rapidly disappears once $\tau$ exceeds $0.1$. These trends closely mirror the dependence predicted by \Cref{cor_obervation_upper_bound}, which indicates that offline data remain informative when $\Delta/\tau$ remains controlled. 

This highlights that the value of offline data is determined by the relative scaling between the environment shift $\Delta$ and coverage $\tau$.  In operational settings where a larger environment shift is anticipated (higher $\Delta$), the designer can proactively increase coverage (e.g., by broadening or randomizing the historical behavior policy) to ensure that $\tau$ grows commensurately. However, the coverage parameter $\tau$ need not be optimized to the maximum possible level (e.g., $\tau = \Theta(1)$); As long as the collected coverage $\tau$ scales appropriately with the shift, regret guarantees continue to improve and offline data remain informative.

\section{Conclusions}
\label{sec_conclusion}

In this work, we study offline--online learning for linear mixture MDPs under environment shift and an uncontrollable offline behavior policy. We develop an algorithm~\ALG for this setting, together with a regret upper bound and nearly matching lower bounds, and characterize when offline data is informative. 
Our results show that offline data is informative when the environment shift $\Delta$ is small and the coverage parameter $\tau$ is large, and that these two factors can compensate for each other. In particular, offline data can remain informative even under non-negligible environment shift, as long as $\Delta$ scales linearly with $\tau$. We further disentangle the factors that determine $\tau$, including the feature mapping of the linear mixture MDP, the offline behavior policy, and the design of offline value function estimates $\{\TildeV_{m,h}\}$.

Several directions for future work arise from our results. One natural question is whether the interplay between environment shift and offline behavior policy extends to broader function approximation settings in RL, such as offline--online learning with low-dimensional representations. Another direction is to go beyond the scalar quantity $\tau$ and study the full spectrum of the offline design matrix, which may provide a more refined characterization of offline data informativeness and potentially unify the lower bounds under different coverage regimes.


\clearpage
\newpage

\bibliographystyle{informs2014} 
\bibliography{references} 





\newpage

\AtBeginEnvironment{APPENDICES}{%
  \renewcommand{\theHsection}{appendix.\Alph{section}}%
  \renewcommand{\theHsubsection}{\theHsection.\arabic{subsection}}%
  \renewcommand{\theHsubsubsection}{\theHsubsection.\arabic{subsubsection}}%
  \renewcommand{\theHfigure}{\theHsection.\arabic{figure}}%
  \renewcommand{\theHtable}{\theHsection.\arabic{table}}%
  \renewcommand{\theHequation}{\theHsection.\arabic{equation}}%
  \renewcommand{\theHtheorem}{\theHsection.\arabic{theorem}}%
}
\makeatother

\crefalias{section}{appendix}
\begin{APPENDICES}
\OneAndAHalfSpacedXI 

\renewcommand{\arraystretch}{1.1}
\begin{table}[t!]
\caption{Common notation.}
\centering
\begin{tabular}{>{\color{edits}}l | >{\color{edits}}l}
\textbf{Symbol} & \textbf{Definition} \\ \hline

\multicolumn{2}{c}{Problem setting specifications} \\
\hline
$\mathcal S, \mathcal A$ & State space and action space \\
$H,K,d$ & Episode horizon, number of online episodes, and feature dimension of the linear mixture MDP \\
$\phi(s' \mid s,a)$ & Feature map $\phi:\mathcal S\times\mathcal A\times\mathcal S\to\mathbb R^d$ \\
$\phi_j(s' \mid s,a)$ & The $j$th coordinate of the feature vector $\phi(s' \mid s,a)$ \\
$P^{\off}, P$ & Offline and online transition probability \\
$\theta^{\off, *}, \theta^{*}$ & Offline and online transition parameters in the linear mixture model \\
$r^{\off}, r$ & Offline and online reward functions \\
$s_1^{\off}, s_1$ & Initial states of the offline and online MDPs \\
$B$ & Known bound on $\|\theta^*\|_2$ and $\|\theta^{\off,*}\|_2$ \\
$\Delta$ & Upper bound over the environment shift, $\|\theta^*-\theta^{\off,*}\|_2\le \Delta$ \\
\hline

\multicolumn{2}{c}{Offline data} \\
\hline
$\pi^{\off}_m$ & Offline behavior policy for the $m$-th trajectory\\
$\mathcal D_{\off}$ & Offline dataset $\{\mathcal H_m^{\off}\}_{m=1}^{M^{\off}}$ \\
$M^{\off}$ & Number of offline trajectories \\
$s^{\off}_{m,h}, a^{\off}_{m,h}$ & State and action at stage $h$ of offline trajectory $m$ \\
$\TildeV_{m,h}$ & Auxiliary value function used in offline regression \\
$x^{\off}_{m,h}$ & Offline feature $\phi(\cdot\mid s^{\off}_{m,h},a^{\off}_{m,h})\TildeV_{m,h+1}$ in the regression \\
$y^{\off}_{m,h}$ & Offline target $\TildeV_{m,h+1}(s^{\off}_{m,h+1})$ in the regression \\
$G_{\off}$ & Offline design matrix $\sum_{m,h} x^{\off}_{m,h}(x^{\off}_{m,h})^\top$ \\
$\tau$ & Offline data coverage parameter \\
\hline

\multicolumn{2}{c}{Online interaction} \\
\hline
$\pi^*$ & An optimal policy \\
$\pi_k$ & Policy executed in episode $k$ \\
$s^{\on}_{k,h}, a^{\on}_{k,h}$ & State and action at stage $h$ of episode $k$ \\
$\Vpi_h, \Qpi_h$ & Value and $Q$ functions at stage $h$ under policy $\pi$ \\
$x^{\on}_{k,h}$ & Online feature $\phi(\cdot\mid s_{k,h},a_{k,h})\Vhat_{k,h+1}$ in the regression\\
$y^{\on}_{k,h}$ & Online target $\Vhat_{k,h+1}(s_{k,h+1})$ in the regression\\
\hline

\multicolumn{2}{c}{Algorithm design} \\
\hline
$\lambda$ & Ridge regularization parameter \\
$M^{\all}_k, M^{\on}_k$ & Offline--online design matrix and online design matrix up to episode $k$ \\
$w^{\all}_k, w^{\on}_k$ 
& Response vectors in the offline--online and online-only ridge regressions up to episode $k$ \\
$\thetaall_k, \thetaon_k$ & Offline--online and online-only ridge estimator \\
$\gamma_k, \beta_k$ & Confidence radius for $\thetaall_k$ and $\thetaon_k$ \\
$\mathcal B_k, \mathcal C_k$ & Offline--online confidence set and online-only confidence set in episode $k$ \\
$\Vhat_{k,h}, \Qhat_{k,h}$ & Optimistic estimation of value and Q-functions in episode $k$ \\
\hline

\multicolumn{2}{c}{Regret and analysis} \\
\hline
$\Regret(K)$ & Expected cumulative regret over $K$ episodes \\
$\lambda_{\min}(\cdot), \lambda_{\max}(\cdot)$ & Minimum and maximum eigenvalues \\
$\|\cdot\|_A$ & Weighted norm $\|x\|_A=\sqrt{x^\top A x}$ ($A$ is positive definite) \\
$\tilde O(\cdot)$ & Big-O notation hiding logarithmic factors \\
\hline

\end{tabular}
\label{table:notation}
\end{table}
\clearpage
\section{Comparisons to Settings in the Literature}
\label{dis_comparison_new}

The offline--online linear mixture MDP setting studied in this paper subsumes several canonical models of interest, including offline--online tabular and linear bandits, offline--online tabular MDPs, and offline linear mixture MDPs. In this section, we show how these three settings can be embedded into our offline--online linear mixture MDP framework, and we compare our main result in \Cref{thm1} with the corresponding guarantees established for each setting in the literature.

\subsection{Offline--Online Tabular and Linear Bandits~\citep{cheung2024leveraging, zhang2025contextual}}
\label{discussion_tabular_bandit}

We start by examining the connection between~\Cref{thm1} and prior work on offline--online tabular bandits~\citep{cheung2024leveraging} and offline--online linear bandits~\citep{zhang2025contextual}. Since tabular bandits can be expressed as linear bandits through one-hot mapping, similar to \Cref{remark_tabularMDP_equivalence}, both models can be viewed as special cases of the offline--online linear mixture MDP framework. We show that (i) the environment shift assumption in \Cref{assump_theta_gap} is no stronger than those adopted in~\citep{cheung2024leveraging, zhang2025contextual}; (ii) how the coverage assumption in \Cref{assumption_eigen} connects with the coverage assumptions in these literature; and (iii) in benign regimes, the regret bound in \Cref{thm1} matches the order of the regret guarantee in~\citet{cheung2024leveraging}. Note that \citet{zhang2025contextual} consider a contextual model, and therefore the regret guarantees are not directly comparable.

\subsubsection{Comparison of Environment Shift Assumptions}
\label{discussion_tabularbandit_envshift}
In stochastic tabular bandits with $A$ arms~\citep{cheung2024leveraging}, the environment shift is characterized by an arm-wise reward discrepancy: for any arm $a$,
\begin{align}
    |\mu(a) - \mu^{\off}(a)| \le \Delta_1 
    \label{eq_tabularbandit}
\end{align}
where $\mu^{\off}(a)$ and $\mu(a)$ denote the expected rewards of arm $a$ in the offline and online environments, respectively. 
Similarly, in stochastic linear bandits~\citep{zhang2025contextual}, the expected reward under action $a \in \mathbb{R}^d$ is modeled as $\mu(a) = \langle \theta^*_B, a \rangle$ for the online environment, and $\mu^{\off}(a) = \langle \theta^{\off, *}_B, a \rangle$ for the offline environment.  The environment shift is then quantified by:
\begin{align}
    \|\theta^{\off,*}_{B} - \theta^{*}_{B}\|_2 \le \Delta_2.
    \label{eq_linearbandit}
\end{align}

Similar to MDPs, a tabular bandit can be represented as a linear bandit via a one-hot feature mapping where $\theta = (\mu(a_1), \ldots, \mu(a_A)) \in \mathbb{R}^A$.  Then the induced linear bandit environment shift,
\begin{align}
\Delta_2(\Delta_1) := \norm{\theta^{\off,*}_B - \theta^*_B}_2 \leq \sqrt{A} \Delta_1 \label{eq_equivalence_tabularbandit_linearbandit}
\end{align}
via the corresponding feature representation, where $A$ denotes the number of arms.
We therefore focus on embedding linear bandits satisfying \cref{eq_linearbandit} into offline--online linear mixture MDPs, which directly induces the relationship between~\cref{eq_tabularbandit} in a tabular bandit model and the environment shift $\Delta$ in \Cref{assump_theta_gap}.

We now describe a mapping between linear bandits and linear mixture MDPs. For simplicity, we consider a stochastic Bernoulli offline--online linear bandit. Given an offline--online linear bandit instance with parameters $\theta^{\off,*}_{B}$ and $\theta^{*}_{B}$, we construct an offline--online linear mixture MDP as follows. The state space consists of three states $\{s_1, s_2, s_3\}$, and the episode length is $H=2$.
At stage $h=1$, the initial state is always $s_1$ and the reward is zero. Upon taking action $a$, the system transitions from $s_1$ to $s_2$ with probability $\langle \theta^{\off,*}_{B}, a \rangle$ in the offline environment and $\langle \theta^{*}_{B}, a \rangle$ in the online environment. With the remaining probability, the system transitions to $s_3$. At stage $h=2$, the reward at $s_2$ is $1$, while the reward at $s_3$ is $0$.

This offline--online linear mixture MDP satisfies all assumptions in the main text.
In particular, the reward functions $r^{\off}$ and $r$ are known,
and the induced stochastic process is equivalent to that of the original linear bandit problem.
As a result, we obtain the following relationship between the environment shifts.

\begin{proposition}
\label{proposition_bandits}
Consider a stochastic offline--online linear bandit problem
for which \cref{eq_linearbandit} holds.
Let $(P^{\off}, r)$ and $(P, r)$ be the equivalent offline and online
linear mixture MDPs constructed above, with parameter vectors
$\theta^{\off,*}$ and $\theta^*$.
We define $\Delta(\Delta_2)$ as the induced environment shift bound under the constructed offline--online linear mixture MDP.
Then \Cref{assump_theta_gap} holds with
\(
    \Delta(\Delta_2) \le \Delta_2.
\)
\end{proposition}
\begin{rproof}
This follows immediately from the construction above, under which the linear mixture MDP parameters coincide with the corresponding linear bandit parameters.
\end{rproof}

Similarly, for any offline–online tabular bandit problem satisfying~\cref{eq_tabularbandit}, there exists an equivalent offline–online linear mixture MDP representation. Let $\Delta(\Delta_1)$ denote the induced environment shift bound under the resulting offline--online linear mixture MDP. We then directly obtain that 
\(
    \Delta(\Delta_1) \le \Delta_2(\Delta_1) \le \sqrt{A}\,\Delta_1,
\)
where the second inequality is from~\cref{eq_equivalence_tabularbandit_linearbandit}. 
Consequently, for any offline–online tabular bandit problem under~\cref{eq_tabularbandit}, or offline–online linear bandit problem under~\cref{eq_linearbandit}, \Cref{thm1} applies.

\subsubsection{Comparison of Coverage Assumptions and the Induced $\tau$}
\label{discussion_tabularbandit_coverage}
In this subsection, we compare \Cref{assumption_eigen} with the coverage assumptions adopted in~\citet{cheung2024leveraging, zhang2025contextual}.

\paragraph{Coverage Assumption in~\citet{cheung2024leveraging}.} 
In \citet{cheung2024leveraging}, the authors assume that the number of offline pulls for each arm is known. We now explain how this assumption specifies the corresponding value of $\tau$ in~\Cref{assumption_eigen} in our framework.

Consider a bandit problem with $A$ arms indexed by $\{a_1, \ldots, a_A\}$. Any finite-armed bandit can be embedded into a linear mixture MDP with three states, as described in \Cref{discussion_tabularbandit_envshift}. Since each tabular bandit is a special case of a tabular MDP, the deterministic construction of $\{\TildeV_{m,h}\}$ in \Cref{sec_comparison} applies directly.
For each $i \in [A]$, $\phi(\cdot \mid s_1,a_i)$ has exactly
one nonzero entry in each column. 
Thus under the deterministic design of $\{\TildeV_{m,h}\}$, namely by selecting the auxiliary value functions $\TildeV_m(s_1, a_i)$ cyclically from the standard basis $\{(1,0,0), (0,1,0), (0,0,1)\}$ for each arm $a_i$, the resulting offline design matrix has the form
\begin{align}
G_{\off} = \operatorname{diag}\Bigl(\frac{N(a_1)}{3}, \frac{N(a_1)}{3}, \frac{N(a_1)}{3}, \frac{N(a_2)}{3}, \ldots, \frac{N(a_A)}{3} \Bigr),
\label{eq_tabular_bandit_deterministic_design}
\end{align}
where $N(a_i)$ denotes the number of times arm $a_i$ is pulled in the offline data. For simplicity, we assume that $N(a_i)$ is divisible by $3$ for all $i \in [A]$. 
As a result, we have that
\(
\tau \ge \min_{i \in [A]} \frac{N(a_i) + \lambda}{3M^{\off}}.
\)
Therefore, the value of $\tau$ depends on the allocation of offline pulls. When the pulls are distributed uniformly across arms, we obtain $\tau \ge \frac{1}{3A}$. In contrast, under highly uneven allocations, one can only guarantee the trivial lower bound $\tau \ge \frac{1 + \lambda}{3 M^\off}$ in the worst case.

\paragraph{Coverage Assumption in~\citet{zhang2025contextual}.} 
Although \citet{zhang2025contextual} study a contextual model that differs from our linear mixture MDP framework, and therefore a direct relation between the two settings is not available as in the previous case, their coverage assumption is closely related at a high level to \Cref{assumption_eigen}.
In the offline phase, they assume a dataset of size $M^\off$, where each sample indexed by $m \in [M^\off]$ is associated with a contextual feature vector $x_m$ (we adopt our notation for consistency). Their coverage assumption is
\begin{align}
    \lambda_{\min}\left(\sum_{m = 1}^{M^\off} x_m x_m^{\top}\right) \ge M^\off \tau 
    \qquad \text{a.s.},
    \label{eq_coverage_zhang2025}
\end{align}
which implies that $\tau = \Theta(1)$ under our definition of $\tau$.
This assumption is structurally similar to \Cref{assumption_eigen}, and is in fact stronger in two respects. First, it requires an almost sure guarantee, whereas \Cref{assumption_eigen} is stated in a high-probability sense. Second, it captures only the ideal regime in which $\tau = \Theta(1)$ (Case 1 in~\Cref{sec_when_offlinedata_informative}), while in our framework under \Cref{assumption_eigen}, we also allow $\tau$ to decay as $M^\off$ increases (Case 2 in~\Cref{sec_when_offlinedata_informative}).
Although the underlying models and feature spaces differ, both \cref{eq_coverage_zhang2025} and \Cref{assumption_eigen} impose a form of uniform coverage over the feature space during the offline phase.

\subsubsection{Comparison of Regret Bounds} 
\label{discussion_tabularbandit_regret}
We finally compare \Cref{thm1} with the offline--online tabular bandit setting studied in \citet{cheung2024leveraging}. The regret bound in \citet{zhang2025contextual} is not directly comparable, as that work considers a contextual model.

By \Cref{discussion_tabularbandit_envshift}, any offline--online tabular bandit problem with environment shift $\Delta_1$ (see~\cref{eq_tabularbandit}), when embedded into the offline--online linear mixture MDP framework, satisfies
\(
    \Delta(\Delta_1) \le \Delta_2(\Delta_1) \le \sqrt{A}\,\Delta_1,
\)
where the first inequality follows from \Cref{proposition_bandits}, and the second follows from the relation between $\Delta_2(\Delta_1)$ and $\Delta_1$ under the one-hot embedding. Furthermore, due to the discussion in~\Cref{discussion_tabularbandit_coverage}, we have that
\(
\tau \ge \min_{i \in [A]} \frac{N(a_i) + \lambda}{3M^{\off}},
\)
where $N(a_i)$ denotes the number of times arm $a_i$ is pulled in the offline data.  Moreover, under this embedding, only transitions from the initial state $s_1$ are nontrivial; the transitions at $s_2$ and $s_3$ are deterministic and can therefore be ignored. Thus the dimension $d = 3A$ is sufficient.  Applying~\Cref{thm1} with $d = 3A$ then gives
\[
\Regret(K) = \tilde O\left(\min\left\{\frac{A K}{\sqrt{\tau M^{\off}}} + A^2\frac{\Delta_1 K}{\tau}, A\sqrt{K}\right\}\right).
\]
When the offline data are evenly distributed across arms, we have that $\tau = \tfrac{1}{3A} + \frac{\lambda}{3M^{\off}}$, which yields
\[
\Regret(K) = \tilde O\left(\min\left\{\frac{A^{3/2} K}{\sqrt{M^{\off}}} + A^{3}\Delta_1 K, A\sqrt{K}\right\}\right).
\]
This matches the dependence on $K$, $M^{\off}$ and $\Delta_1$ established in~\citet{cheung2024leveraging} (see Theorem 4.9 of~\citet{cheung2024leveraging}).  However, if the offline samples are highly imbalanced across arms, e.g., some arm receives only $O(1)$ pulls as $M^{\off}\to\infty$, then $\tau=O(1/M^{\off})$ and the bound in~\Cref{thm1} reduces to the purely online bound $\tilde{O}(A\sqrt{K})$.  This reflects that our guarantee in~\Cref{thm1} is governed by the worst-covered direction through $\tau$.  In contrast, the guarantees in~\citet{cheung2024leveraging} are expressed in terms of per-arm counts, and may still yield improvements under imbalanced offline data.  These differences arise because our analysis is designed to cover general linear mixture models beyond the tabular setting.

\subsection{Offline--Online Tabular MDPs~\citep{chen2022data}}
\label{discussion_tabular_MDPs}
By \Cref{remark_tabularMDP_equivalence}, any offline--online tabular MDP can be embedded into the offline--online linear mixture MDP framework. Analogous to \Cref{discussion_tabular_bandit}, we examine the relation between \Cref{thm1} and prior work on offline--online tabular MDPs~\citep{chen2022data}, focusing on the respective environment shift assumptions, coverage assumptions, and regret guarantees. Unlike the regret bound in~\citet{chen2022data}, which incurs linear dependence on the environment shift in all regimes, the bound in~\Cref{thm1} avoids this dependence when the offline data are uninformative, typically when the environment shift is large, due to the safeguard structure discussed in~\Cref{sec_when_offlinedata_informative}.

\subsubsection{Comparison of Environment Shift Assumptions}
\label{discusson_delta_in_tabular_MDP}
In this subsection, we relate \Cref{assump_theta_gap} to the environment shift assumption in tabular MDPs proposed by \citet{chen2022data}, which assumes that for all $(s,a)$,
\begin{align}
    \bigl\|P(\cdot\mid s,a) - P^{\off}(\cdot\mid s,a)\bigr\|_1 \le \Delta_3 .
    \label{eq_tabularenvshift}
\end{align}
Using the standard equivalence that represents any tabular MDP as a linear mixture MDP (see \cref{remark_tabularMDP_equivalence}), we show that \cref{eq_tabularenvshift} implies \Cref{assump_theta_gap} with an explicit conversion between the two environment shifts.

\begin{proposition}
\label{prop_tabular_shift_implies_theta_gap}
Consider a tabular offline--online MDP for which \cref{eq_tabularenvshift} holds.  
Let $(P^{\off},r)$ and $(P,r)$ be rewritten as the equivalent offline and online linear mixture MDPs following \cref{remark_tabularMDP_equivalence}, with parameter vectors $\theta^*$ and $\theta^{\off,*}$ of dimension $d$. 
We define $\Delta(\Delta_3)$ as the induced environment shift bound under the corresponding offline--online linear mixture MDP.
Then \Cref{assump_theta_gap} holds with\footnote{\Cref{prop_tabular_shift_implies_theta_gap} shows that $\Delta(\Delta_3) \leq O(\Delta_3)$.  To obtain the reverse direction and conclude that $\Delta(\Delta_3) = \Theta(\Delta_3)$, one can instead define $\Delta(\Delta_3)$ and $\Delta_3$ to measure the actual discrepancy between the offline and online environments, rather than an upper bound on that discrepancy.}
\(
\Delta(\Delta_3) \le \sqrt{SA}\Delta_3
\le \sqrt{d}\Delta_3,
\)
where $S=|\mathcal S|$ and $A=|\mathcal A|$, and $\Delta(\Delta_3)$ is defined analogously to $\Delta_2(\Delta_1)$.
\end{proposition}

\begin{rproof}
Under the equivalence that represents a tabular MDP as a linear mixture MDP (see \cref{remark_tabularMDP_equivalence}), each coordinate of the parameter vector $\theta$ corresponds to a transition probability $P(s' \mid s,a)$.  Consequently, for each $(s,a)$, the vector of $\theta^* - \theta^{\off,*}$ associated with $(s,a)$ is exactly $P(\cdot\mid s,a) - P^{\off}(\cdot\mid s,a)$. Therefore,
\[
\|\theta^* - \theta^{\off,*}\|_2^2
= \sum_{s\in\mathcal S}\sum_{a\in\mathcal A}
\bigl\|P(\cdot\mid s,a) - P^{\off}(\cdot\mid s,a)\bigr\|_2^2.
\]
Using the inequality $\|\cdot\|_2 \le \|\cdot\|_1$ and the tabular environment shift assumption \eqref{eq_tabularenvshift}, we obtain
\(
\|\theta^* - \theta^{\off,*}\|_2^2
\le \sum_{s\in\mathcal S}\sum_{a\in\mathcal A} \Delta_3^2
= SA\Delta_3^2.
\)
Finally, since $d = S^2A$ under the tabular equivalence, we have $\sqrt{SA} \le \sqrt d$, which completes the proof.
\end{rproof}

\begin{remark}
Note that \cref{eq_tabularenvshift} captures only part of the environment shift assumption in \citet{chen2022data} (their Assumption~1). In particular, \citet{chen2022data} also allow discrepancies between $r$ and $r^{\off}$, treating rewards as stochastic and unknown, and consider time-inhomogeneous transition models. Here we restate their assumption in the form of \cref{eq_tabularenvshift} to align with the setting in our main text. Extensions of~\Cref{prop_tabular_shift_implies_theta_gap} to unknown reward functions and time-inhomogeneous transitions are conceptually straightforward and therefore omitted.
\end{remark}

\subsubsection{Comparison of Coverage Assumptions and the Induced $\tau$}
\label{discussion_tabularmdp_coverage}
In~\citet{chen2022data}, their coverage assumption requires that, for every $(h,s,a)$, the number of offline visits equals $M^{\off}/(|\mathcal S||\mathcal A|)$ (see Theorem~2 of~\citet{chen2022data}). Under this uniform allocation, the deterministic construction of $\{\TildeV_{m,h}\}$ in \Cref{sec_comparison} applies directly. The resulting offline design matrix satisfies
\begin{align}
G_{\off} = \operatorname{diag}\Bigl(\frac{H M^{\off}}{|\mathcal S|^2|\mathcal A|}, \ldots, \frac{H M^{\off}}{|\mathcal S|^2|\mathcal A|}\Bigr) \qquad \text{a.s.},
\label{eq_appendix_b22}
\end{align}
so that $\lambda_{\min}(G_{\off})=\frac{H M^{\off}}{|\mathcal S|^2|\mathcal A|}+\lambda$ almost surely.
Consequently, $\tau \ge \frac{H}{|\mathcal S|^2|\mathcal A|}$. In other words, the coverage assumption in \citet{chen2022data} guarantees a nontrivial lower bound on $\tau$ in our framework, namely $\tau = \Theta(1)$. 
Furthermore, similar to the comparison between \Cref{assumption_eigen} and the coverage assumption in~\citet{zhang2025contextual} in the previous section, our framework under \Cref{assumption_eigen} also accommodates the regime in which $\tau$ decays as $M^\off$ increases.

\subsubsection{Comparison of Regret Bounds} 
\label{discussion_tabular_mdp_regret}

Combining the discussions in \Cref{discusson_delta_in_tabular_MDP} and \Cref{discussion_tabularmdp_coverage}, we obtain the following corollary.

\begin{restatable}{corollary}{cortabularMDP}
\label{cor_tabularMDP}
Under the assumptions of~\citet{chen2022data} where the environment shift is bounded by $\Delta_3$ (see~\cref{eq_tabularenvshift}), we have $\tau \ge H/(|\mathcal S|^2|\mathcal A|)$, and the regret bound in~\Cref{thm1} becomes
\begin{align*}
\Regret(K) = \tilde O \Biggl( \min\Biggl\{|\mathcal S|^3 |\mathcal A|^{3/2} K \sqrt{\frac{1}{M^{\off}}} + |\mathcal S|^6 |\mathcal A|^{3} \Delta_3 K, |\mathcal S|^2 |\mathcal A| \sqrt{K} \Biggr\}\Biggr),
\end{align*}
where the dependence on $H$ and $B$ is omitted.
\end{restatable}

\begin{rproofof}{\Cref{cor_tabularMDP}}\label{proof_tabular_MDP}
Note that $\tau \ge \frac{H}{|\mathcal S|^2|\mathcal A|}$ due to~\Cref{discussion_tabularmdp_coverage}. Substituting $d = |\mathcal S|^2 |\mathcal A|$ (see~\Cref{remark_tabularMDP_equivalence}) and $\Delta(\Delta_3) \le \sqrt{d}\Delta_3$ into the regret bound in~\Cref{thm1} directly yields
\[
\Regret(K) =
\tilde O\left(\min\left\{|\mathcal S|^3 |\mathcal A|^{3/2} K \sqrt{\frac{1}{M^{\off}}} + |\mathcal S|^6 |\mathcal A|^{3} \Delta_3 K, |\mathcal S|^2 |\mathcal A| \sqrt{K}\right\}\right).
\]
\end{rproofof}

In comparison, the regret bound of~\citet{chen2022data} scales as
$
|\mathcal S||\mathcal A|^{1/2}\frac{K}{\sqrt{M^{\off}}} + \Delta_3 K,
$
again omitting the dependence on $H$ (see~\Cref{discussion_regret_chen2022}). When $M^{\off}/K$ is sufficiently large, the two bounds agree in their dependence on $K$, $\Delta_3$, and $M^{\off}$. A structural difference is that our bound has a ``min-of-two'' structure, as discussed in~\Cref{sec_when_offlinedata_informative}, thereby retaining a safeguard in regimes where the offline data are uninformative. In contrast, the bound of \citet{chen2022data} always incurs linear dependence on $\Delta_3$, regardless of the quality of the offline data. 

\subsubsection{Regret Bound for Offline--Online Tabular MDPs in~\citet{chen2022data}} 
\label{discussion_regret_chen2022}
We briefly discuss how the regret bound of~\citet{chen2022data} leads to the scaling
\(
|\mathcal S||\mathcal A|^{1/2}\frac{K}{\sqrt{M^{\off}}} + \Delta_3 K,
\)
which we use for comparison in~\Cref{discussion_tabular_mdp_regret}.

Although~\citet{chen2022data} do not explicitly state a closed-form regret bound, their Theorem~2 provides a bound expressed in terms of confidence radii that depend on the environment shift. In their notation, three confidence radii are defined:
\(
\epsilon_{R}^{\mathrm{DP}}(n, N),
\epsilon_{P}^{\mathrm{DP}}(n, N),
\epsilon_{V}^{\mathrm{DP}}(n, N),
\)
where \(n \in \{0,\ldots, K/(|\mathcal S||\mathcal A|)\}\) and
\(N = M^{\off}/(|\mathcal S||\mathcal A|)\).
These radii are given by
\(
\epsilon_{R}^{\mathrm{DP}}(n, N) 
=
\sqrt{\frac{\mu^2}{2n} + \frac{(1-\mu)^2}{2N}} + \Delta_3(1-\mu), 
\epsilon_{P}^{\mathrm{DP}}(n, N)
=
\sqrt{2|\mathcal S|\left(\frac{\mu^2}{2n}+\frac{(1-\mu)^2}{2N}\right)}+ \Delta_3(1-\mu), 
\epsilon_{V}^{\mathrm{DP}}(n, N)
=
H\epsilon_{R}^{\mathrm{DP}}(n, N),
\)
where \(\mu \in [0,1]\) is a weighting parameter (see Section~4.2 of~\citet{chen2022data} for details).

As noted in~\citet{chen2022data}, the function
\(
\sqrt{ \frac{\mu^2}{2n} + \frac{(1-\mu)^2}{2N}}+ \Delta_3(1-\mu)
\)
is convex in \(\mu\). Therefore, by setting \(\mu = 0\) and applying Equation~(EC.11) of~\citet{chen2022data}, the regret satisfies
\[
\Regret(K) 
\le
|\mathcal S||\mathcal A| 
\sum_{n=0}^{K/(|\mathcal S||\mathcal A|)}
\Bigl(
\epsilon_{R}^{\mathrm{DP}}(n,N)
+
\epsilon_{P}^{\mathrm{DP}}(n,N)
+
\epsilon_{V}^{\mathrm{DP}}(n,N)
\Bigr).
\]
Substituting the above expressions and summing over \(n\) yields
\[
\Regret(K) = \tilde O\left(|\mathcal S| |\mathcal A|^{1/2} \sqrt{K^2/M^{\off}} + \Delta_3 K \right),
\]
where we omit logarithmic factors and constants, as well as the dependence on \(H\). We emphasize that setting $\mu = 0$ and exploiting convexity does not materially loosen the bound derived in~\citet{chen2022data}. As long as the optimal choice of $\mu$ satisfies $\mu<1$ for all $n$ (which holds in most regimes of interest), the gap between the tightest bound obtainable from~\citet{chen2022data} and the simplified bound used in our discussion is at most a constant factor.

\subsection{Offline Linear Mixture MDPs~\citep{nguyen2023instance}}
\label{discussion_offlineRL}
We finally compare \Cref{thm1} with the suboptimality guarantees for offline linear mixture MDPs studied in \citet{nguyen2023instance}. When $K = 1$ and $\Delta = 0$, our regret bound reduces to a purely offline suboptimality bound; see \Cref{thm1_in_offline_RL} for details. Since this regime involves no environment shift, we focus on comparing the respective coverage assumptions and suboptimality gap guarantees in \Cref{thm1} and~\citet{nguyen2023instance}. As in \Cref{sec_comparison}, for clarity of exposition, we assume in this section that the state and action spaces are finite, so all probability measures are treated as probability mass functions.

\subsubsection{Comparison of Coverage Assumptions and the Induced $\tau$}
\label{discussion_offlineRL_coverage}
Although the primary contribution of \citet{nguyen2023instance} is to establish suboptimality guarantees under a partial coverage assumption (Assumptions 4.1 and 4.2 in~\citet{nguyen2023instance}), the uniform data coverage assumption, \Cref{assumption_data_coverage_main}, as discussed in~\citet{nguyen2023instance}, is also sufficient to obtain their suboptimality results for linear mixture MDPs (Theorem 7 and Theorem 9 in~\citet{nguyen2023instance}). Under the stochastic construction of $\{\TildeV_{m,h}\}$ in \Cref{def_construction_tildeV} and the uniform learnability assumption (\Cref{assumption_offlineRL_span_main}), the uniform data coverage condition in \Cref{assumption_data_coverage_main} implies that $\tau = \Theta(1)$.
This argument is formalized in \Cref{cor_uniformdata_coverage} in \Cref{sec_comparison}.
We defer the discussion of the partial coverage assumptions to~\Cref{discussion_coverage_2} and show that these assumptions are not applicable for the offline--online learning setting considered here.

\subsubsection{Comparison of Suboptimality Gap} 
With discussions in~\Cref{discussion_offlineRL_coverage}, we directly obtain the following corollary.

\begin{corollary}
\label{cor_offlineRL}
Under \Cref{def_construction_tildeV}, \Cref{assumption_offlineRL_span_main} and~\Cref{assumption_data_coverage_main}, when $K = 1$ and $\Delta = 0$, we have $\tau = \Theta(1)$, and the regret bound in~\Cref{thm1}
reduces to an offline suboptimality gap of order
$\tilde O(\frac{1}{\sqrt{M^{\off}}})$.
\end{corollary}
\Cref{cor_offlineRL} follows directly from combining~\Cref{cor_uniformdata_coverage} with~\Cref{thm1}. 
The $\tilde O(\frac{1}{\sqrt{M^{\off}}})$ rate matches the rate obtained in Theorems~7 and Theorem~9 of~\citet{nguyen2023instance}, up to logarithmic
factors, and is statistically optimal.
This shows that, in the purely offline regime, our bound is sharp.


\section{Omitted Discussion from \cref{sec:preliminary} to~\Cref{sec_main_results}}
\label{sec:discussion_preliminary}

In this section, we supplement \Cref{dis_comparison_new} with additional discussions beyond comparisons to existing literature. Specifically, we show (i) the environment shift $\Delta$ and the resulting discrepancy between offline and online value functions are nearly of the same order (\Cref{discussion_theta_valuefunction}); (ii) the necessity of the assumption that $\Delta$ is known in~\Cref{assump_theta_gap} (\Cref{discussion_knowingdelta}); (iii) a comparison of~\Cref{assumption_eigen} and other coverage assumptions in the literature (\Cref{discussion_coverage}); (iv) \cref{eq_principle_tildev} is neither convex nor concave in general (\cref{discussion_concavity_tau}); and (v) the technical details omitted in translating~\Cref{thm1} into a suboptimality gap bound for comparison with~\citet{nguyen2023instance} in~\Cref{discussion_offlineRL} (\Cref{thm1_in_offline_RL}).

\subsection{Environment Shift in terms of Value Functions}
\label{discussion_theta_valuefunction}

We start off by studying \Cref{assump_theta_gap} from a value function perspective. Although one could alternatively define the environment shift $\Delta$ directly in terms of value function discrepancies between the offline and online MDPs, we show that \Cref{assump_theta_gap} already implies an explicit bound on the difference between value functions induced by the same policy in the offline and online environments. Moreover, there exist instances in which $\Delta$ exactly matches this value function gap. We therefore retain the parameter-space definition of $\Delta$ in~\Cref{assump_theta_gap}, which more naturally reflects the model geometry and supports the regret analysis.

Denote $V_h^{\off,\pi}$ as the value function at stage $h$ under policy $\pi$ in the offline MDP, defined analogously to $V_h^{\pi}$ in the online MDP in \cref{eq_vdefinition}.  We start by showing that when the reward functions $r$ and $r^{\off}$ are identical, the value difference at the initial state is upper bounded at the same order of $\|\theta^* - \theta^{\off,*}\|_2$, up to a polynomial dependence on the horizon $H$ and the feature dimension $d$.

\begin{proposition}
\label{prop_upper_theta_gap}
Assume that the reward functions are identical, i.e., $r = r^{\off}$.  Under~\Cref{def_mdp_1} and \Cref{assump_theta_gap}, for any fixed policy $\pi$, we have
\(
\abs{V_1^{\pi}(s_1) - V_1^{\off,\pi}(s_1)} \leq \frac{H(H-1)}{2}\Delta \sqrt{d}.
\)
\end{proposition}

\begin{rproof}
Fix an arbitrary policy $\pi$. For any stage $h$, since the reward functions are identical, the value functions satisfy the Bellman recursions
\[
V_h^{\pi}(s) = r(s,\pi_h(s)) + \sum_{s'} P(s' \mid s,\pi_h(s)) V_{h+1}^{\pi}(s'),
\ 
V_h^{\off,\pi}(s) = r(s,\pi_h(s)) + \sum_{s'} P^{\off}(s' \mid s,\pi_h(s)) V_{h+1}^{\off,\pi}(s'),
\]
with terminal condition $V_{H+1}^{\pi} \equiv V_{H+1}^{\off,\pi} \equiv 0$.
Subtracting the two equations and adding and subtracting
$\sum_{s'} P^{\off}(s' \mid s,\pi_h(s)) V_{h+1}^{\pi}(s')$,
we obtain
\begin{align*}
V_h^{\pi}(s) - V_h^{\off,\pi}(s) = \sum_{s'} \bigl(P - P^{\off}\bigr)\bigl(s' \mid s,\pi_h(s)\bigr) V_{h+1}^{\pi}(s') + \sum_{s'} P^{\off}\bigl(s' \mid s,\pi_h(s)\bigr)
\bigl( V_{h+1}^{\pi}(s') - V_{h+1}^{\off,\pi}(s') \bigr).
\end{align*}

We bound the two terms separately.  For the first term, by the linear mixture structure,
\[
\sum_{s'} \bigl(P - P^{\off}\bigr)\bigl(s' \mid s,a\bigr) V_{h+1}^{\pi}(s')
= \sum_{j=1}^d \bigl(\theta_j^* - \theta_j^{\off,*}\bigr)
\sum_{s'} \phi_j(s' \mid s,a) V_{h+1}^{\pi}(s').
\]
Since $0 \le V_{h+1}^{\pi}(s') \le H-h$ for all $s'$ and 
$\sum_{s'} |\phi_j(s' \mid s,a)| \le 1$,
we have
\(
\left| \sum_{s'} \phi_j(s' \mid s,a) V_{h+1}^{\pi}(s') \right| \leq H-h.
\)
Applying Cauchy--Schwarz yields
\[
\left|
\sum_{s'} \bigl(P - P^{\off}\bigr)\bigl(s' \mid s,a\bigr) V_{h+1}^{\pi}(s')
\right|
\le \|\theta^* - \theta^{\off,*}\|_2 \cdot \sqrt{d}(H-h)
\le \Delta \sqrt{d}(H-h).
\]
For the second term, since $P^{\off}(\cdot \mid s,a)$ is a probability distribution,
\[
\left|
\sum_{s'} P^{\off}\bigl(s' \mid s,a\bigr)
\bigl( V_{h+1}^{\pi}(s') - V_{h+1}^{\off,\pi}(s') \bigr)
\right|
\le \sup_{s'} \bigl| V_{h+1}^{\pi}(s') - V_{h+1}^{\off,\pi}(s') \bigr|.
\]

Define
\(
D_h := \sup_{s} \bigl| V_h^{\pi}(s) - V_h^{\off,\pi}(s) \bigr|.
\)
Combining the above bounds gives the recursion
\[
D_h \le \Delta \sqrt{d}(H-h) + D_{h+1},
\qquad D_{H+1} = 0.
\]
Unrolling the recursion yields
\(
D_1 \le \Delta \sqrt{d} \sum_{h=1}^H (H-h)
= \Delta \sqrt{d} \cdot \frac{H(H-1)}{2}.
\)
Since
$\bigl| V_1^{\pi}(s_1) - V_1^{\off,\pi}(s_1) \bigr| \le D_1$,
the proof is complete.
\end{rproof}

We now show that the above relationship is nearly tight in a worst-case sense. Specifically, there exist instances for which the discrepancy between the offline and online value functions scales linearly with the parameter shift $\|\theta^* - \theta^{\off,*}\|_2$.

\begin{proposition}
\label{prop_lower_theta_gap}
For any $\Delta>0$, there exist parameters $\theta^*$ and $\theta^{\off,*}$
with $\|\theta^* - \theta^{\off,*}\|_2 = \Delta$ such that
\(
\sup_{\pi}
\bigl| V_1^{\pi}(s_1) - V_1^{\off,\pi}(s_1) \bigr|
= \Omega(\Delta).
\)
\end{proposition}

\begin{rproof}
Consider a linear bandit instance, which can be equivalently represented as a linear mixture MDP via the construction in \Cref{discussion_tabularbandit_envshift}. Under this construction, the parameter vectors of the induced linear mixture MDP coincide exactly with the bandit parameters, and the induced environment shift satisfies $\Delta = \|\theta^* - \theta^{\off,*}\|_2$. Moreover, for any policy $\pi \in \mathbb{R}^d$ (which is valid in the linear bandit setting), the corresponding value functions satisfy
\(
V_1^{\pi}(s_1) - V_1^{\off,\pi}(s_1)
= \langle \theta^* - \theta^{\off,*}, \pi \rangle .
\)
Taking $\pi = \frac{\theta^* - \theta^{\off,*}}{\|\theta^* - \theta^{\off,*}\|_2}$ yields
\(
\sup_{\pi} \bigl| V_1^{\pi}(s_1) - V_1^{\off,\pi}(s_1) \bigr|
= \|\theta^* - \theta^{\off,*}\|_2
= \Delta,
\)
which completes the proof.
\end{rproof}

\subsection{Impossibility of Unknown $\Delta$}
\label{discussion_knowingdelta}

We briefly discuss the offline--online linear mixture MDP setting in which no explicit upper bound on $\|\theta^* - \theta^{\off,*}\|_2$ (or any equivalent measure of environment shift) is given, i.e., $\Delta$ is unknown. The main takeaway is that, without knowing $\Delta$, any algorithm that achieves a regret rate strictly better than $\sqrt{K}$ in the no-shift case $\theta^*=\theta^{\off,*}$ must necessarily suffer a $\sqrt{K}$-order regret on some shifted instance. Thus, knowledge of an upper bound on $\Delta$ is essential for obtaining a worst-case guarantee that uniformly improves upon the online-only rate while remaining strictly better in the no-shift regime.

\begin{proposition}
\label{impossibility_delta}
Consider the class of offline--online linear mixture MDP instances for which all assumptions in the main text hold except \Cref{assump_theta_gap}, i.e., the environment shift is not assumed to be bounded by a known constant. Suppose an algorithm satisfies 
\(
\Regret(K)=o(\sqrt{K})
\)
over the instance class when $\theta^*=\theta^{\off,*}$, then there exists another instance in the same class with $\theta^*\neq\theta^{\off,*}$ for which the algorithm incurs regret
\(
\Regret(K)>\Omega(\sqrt{K}).
\)
\end{proposition}

We omit the proof of~\Cref{impossibility_delta}, as \citet{cheung2024leveraging} provides an explicit hard instance construction in the tabular bandit setting (see Proposition 3.1 in~\citet{cheung2024leveraging}).  By the equivalence discussed in \Cref{discussion_tabularbandit_envshift}, such instances can be embedded as a special case of offline--online linear mixture MDPs. 

\subsection{Comparison of~\Cref{assumption_eigen} and~\Cref{assumption_data_coverage_main} with Other Coverage Assumptions}
\label{discussion_coverage}

\subsubsection{Comparison with Feature Coverage Assumptions}
\label{discussion_coverage_1}

Beyond the uniform data coverage assumption (\Cref{assumption_data_coverage_main}) discussed in~\Cref{sec_comparison}, 
feature coverage assumptions, often formulated in a form similar to 
\Cref{assumption_eigen}, are also widely adopted in offline RL models 
with linear structure. In this subsection, we compare \Cref{assumption_eigen} with other feature coverage assumptions used in related models in the literature.

\paragraph{Comparison with~\citet{yin2022near}.} 
Although~\citet{yin2022near} study linear MDPs, whose feature space differs from that of linear mixture MDPs, the underlying coverage assumption is essentially analogous to~\Cref{assumption_eigen}.  
Specifically, they assume that, with high probability,
\(
    \lambda_{\min}\left(\sum_{m=1}^{M^{\off}}\phi_m \phi_m^\top\right) \ge \tau M^{\off},
\)
where $\phi_m$ denotes the feature vector associated with the $m$-th offline sample (see Assumption~2.2 and Lemma~H.5 in~\citet{yin2022near}).  
This condition ensures that the empirical design matrix grows linearly in the number of offline samples in every direction in the feature space.

In our linear mixture MDP setting, the role of $\phi_m$ is played by the induced feature $\phi(\cdot \mid s,a)\TildeV$.  
While linear MDPs do not require the design of $\{\TildeV_{m,h}\}$ due to their structural properties, both~\citet{yin2022near} and~\Cref{assumption_eigen} formulate coverage in terms of a lower bound on the minimum eigenvalue of an empirical design matrix. 
However, the scope of \Cref{assumption_eigen} is strictly broader.  
The assumption in~\citet{yin2022near} enforces a non-trivial bound $\tau = \Theta(1)$, corresponding to the ideal case discussed in~\Cref{sec_when_offlinedata_informative}.
In contrast, \Cref{assumption_eigen} accommodates more general regimes in which $\tau$ may decay as $M^\off$ increases (Case 2 in~\Cref{sec_when_offlinedata_informative}).


\subsubsection{Comparison with Partial Data Coverage Assumptions}
\label{discussion_coverage_2}
Unlike in offline RL, canonical partial data coverage assumptions may not be directly applicable in the offline--online setting due to the presence of environment shift. In this subsection we illustrate this issue using the partial data coverage assumption in~\citet{nguyen2021sample} (Assumption~4.1 therein).

Concretely, the partial data coverage assumption in~\citet{nguyen2021sample} states that for any $(h,s,a)$ tuple,
\(
    d_h^{\pi^{\off, *}}(s,a) > 0 \Rightarrow d_h^{\pi^{\off}}(s,a) > 0,
\)
where $\pi^{\off,*}$ denotes an optimal policy of the offline MDP, and $d_h^{\pi^{\off,*}}$ denotes the marginal state--action probability measure under $\pi^{\off,*}$, defined analogously to $d_h^{\pi^{\off}}$ in~\Cref{sec_comparison}.

We now present a simple counterexample showing that, for any $M^\off, K$, and $\Delta > 0$, this assumption is not sufficient to guarantee that the offline data is informative according to \Cref{def_informative}.
Consider an offline--online tabular bandit with three arms.
Suppose the mean rewards of the arms satisfy
\[
    \mu(1) = 0.5, \ \mu^{\off}(1) = 0.5,\ 
    \mu(2) = 0.5 - \frac{1}{\sqrt{K}}, \ \mu^{\off}(2) = 0.5 - \frac{1}{\sqrt{K}},
    \mu(3) = 0.5 - \frac{1}{4}\Delta, \ \mu^{\off}(3) = 0.5 + \frac{1}{4}\Delta,
\]
where $\mu$ and $\mu^{\off}$ follow the definitions in \Cref{discussion_tabular_bandit}. As discussed in \Cref{proposition_bandits}, this instance satisfies \Cref{assump_theta_gap}.

In the offline environment, arm $3$ is optimal.
Suppose the offline behavior policy always pulls arm $3$.
Then the partial coverage assumption above is satisfied, since the optimal offline policy also selects arm $3$.
However, in the online environment arm $1$ is optimal.
Since the offline data contains no information about arm $1$ and arm $2$, the learner must explore arm $2$ during the online phase.
Consequently, the regret remains of order $\sqrt{K}$ regardless of the value of $M^\off$, and the offline data is therefore not informative. 
This completes the argument, since under the uniform data coverage assumption (\Cref{assumption_data_coverage_main}), the offline data can be informative for appropriate values of $M^\off$, $K$, and $\Delta > 0$, as discussed in~\Cref{discussion_tabular_bandit}.

If the partial coverage assumption were instead defined with respect to the optimal policy of the online environment, one might obtain positive guarantees when the offline and online MDPs share the same state and action spaces. However, in general offline--online settings the two environments may differ substantially, and such a definition may not even be well-defined.
We leave the investigation of such coverage conditions for future work.

\subsection{Discussion on the Convexity of~\cref{eq_principle_tildev}}
\label{discussion_concavity_tau}

In this section we provide a counterexample showing that, in general,
\cref{eq_principle_tildev} is neither concave nor convex with respect to $V$. Consequently, one cannot expect a closed-form solution or to leverage convex optimization algorithms to solve it in general.

For the counterexample, we consider a bandit embedded linear mixture MDP with $|\A| = 1$. Following the discussion in~\Cref{discussion_tabular_bandit}, the corresponding offline--online linear mixture MDP has $|\A|=1$, $|\S|=2$, and $H=2$. Choosing $d=2$ is sufficient as discussed in~\Cref{discussion_tabular_bandit}. The feature mapping satisfies $\phi(\cdot\mid s,a)=I\in\mathbb{R}^{2\times 2}$ for any $(s,a)$.

We show that for the adaptively obtained real symmetric PSD matrix $A_{m,h}$, which corresponds to
$$
    \sum_{(m',h') < (m,h)} x^{\off}_{m',h'}(\TildeV_{m',h'+1})\bigl(x^{\off}_{m',h'}(\TildeV_{m',h'+1})\bigr)^{\top}
$$
in~\cref{eq_principle_tildev}, the problem 
$$
    \min_{V \in [-1,1]^{|\S|}} \lambda_{\min}(A_{m,h} + \phi(\cdot|s_{m,h},a_{m,h})V(\phi(\cdot|s_{m,h},a_{m,h})V)^{\top}) =  \min_{V \in [-1,1]^{|\S|}} \lambda_{\min}(A_{m,h} + VV^{\top})
$$
is neither concave nor convex with respect to $V$.

We first consider the step $(m,h)=(1,1)$. In this case the matrix in \cref{eq_principle_tildev} satisfies $A_{1,1}=H^2 d I_2$ following~\Cref{thm1}.
For any $V \in \mathbb{R}^2$, the matrix $VV^\top$ has rank at most one.
Therefore no choice of $V$ can increase the minimum eigenvalue of $A_{1,1} + VV^{\top}$.
Thus it's valid to choose 
\(
V=(1,0)^\top.
\)

After this step, the matrix becomes
\[
A_{2,1} = A_{1,1}+VV^\top =
\begin{pmatrix}
9 & 0\\
0 & 8
\end{pmatrix}.
\]
Since we consider a bandit problem, the horizon index can be omitted in the following discussion.

Define
$f(V)=\lambda_{\min}\bigl(A_{2,1}+VV^\top\bigr)$ and $V\in[-1,1]^2$.
We first show that $f(V)$ is not convex. Let
\[
V_1=
\begin{pmatrix}
1\\
1
\end{pmatrix},
\quad
V_2=
\begin{pmatrix}
-1\\
1
\end{pmatrix},
\quad
\bar V=\frac{1}{2}(V_1+V_2)=
\begin{pmatrix}
0\\
1
\end{pmatrix}.
\]
Then
\[
A_{2,1}+V_1V_1^\top=
\begin{pmatrix}
10 & 1\\
1 & 9
\end{pmatrix},
\qquad
A_{2,1}+V_2V_2^\top=
\begin{pmatrix}
10 & -1\\
-1 & 9
\end{pmatrix}.
\]
Both matrices have smallest eigenvalue
\(
f(V_1)=f(V_2)=\frac{19-\sqrt5}{2}.
\)
On the other hand,
\[
A_{2,1}+\bar V\bar V^\top=
\begin{pmatrix}
9 & 0\\
0 & 9
\end{pmatrix},
\qquad
f(\bar V)=9.
\]
Hence
\(
f(\bar V) > \frac{f(V_1)+f(V_2)}{2},
\)
which violates convexity.

Next we show that $f(V)$ is not concave. Let
\[
W_1=
\begin{pmatrix}
0\\
1
\end{pmatrix},
\quad
W_2=
\begin{pmatrix}
0\\
-1
\end{pmatrix},
\quad
\bar W=\frac{1}{2}(W_1+W_2)=
\begin{pmatrix}
0\\
0
\end{pmatrix}.
\]
Then
\[
A_{2,1}+W_1W_1^\top=
\begin{pmatrix}
9 & 0\\
0 & 9
\end{pmatrix},
\qquad
A_{2,1}+W_2W_2^\top=
\begin{pmatrix}
9 & 0\\
0 & 9
\end{pmatrix}.
\]
Hence
\(
f(W_1)=f(W_2)=9.
\)
On the other hand,
\[
A_{2,1}+\bar W\bar W^\top=A_{2,1}=
\begin{pmatrix}
9 & 0\\
0 & 8
\end{pmatrix},
\qquad
f(\bar W)=8.
\]
Therefore
\(
f(\bar W)
<
\frac{f(W_1)+f(W_2)}{2},
\)
which shows that $f(V)$ is not concave.

\subsection{Specializing \Cref{thm1} to the Purely Offline Regime}
\label{thm1_in_offline_RL}
It is not fully precise to directly conclude from the statement of \Cref{thm1} that the suboptimality gap is of order $1/\sqrt{M^{\off}}$ by simply setting $K=1$ and $\Delta=0$. Indeed, the proof of~\Cref{thm1} implicitly operates in a regime where $K$ is sufficiently large and several constant terms are absorbed into the $\tilde O(\cdot)$ notation (see~\cref{eq_final}).

To make the comparison rigorous, we return to the explicit regret upper bound for any $K$ and $M^\off$
in~\cref{eq_final}:
\begin{align*}
\Regret(K) \le \min\Biggl\{
&
\Ex \Biggl[2\sqrt{2H^2K} \Biggl[H \sqrt{2 \left(\log \left(\frac{1}{\delta}\right)+ \frac{1}{2} d \log\left(1 + \frac{\lambda_{\max}(G_{\off})}{H^2 d}+ \frac{HK}{d}\right)\right)} + H \sqrt{d} B \nonumber
\\
&\qquad
+ \Delta \frac{\lambda_{\max}(G_{\off})}{\sqrt{\tau(\delta')M^{\off}}}
\Biggr] \times
\sqrt{d \log\left(1 + \frac{H^3K}{\tau(\delta')M^\off}\right)} \Bigm| \E_2\Biggr], \nonumber
\\
& \Ex \Biggl[2\sqrt{2HT} \left(H \sqrt{2 \left(\log \left(\frac{1}{\delta}\right) + \frac{d}{2} \log\left(1 + \frac{HK}{d}\right)\right)} + H \sqrt{d} B \right) \nonumber
\\
&\qquad\times
\sqrt{d \log\left(1 + \frac{HK}{d}\right)} \Bigm | \E_2 \Biggr]
\Biggr\}
+ H^2 K (2\delta + \delta') + H K (2\delta + \delta').
\end{align*}
Setting $K=1$ and $\Delta=0$, and choosing $\delta=\delta' = 1/M^{\off}$, due to $\lambda_{\max}(G_{\off}) \le H^3dM^\off$, we obtain
\begin{align*}
\Regret(K) = \tilde O\Biggl(BH^2d \sqrt{\log\left(1 + \frac{H^3}{\tau(\delta')M^{\off}}\right)}\Biggr).
\end{align*}
In particular, under the uniform coverage condition of \Cref{cor_uniformdata_coverage}, where $\tau(\delta') = \Theta(1)$, this bound simplifies to $\tilde O\left(\frac{1}{\sqrt{M^{\off}}}\right)$, which is directly comparable to the suboptimality gap established in Theorems~7 and~9 of~\citet{nguyen2023instance}.


\section{Proofs}\label{sec_appendix_proof}


\subsection{Proofs in~\Cref{sec_upper_bound}}\label{appendix_proofs}

This section provides a complete proof of~\Cref{thm1}.

\RegretBound*

Before presenting the proof of~\Cref{thm1}, we introduce a filtration that will play a central role in the analysis. The filtration jointly accounts for the offline and online phases.  To this end, we introduce a unified indexing scheme that orders all samples from both phases. Specifically, consider the collection of offline indices
\(
\mathcal I_{\off} \coloneqq \{(m,h): m=1,\ldots,M^{\off}; h=1,\ldots,H\},
\)
and the collection of online indices
\(
\mathcal I_{\on} \coloneqq \{(k,h): k=1,\ldots,K; h=1,\ldots,H\}.
\)
We consider the unique one-one mapping
\(
t \longleftrightarrow (i_t,h_t) \in \mathcal I_{\off}\cup\mathcal I_{\on}
\)
that orders all samples, such that all offline indices in $\mathcal I_{\off}$ precede those in $\mathcal I_{\on}$. 
Note that the mapping is unique as (i) within each trajectory, samples are indexed by $h = 1, \ldots, H$; (ii) online trajectories are indexed by $k = 1, \ldots, K$; and (iii) offline trajectories are indexed by $m = 1, \ldots, M^{\off}$, which induces a unique order due to the adaptive behavior policy.

For each time index $t$, we define the filtration $\mathcal F_t$ as follows.

\begin{itemize}
\item If $t$ corresponds to an offline index $(m,h)\in\mathcal I_{\off}$, then $\mathcal F_t$ is the $\sigma$-algebra generated by the offline observations up to and including $(m,h)$, that is,
\[
\mathcal F_t = \sigma\bigl(\{(S^{\off}_{m',h'},A^{\off}_{m',h'},r^{\off}(S^{\off}_{m',h'},A^{\off}_{m',h'})): m'<m \text{ or } (m'=m,h'\le h)\}\bigr).
\]

\item If $t$ corresponds to an online index $(k,h)\in\mathcal I_{\on}$, then $\mathcal F_t$ is the $\sigma$-algebra generated by the entire offline dataset together with the online observations up to and including stage $h$ of episode $k$, namely
\[
\mathcal F_t = \sigma\bigl(\{\mathcal H^{\off}_m\}_{m=1}^{M^{\off}}, \{(s^{k'}_{h'},a^{k'}_{h'},r(s^{k'}_{h'},a^{k'}_{h'})):k'<k \text{ or } (k'=k,h'\le h)\}\bigr).
\]
\end{itemize}
We use $\mathcal F_t$ and $\mathcal F_{m,h}$ (or $\mathcal F_{k,h}$) interchangeably when there is no ambiguity, since the indices are in one-to-one correspondence.

To prove~\Cref{thm1}, we first establish a key concentration result for the value-targeted regression introduced in~\Cref{sec_algo}. Specifically, we rely on a self-normalized concentration inequality adapted from~\citet{abbasi2011improved}.

\begin{lemma}
\label{lemma_biased_ellipsoid}
For any $\delta \in (0,1)$, with probability at least $1-2\delta$, for all $k \ge 1$,
\[
\bigl\|\thetaall_{k} - \theta^*\bigr\|_{M^{\all}_{k}} \le \gamma_k, 
\qquad 
\bigl\|\thetaon_{k} - \theta^*\bigr\|_{M^{\on}_{k}} \le \beta_k,
\]
where
\[
\gamma_k = H \sqrt{2\log\Bigl(\frac{\det(M^{\all}_{k})^{1/2}}{\det(\lambda I)^{1/2}\delta} \Bigr)} + \Delta\frac{\lambda_{\max}(G_{\off})}{\sqrt{\lambda_{\min}(\lambda I + G_{\off})}} + \sqrt{\lambda}B,
\qquad
\beta_k = H \sqrt{2\log\Bigl(\frac{\det(M^{\on}_{k})^{1/2}}{\det(\lambda I)^{1/2}\delta}\Bigr)} + \sqrt{\lambda}B.
\]
Equivalently, define the good event
\(
\E_0 = \bigl\{\theta^* \in \B_k \cap \C_k, \text{ for all } k \ge 1 \bigr\},
\)
where
\(
\B_k = \Bigl\{\theta \in \mathbb{R}^d : \|\theta - \thetaall_k\|_{M^{\all}_k} \le \gamma_k \Bigr\},
\C_k = \Bigl\{\theta \in \mathbb{R}^d : \|\theta - \thetaon_k\|_{M^{\on}_k} \le \beta_k \Bigr\}.
\)
Then the event $\E_0$ holds with probability at least $1-2\delta$.
\end{lemma}

\begin{rproof}
We discuss the offline--online estimator first. Note that $M^{\all}_{k}$ is symmetric and satisfies $M^{\all}_{k}\succeq \lambda I$, hence it is positive definite and invertible. Therefore, using the fact that $M_k^\all = M_k^\on + G_\off$,
\begin{align*}
M^{\all}_{k}&\bigl(\thetaall_{k}-\theta^*\bigr)
= w^{\all}_{k}-M^{\all}_{k}\theta^* 
= \bigl(w^{\all}_{k}-M^{\on}_{k}\theta^{*}\bigr) - G_{\off}\theta^{*} 
= \bigl(w^{\all}_{k}-M^{\on}_{k}\theta^{*}\bigr) - G_{\off}\theta^{\off, *} + (G_{\off}\theta^{\off, *} - G_{\off}\theta^{*}) \\
&= \sum_{k'=1}^{k-1}\sum_{h=1}^{H}
x^{\on}_{k',h}\Bigl(y^{\on}_{k',h}-\langle x^{\on}_{k',h},\theta^*\rangle\Bigr) -
\lambda\theta^* + \sum_{m=1}^{M^{\off}}\sum_{h=1}^{H} x^{\off}_{m,h}\Bigl(y^{\off}_{m,h}-\langle x^{\off}_{m,h},\theta^{\off, *}\rangle\Bigr) \\
&\qquad + \sum_{m=1}^{M^{\off}}\sum_{h=1}^{H} x^{\off}_{m,h}\bigl(x^{\off}_{m,h}\bigr)^\top(\theta^{\off, *}-\theta^*) .
\end{align*}

Define
\(
\eta^{\on}_{k',h} = y^{\on}_{k',h}-\langle x^{\on}_{k',h},\theta^*\rangle,
\eta^{\off}_{m,h} = y^{\off}_{m,h}-\langle x^{\off}_{m,h},\theta^{\off, *}\rangle, 
b^{\off} = \sum_{m=1}^{M^{\off}} \sum_{h=1}^{H} x^{\off}_{m,h}\bigl(x^{\off}_{m,h}\bigr)^{\top} (\theta^{\off, *} - \theta^*),
\)
and let
\(
S_k = \sum_{m=1}^{M^{\off}}\sum_{h=1}^{H} x^{\off}_{m,h}\eta^{\off}_{m,h} + \sum_{k'=1}^{k-1}\sum_{h=1}^{H} x^{\on}_{k',h}\eta^{\on}_{k',h}.
\)
Here both $\{\eta^{\on}_{k',h}\}$ and $\{\eta^{\off}_{m,h}\}$ form martingale difference sequences with respect to the filtration $\{\F_t\}$, and $S_k$ collects the corresponding self-normalized noise terms that can be controlled via a self-normalized concentration inequality. Then the decomposition becomes
\(
M^{\all}_{k}\bigl(\thetaall_{k}-\theta^*\bigr) = S_k + b^{\off} - \lambda\theta^*.
\)
For the online-only estimator, following the same approach, we have that
$$
    M^{\on}_{k}\bigl(\thetaon_{k}-\theta^*\bigr) = \sum_{k'=1}^{k-1}\sum_{h=1}^{H} x^{\on}_{k',h}\eta^{\on}_{k',h} - \lambda \theta^*.
$$
Since $0\le \Vhat_{k', h},\TildeV_{m, h}\le H$, for any $h,k',m$, both $\eta^{\on}_{k',h}$ and $\eta^{\off}_{m,h}$ are uniformly bounded almost surely:
\[
|\eta^{\on}_{k',h}|\le H,
\qquad
|\eta^{\off}_{m,h}|\le H
\quad\text{a.s.}.
\]

Recall the definition of the filtration $\F_{k,h}$ and $\F_{m,h}$, we have that 
\(
\Ex\left[\eta^{\on}_{k',h}\mid \mathcal{F}_{k',h}\right]=0,
\Ex\left[\eta^{\off}_{m,h}\mid \mathcal{F}_{m,h}\right]=0,
\)
for all $k',h,m$. 
Then all requirements of~\cref{lemma_concentration} (see \cref{sec:useful_facts}) are satisfied. By applying~\cref{lemma_concentration} to the offline--online estimator, we have that with probability at least $1-\delta$, for all $k\ge 1$,
\[
\left\|S_k\right\|_{(M^{\all}_{k})^{-1}} \le H\sqrt{2\log\Bigl(\frac{\det(M^{\all}_{k})^{1/2}}{\det(\lambda I)^{1/2}\delta} \Bigr)}.
\]
Similarly, by applying~\cref{lemma_concentration} to the online-only estimator, we have that with probability at least $1 - \delta$,
$$
    \left\|\sum_{k'=1}^{k-1}\sum_{h=1}^{H} x^{\on}_{k',h}\eta^{\on}_{k',h}\right\|_{(M_k^{\on})^{-1}} \le H\sqrt{ 2\log\Bigl(\frac{\det(M^{\on}_{k})^{1/2}}{\det(\lambda I)^{1/2}\delta}\Bigr)}.
$$

Since $M^{\all}_{k}$ and $M^{\on}_{k}$ are positive definite and symmetric, $\|\cdot\|_{M^{\all}_{k}}$ and $\|\cdot\|_{M^{\on}_{k}}$ are well-defined norms. Thus by the triangle inequality,
\[
\|\thetaall_{k}-\theta^*\|_{M^{\all}_{k}} 
= 
\|M^{\all}_{k}(\thetaall_{k}-\theta^*)\|_{(M^{\all}_{k})^{-1}}
\le
\|S_k\|_{(M^{\all}_{k})^{-1}} + \|b^{\off}\|_{(M^{\all}_{k})^{-1}} + \|\lambda\theta^*\|_{(M^{\all}_{k})^{-1}},
\]
also,
$$
\|\thetaon_{k}-\theta^*\|_{M_k^{\on}} 
= 
\|M_k^{\on}(\thetaon_{k}-\theta^*)\|_{(M_k^{\on})^{-1}} 
\le 
\left\|\sum_{k'=1}^{k-1}\sum_{h=1}^{H} x^{\on}_{k',h}\eta^{\on}_{k',h}\right\|_{(M_k^{\on})^{-1}} + \|\lambda\theta^*\|_{(M^{\on}_{k})^{-1}}.
$$
Furthermore, by definition,
\[
\|b^{\off}\|_{(M^{\all}_k)^{-1}}
\le
\|b^{\off}\|_{(M^{\all}_1)^{-1}}
\le
\sqrt{\lambda_{\max}\bigl((M^{\all}_1)^{-1}\bigr)} \|b^{\off}\|_2
=
\frac{1}{\sqrt{\lambda_{\min}(M^{\all}_1)}} \|b^{\off}\|_2,
\]
where the first inequality follows from $M^{\all}_k \succeq M^{\all}_1$ and Lemmas~\ref{lem_monotone_norm} and~\ref{lem_inverse_order} (see \cref{sec:useful_facts}), which together imply $(M^{\all}_k)^{-1} \preceq (M^{\all}_1)^{-1}$, the second inequality is from~\Cref{lem_weighted_to_l2} (\cref{sec:useful_facts}), and the final equality follows from the spectral identity $\lambda_{\max}(A^{-1}) = 1/\lambda_{\min}(A)$ for any symmetric positive definite matrix.
Thus
\begin{align*}
\|b^{\off}\|_{(M^{\all}_k)^{-1}} 
\le \frac{1}{\sqrt{\lambda_{\min}(M^{\all}_1)}} \|b^{\off}\|_2
\le \frac{1}{\sqrt{\lambda_{\min}(M^{\all}_1)}}\lambda_{\max}(G_{\off})\|\theta^{\off,*}-\theta^*\|_2
\le \Delta\frac{\lambda_{\max}(G_{\off})}{\sqrt{\lambda_{\min}(\lambda I + G_{\off})}},
\end{align*}
where the first inequality again follows from~\Cref{lem_weighted_to_l2}, noting that the bound holds for any symmetric positive semi-definite matrix, even though the induced quantity $\|\cdot\|_A$ is only a seminorm when $A$ is not strictly positive definite. The last inequality follows directly from~\Cref{assump_theta_gap}, which ensures $\|\theta^{\off,*}-\theta^*\|_2 \le \Delta$, together with the definition of $M^{\all}_1$.

Finally, since $M^{\all}_{k}\succeq \lambda I$ and $M^{\on}_{k}\succeq \lambda I$ for all $k$,
\(
\|\lambda\theta^*\|_{(M^{\all}_{k})^{-1}} \le \sqrt{\lambda}\|\theta^*\|_2 \le
\sqrt{\lambda}B, 
\|\lambda\theta^*\|_{(M^{\on}_{k})^{-1}} \le  \sqrt{\lambda}\|\theta^*\|_2 \le \sqrt{\lambda}B,
\)
where we use the bound $\max\{\|\theta^{\off,*}\|_2,\|\theta^*\|_2\}\le B$ from~\Cref{def_mdp_1}. The proof is complete with a union bound.
\end{rproof}

With the self-normalized concentration bounds~\Cref{lemma_biased_ellipsoid} in place, we next establish the key optimism property of the value estimates produced by~\Cref{alg_main}, which is crucial for the subsequent regret analysis.

\begin{lemma}
\label{lemma_optimism}
Conditional on $\E_0$, for any episode $k$, the online optimistic Q-function and value function satisfy
\[
\Qoptimal_{h}(s,a) \le \Qhat_{k,h}(s,a), 
\qquad
\Voptimal_{h}(s) \le \Vhat_{k,h}(s),
\]
for all states $s$, actions $a$, and stages $h=1,\ldots,H+1$.
\end{lemma}

\begin{rproof}
We prove the result by induction. For the base case $h = H+1$, by definition,
\[
\Qoptimal_{H+1}(s,a)=0,
\qquad
\Qhat_{k,H+1}(s,a)=0,
\]
for all $s,a$. Hence the claim holds trivially. Assume that for some $h+1\le H$,
\[
\Qoptimal_{h+1}(s,a)\le \Qhat_{k,h+1}(s,a) \quad \text{for all } s,a.
\]
Then, by the definition of the value function,
\[
\Voptimal_{h+1}(s) = \max_{a\in\mathcal A} \Qoptimal_{h+1}(s,a) \le \max_{a\in\mathcal A} \Qhat_{k,h+1}(s,a) = \Vhat_{k,h+1}(s),
\]
for all states $s$. Furthermore, for any state--action pair $(s,a)$, we have
\[
\Qoptimal_{h}(s,a) = r(s,a) + P(\cdot\mid s,a)^\top \Voptimal_{h+1}.
\]
By the linear mixture model,
\[
P(\cdot\mid s,a) =\langle \theta^*, \phi(\cdot\mid s,a)\rangle,
\]
hence
\[
P(\cdot\mid s,a)^\top \Voptimal_{h+1} = \sum_{j=1}^d \theta^*_j \phi_j(\cdot\mid s,a)^\top \Voptimal_{h+1} \le \sum_{j=1}^d \theta^*_j \phi_j(\cdot\mid s,a)^\top \Vhat_{k,h+1}.
\]
Since $\theta^* \in \mathcal B_k\cap C_k$, we further have via the induction hypothesis that
\begin{align*}
\Qoptimal_{h}(s,a)
&= r(s,a) + P(\cdot\mid s,a)^\top \Voptimal_{h+1} \le r(s,a) + \sum_{j=1}^d \theta^*_j \phi_j(\cdot\mid s,a)^\top \Vhat_{k,h+1} \\
&\le r(s,a) + \max_{\theta\in\mathcal B_k \cap \C_k} \sum_{j=1}^d \theta_j \phi_j(\cdot\mid s,a)^\top \Vhat_{k,h+1} = \Qhat_{k,h}(s,a).
\end{align*}
Taking the maximum over $a$ on both sides gives $\Voptimal_{h}(s)\le \Vhat_{k,h}(s)$, which completes the proof.
\end{rproof}

We next bound the gap between the optimistic value estimates produced by the algorithm and the true value functions induced by the executed policy $\pi_k$.
The analysis follows a standard telescoping argument along the trajectory within an episode: the gap at stage $h$ is related to the gap at stage $h+1$, plus a martingale difference term accounting for the randomness of the next state.
The remaining term is controlled by the uncertainty in the transition parameter, which is quantified through the confidence radii associated with the sets $\mathcal C_k$ and $\mathcal B_k$.

\begin{lemma}
\label{lemma_telescoping}
Let $\pi_k$ be the implemented policy at episode $k$. Then conditional on $\E_0$ and $\F_{k,h+1}$,
\begin{align*}
&\Vhat_{k,h}\bigl(s^{\on}_{k,h}\bigr) - \Vpik_{h}\bigl(s^{\on}_{k,h}\bigr)
\le
\Vhat_{k,h+1}\bigl(s^{\on}_{k,h+1}\bigr) - \Vpik_{h+1}\bigl(s^{\on}_{k,h+1}\bigr) + \xi^{\on}_{k,h+1}
\\
&\qquad
+2 \min\left\{\gamma_k\left\|\phi(\cdot\mid s^{\on}_{k,h},a^{\on}_{k,h})\Vhat_{k,h+1}\right\|_{(M^{\all}_k)^{-1}}, \beta_k\left\|\phi(\cdot\mid s^{\on}_{k,h},a^{\on}_{k,h})\Vhat_{k,h+1}\right\|_{(M^{\on}_k)^{-1}}\right\},
\end{align*}
where
\[
\xi^{\on}_{k,h+1} = P(\cdot\mid s^{\on}_{k,h},a^{\on}_{k,h})^\top \bigl(\Vhat_{k,h+1}-\Vpik_{h+1}\bigr) - \Bigl(\Vhat_{k,h+1}(s^{\on}_{k,h+1})-\Vpik_{h+1}(s^{\on}_{k,h+1})\Bigr).
\]
\end{lemma}

\begin{rproof}
Recall that $a^{\on}_{k,h}=\pi_{k,h}(s^{\on}_{k,h})$ and $\Vpik_{h}(s)=\Qpik_{h}(s,\pi_{k,h}(s))$. Hence
\[
\Vhat_{k,h}\left(s^{\on}_{k,h}\right)-\Vpik_{h}\left(s^{\on}_{k,h}\right) = \Qhat_{k,h}\left(s^{\on}_{k,h},a^{\on}_{k,h}\right) - \Qpik_{h}\left(s^{\on}_{k,h},a^{\on}_{k,h}\right).
\]
Recall also that
\[
x^{\on}_{k,h} = \phi(\cdot\mid s^{\on}_{k,h},a^{\on}_{k,h})\Vhat_{k,h+1}.
\]
By the definition of $\Qhat_{k,h}$ and the linear mixture model,
\begin{align*}
\Qhat_{k,h}&\left(s^{\on}_{k,h},a^{\on}_{k,h}\right) - \Qpik_{h}\left(s^{\on}_{k,h},a^{\on}_{k,h}\right) \\
&= 
r\left(s^{\on}_{k,h},a^{\on}_{k,h}\right) + \max_{\theta\in\B_k \cap \C_k} \theta^\top x^{\on}_{k,h} - \Bigl(r\left(s^{\on}_{k,h},a^{\on}_{k,h}\right) + (\theta^*)^\top \phi(\cdot\mid s^{\on}_{k,h},a^{\on}_{k,h})\Vpik_{h+1} \Bigr) \\
&=
(\theta^*)^\top x^{\on}_{k,h} - (\theta^*)^\top \phi(\cdot\mid s^{\on}_{k,h},a^{\on}_{k,h})\Vpik_{h+1} + \max_{\theta\in\B_k \cap \C_k}(\theta-\theta^*)^\top x^{\on}_{k,h} \\
&=
P(\cdot\mid s^{\on}_{k,h},a^{\on}_{k,h})^\top \bigl(\Vhat_{k,h+1}-\Vpik_{h+1}\bigr) + \max_{\theta\in\B_k \cap \C_k}(\theta-\theta^*)^\top x^{\on}_{k,h},
\end{align*}
where the first equality is from the definition of $\Qhat_{k,h}$. Since $\theta^*\in\B_k \cap \C_k$, $\B_k=\{\theta:\|\theta - \thetaall_k\|_{M^{\all}_k}\le\gamma_k\}$, and $\C_k = \bigl\{\theta \in \mathbb{R}^d : \|\theta - \thetaon_{k}\|_{M^{\on}_{k}} \le \beta_k\bigr\}$,
\begin{align*}
\max_{\theta\in\B_k \cap \C_k}(\theta-\theta^*)^\top x^{\on}_{k,h}
&\le
\max_{\theta\in\B_k \cap \C_k}\bigl|(\theta-\thetaall_k)^\top x^{\on}_{k,h}\bigr| + \bigl|(\theta^*-\thetaall_k)^\top x^{\on}_{k,h}\bigr| \\
&\le
\max_{\theta\in\B_k \cap \C_k}\|(M^{\all}_k)^{\frac{1}{2}}(\theta-\thetaall_k)\|_2
\|(M^{\all}_k)^{-\frac{1}{2}}x^{\on}_{k,h}\|_2 \\
&\qquad
+ \|(M^{\all}_k)^{\frac{1}{2}}(\theta^*-\thetaall_k)\|_2 \|(M^{\all}_k)^{-\frac{1}{2}}x^{\on}_{k,h}\|_2 \\
&\le
\gamma_k\|x^{\on}_{k,h}\|_{(M^{\all}_k)^{-1}} + \gamma_k\|x^{\on}_{k,h}\|_{(M^{\all}_k)^{-1}}
=
2\gamma_k\|x^{\on}_{k,h}\|_{(M^{\all}_k)^{-1}}.
\end{align*}
where the second inequality is from Cauchy-Schwarz inequality. Similarly, we have that
\begin{align*}
\max_{\theta\in\B_k \cap \C_k}(\theta-\theta^*)^\top x^{\on}_{k,h}
&\le
\max_{\theta\in\B_k \cap \C_k}\bigl|(\theta-\thetaon_k)^\top x^{\on}_{k,h}\bigr| + \bigl|(\theta^*-\thetaon_k)^\top x^{\on}_{k,h}\bigr| \\
&\le
\max_{\theta\in\B_k \cap \C_k}\|(M^{\on}_k)^{\frac{1}{2}}(\theta-\thetaon_k)\|_2 \|(M^{\on}_k)^{-\frac{1}{2}}x^{\on}_{k,h}\|_2 \\
&\qquad
+ \|(M^{\on}_k)^{\frac{1}{2}}(\theta^*-\thetaon_k)\|_2 \|(M^{\on}_k)^{-\frac{1}{2}}x^{\on}_{k,h}\|_2 \\
&\le
\beta_k\|x^{\on}_{k,h}\|_{(M^{\on}_k)^{-1}} + \beta_k\|x^{\on}_{k,h}\|_{(M^{\on}_k)^{-1}}
= 
2\beta_k\|x^{\on}_{k,h}\|_{(M^{\on}_k)^{-1}}.
\end{align*}
The proof is complete with the definition of $\xi^{\on}_{k,h+1}$.
\end{rproof}

The next lemma allows us to bound $\sum_{k = 1}^K\sum_{h = 1}^H \gamma_k\|x^{\on}_{k,h}\|_{(M^{\all}_k)^{-1}}$ and $\sum_{k = 1}^K\sum_{h = 1}^H \beta_k\|x^{\on}_{k,h}\|_{(M^{\on}_k)^{-1}}$ to further simplify the results in~\Cref{lemma_telescoping}.

\begin{lemma}\label{lemma_sum_over_radius}
    For all $k\ge 1$, with $\lambda = H^2d$, we have that 
\begin{align*}
\sum_{k = 1}^K \sum_{h = 1}^H \gamma_k  \|x^{\on}_{k,h}\|_{(M^{\all}_k)^{-1}}
&\le
\sqrt{2HT} \Biggl(H \sqrt{2 \log\Bigl(\frac{\det(M^{\all}_{K})^{1/2}}{\det(H^2 d I)^{1/2}\delta}\Bigr)} + \sqrt{H^2 d} B + \Delta\frac{\lambda_{\max}(G_{\off})}{\sqrt{\lambda_{\min}(H^2d I + G_{\off})}}\Biggr) \\
&\qquad\times
\sqrt{\Bigl(\log(\det(M^{\all}_{K+1}))- \log(\det(M^{\all}_{1}))\Bigr)}, \\
\sum_{k = 1}^K \sum_{h = 1}^H \beta_k  \|x^{\on}_{k,h}\|_{(M^{\on}_k)^{-1}}
&\le
\sqrt{2HT}\Biggl(H \sqrt{2 \log\Bigl(\frac{\det(M^{\on}_{K})^{1/2}}{\det(H^2 d I)^{1/2}\delta}\Bigr)}+ \sqrt{H^2 d} B\Biggr)
\times 
\sqrt{\Bigl(\log(\det(M^{\on}_{K+1}))- \log(\det(H^2dI))\Bigr)}.
\end{align*}
\end{lemma}
\begin{rproof}
Note that due to 
$$
M^{\all}_{k}= \lambda I + \sum_{m=1}^{M^{\off}} \sum_{h=1}^{H} x^{\off}_{m,h}\bigl(x^{\off}_{m,h}\bigr)^{\top} + \sum_{k'=1}^{k-1} \sum_{h=1}^{H} x^{\on}_{k',h}\bigl(x^{\on}_{k',h}\bigr)^{\top}, 
\qquad
M_k^{\on} = \lambda I + \sum_{k'=1}^{k-1} \sum_{h=1}^{H} x^{\on}_{k',h}\bigl(x^{\on}_{k',h}\bigr)^{\top}
$$
we have that
\begin{align}
\|x^{\on}_{k,h}\|_{(M^{\all}_k)^{-1}}^2
& = (x^{\on}_{k,h})^\top (M^{\all}_k)^{-1} x^{\on}_{k,h} \le (x^{\on}_{k,h})^\top (\lambda I)^{-1} x^{\on}_{k,h} = \frac{1}{\lambda}\|x^{\on}_{k,h}\|_2^2 \nonumber\\
&= \frac{1}{\lambda} \sum_{j=1}^d\bigl(\phi_j(\cdot \mid s^{\on}_{k,h}, a^{\on}_{k,h})^\top \Vhat_{k,h+1}\bigr)^2 \le \frac{1}{\lambda} \sum_{j=1}^d \|\phi_j(\cdot \mid s^{\on}_{k,h}, a^{\on}_{k,h})\|_1^2 \|\Vhat_{k,h+1}\|_\infty^2\le \frac{H^2 d}{\lambda} = 1,
\label{eq_bound_of_c2}
\end{align}
where the first inequality is from~\Cref{lem_monotone_norm}, and the last inequality is from the assumption that $\|\phi_j(\cdot \mid s,a)\|_1 \le 1$ for all $(s,a,j)$.  Also, 
\begin{align*}
\|x^{\on}_{k,h}\|_{(M^{\on}_k)^{-1}}^2 = (x^{\on}_{k,h})^\top (M^{\on}_k)^{-1} x^{\on}_{k,h} \le (x^{\on}_{k,h})^\top (\lambda I)^{-1} x^{\on}_{k,h} = \frac{1}{\lambda}\|x^{\on}_{k,h}\|_2^2 \le \frac{H^2 d}{\lambda} = 1.
\end{align*}

Thus by denoting 
$$
\gamma_{k,1} = H \sqrt{2 \log\Bigl(\frac{\det(M^{\all}_{k})^{1/2}}{\det(\lambda I)^{1/2}\delta}
\Bigr)} + \sqrt{\lambda}B, 
\qquad
\gamma_{\Delta} = \Delta\frac{\lambda_{\max}(G_{\off})}{\sqrt{\lambda_{\min}(\lambda I + G_{\off})}},
$$
it's easy to see that $\gamma_k = \gamma_{k,1} + \gamma_{\Delta}$, and (1) $\gamma_{k,1}$ is non-decreasing w.r.t. $k$; (2) $\gamma_{\Delta}$ is a constant after observing all offline data.  Thus,
\begin{align*}
\sum_{k = 1}^K\sum_{h = 1}^H \gamma_k\|x^{\on}_{k,h}\|_{(M^{\all}_k)^{-1}} 
&\le \sum_{k = 1}^K\sum_{h = 1}^H \gamma_k\min\{1, \|x^{\on}_{k,h}\|_{(M^{\all}_k)^{-1}}\} \\
&\le \left(\gamma_{K,1} + \gamma_{\Delta}\right) \sqrt{HK}\sqrt{\sum_{k = 1}^K\sum_{h = 1}^H \min\{1,\|x^{\on}_{k,h}\|_{(M^{\all}_k)^{-1}}^2\}} \\
&\le \left(\gamma_{K,1} + \gamma_{\Delta}\right) \sqrt{2HK}\sqrt{\sum_{k = 1}^K\sum_{h = 1}^H \log (1 + \|x^{\on}_{k,h}\|_{(M^{\all}_k)^{-1}}^2)} \\
&\le \left(\gamma_{K,1} + \gamma_{\Delta}\right) \sqrt{2HK}\sqrt{\sum_{k = 1}^K H\log(1 + \sum_{h = 1}^H \|x^{\on}_{k,h}\|_{(M^{\all}_k)^{-1}}^2)},
\end{align*}
where the first inequality is from~\cref{eq_bound_of_c2}, the second inequality is from Cauchy-Schwarz inequality and two arguments over $\gamma_k$ above, the third inequality is from that $\min\{1, x\} \le 2\log(1 + x)$ for all $x \ge 0$, and the last inequality is from that for any $\{x_h\}_{h = 1}^H$ with $x_h \ge 0$ for all $h \in [H]$, $1 + \sum_{h = 1}^H x_h = \frac{\sum_{h = 1}^H (1 + Hx_h)}{H} \ge \Pi_{h =1}^H (1 + Hx_h)^{1/H} \ge \Pi_{h =1}^H (1 + x_h)^{1/H}$, thus $\sum_{h = 1}^H \log(1 + x_h) \le H\log(1 + \sum_{h = 1}^H x_h)$. The same argument applies to the term $\sum_{k = 1}^K\sum_{h = 1}^H \beta_k\|x^{\on}_{k,h}\|_{(M^{\on}_k)^{-1}}$, and we have that
\begin{align*}
    \sum_{k = 1}^K\sum_{h = 1}^H \beta_k\|x^{\on}_{k,h}\|_{(M^{\on}_k)^{-1}} \le \beta_{K} \sqrt{2HK}\sqrt{\sum_{k = 1}^K H\log(1 + \sum_{h = 1}^H \|x^{\on}_{k,h}\|_{(M^{\on}_k)^{-1}}^2)}.
\end{align*}

Furthermore, note that 
\begin{align*}
1 + \sum_{h = 1}^H\|x^{\on}_{k,h}\|_{(M^{\all}_k)^{-1}}^2 
&= 1 + \operatorname{tr}(\sum_{h = 1}^H (M^{\all}_k)^{-\frac{1}{2}}x^{\on}_{k,h}(x^{\on}_{k,h})^{\top}(M^{\all}_k)^{-\frac{1}{2}}) \\
&\le \det(I + \sum_{h = 1}^H(M^{\all}_k)^{-\frac{1}{2}}x^{\on}_{k,h}(x^{\on}_{k,h})^{\top}(M^{\all}_k)^{-\frac{1}{2}}) = \frac{\det(M^{\all}_{k+1})}{\det(M^{\all}_k)},
\end{align*}
where the second inequality is due to the fact that $\Pi_i (1 + x_i) \ge 1 + \sum_i x_i$ for all $x \ge 0$, and the third line is due to the definition of $M^{\all}_k$. Similarly, for the online-only estimator, we have
\begin{align*}
1 + \sum_{h = 1}^H\|x^{\on}_{k,h}\|_{(M^{\on}_k)^{-1}}^2 
&= 1 + \operatorname{tr}( \sum_{h = 1}^H (M^{\on}_k)^{-\frac{1}{2}}x^{\on}_{k,h}(x^{\on}_{k,h})^{\top}(M^{\on}_k)^{-\frac{1}{2}}) \\
&\le \det(I + \sum_{h = 1}^H(M^{\on}_k)^{-\frac{1}{2}}x^{\on}_{k,h}(x^{\on}_{k,h})^{\top}(M^{\on}_k)^{-\frac{1}{2}}) = \frac{\det(M^{\on}_{k+1})}{\det(M^{\on}_k)}.
\end{align*}
Correspondingly, we have that 
\begin{align*}
&\sum_{k = 1}^K \sum_{h = 1}^H \gamma_k  \|x^{\on}_{k,h}\|_{(M^{\all}_k)^{-1}}
\le
\left(\gamma_{K,1} + \gamma_{\Delta}\right) \sqrt{2HK} \sqrt{H\log(\det(M^{\all}_{K+1}))-H\log(\det(M^{\all}_{1}))} \\
&=
\sqrt{2H^2K} \Biggl(H \sqrt{2 \log\Bigl(\frac{\det(M^{\all}_{K})^{1/2}}{\det(H^2 d I)^{1/2}\delta}\Bigr)}+ \sqrt{\lambda} B + \Delta\frac{\lambda_{\max}(G_{\off})}{\sqrt{\lambda_{\min}(H^2dI + G_{\off})}}\Biggr)
\sqrt{\log(\det(M^{\all}_{K+1})) - \log(\det(M^{\all}_{1}))}.
\end{align*}
The same argument applies to the term \(\sum_{k=1}^K \sum_{h=1}^H \beta_k \|x^{\on}_{k,h}\|_{(M^{\on}_k)^{-1}}\), which completes the proof.
\end{rproof}

With the previous lemmas in hand we are ready to prove \Cref{thm1}.

\begin{rproofof}{\cref{thm1}}
Recall that the good event $\E_0$ occurs with probability at least $1- 2\delta$ due to~\cref{lemma_biased_ellipsoid}, where 
\(
\E_0 = \bigl\{\theta^* \in \B_k \cap \C_k, \text{ for all } k \ge 1 \bigr\}.
\)
Furthermore, for any fixed $\delta' \in (0,1)$, we denote 
\(
    \E_1 = \left\{\frac{1}{M^{\off}}\lambda_{\min}(\lambda I + G_{\off})\ge \tau(\delta')\right\}.
\)
From~\Cref{assumption_eigen}, we have that $\Pr(\E_1) \ge 1 - \delta'$. Now we define
\(
    \E_2 = \E_0 \cap \E_1.
\)
Then $\Pr(\E_2) \ge 1 - 2\delta - \delta'$ through a union bound.
Thus we have that
\begin{align*}
&\Ex\left[\sum_{k=1}^K \bigl(\Voptimal_1(s_1) - \Vpik_1(s_1) \bigr) \Bigm| \E_2\right] 
\le 
\Ex\left[\sum_{k=1}^K\bigl(\Vhat_{k,1}(s_1) - \Vpik_1(s_1) \bigr)\Bigm|\E_2\right] \\
&\qquad = \Ex\left[\sum_{k=1}^K\sum_{h = 1}^H\bigl(\Vhat_{k,h}(s^{\on}_{k,h}) - \Vpik_h(s^{\on}_{k,h}) - \Vhat_{k,h+1}(s^{\on}_{k,h+1}) + \Vpik_{h+1}(s^{\on}_{k,h+1}) \bigr) \Bigm| \E_2\right] \\
&\qquad= \sum_{k=1}^K\sum_{h = 1}^H\Ex[ \Ex[ \Vhat_{k,h}(s^{\on}_{k,h}) - \Vpik_h(s^{\on}_{k,h}) - \Vhat_{k,h+1}(s^{\on}_{k,h+1})  + \Vpik_{h+1}(s^{\on}_{k,h+1})\mid \{\E_2, \F_{k,h+1}\}]\mid\E_2] \\
&\qquad\le \sum_{k=1}^K \sum_{h=1}^H \mathbb{E}[\mathbb{E}[\xi^{\on}_{k,h+1} +2 \min\{\gamma_k \|\phi(\cdot\mid s^{\on}_{k,h},a^{\on}_{k,h}) \Vhat_{k,h+1} \|_{(M^{\all}_k)^{-1}}, \\
&\qquad\qquad
\beta_k\|\phi(\cdot\mid s^{\on}_{k,h},a^{\on}_{k,h})\Vhat_{k,h+1}\|_{(M^{\on}_k)^{-1}}\} \mid \{\E_2,\mathcal F_{k,h+1}\}] \mid \E_2],
\end{align*}
where the first inequality is from~\cref{lemma_optimism} and that $\E_2 \subset \E_0$, and the second inequality is from~\cref{lemma_telescoping} and that $\E_2 \subset \E_0$. Thus we have that 
\begin{align*}
\Ex\left[\sum_{k=1}^K \bigl(\Voptimal_1(s_1) - \Vpik_1(s_1) \bigr) \Bigm| \E_2\right]
&\le
\min\Biggl\{ \Ex\Biggl[2\sqrt{2H^2K} \Biggl(H \sqrt{2 \log\left(\frac{\det(M^{\all}_{K})^{1/2}}{\det(H^2 d I)^{1/2}\delta} \right)}+ \sqrt{H^2 d} B
\\
&\qquad\qquad
+ \Delta\frac{\lambda_{\max}(G_{\off})}{\sqrt{\lambda_{\min}(H^2d I + G_{\off})}}\Biggr) \times \sqrt{\log(\det(M^{\all}_{K+1})) - \log(\det(M^{\all}_{1}))} \Bigm| \E_2 \Biggr],
\\
&\qquad
\Ex \Biggl[ 2\sqrt{2H^2K} \Biggl(H \sqrt{2 \log\left(\frac{\det(M^{\on}_{K})^{1/2}}{\det(H^2 d I)^{1/2}\delta}\right)}+ \sqrt{H^2 d} B\Biggr)
\\
&\qquad\qquad
\times \sqrt{\log(\det(M^{\on}_{K+1})) - \log(\det(H^2 d I))} \Bigm| \E_2 \Biggr]
\Biggr\}
\\
&\qquad
+ \Ex\left[\sum_{k=1}^K \sum_{h=1}^H \xi^{\on}_{k,h+1} \Bigm| \E_2 \right],
\end{align*}
where the second inequality is from~\cref{lemma_sum_over_radius}. As we have $\Ex[\sum_{k=1}^K\sum_{h = 1}^H \xi^{\on}_{k,h+1}] = 0$ and $|\xi^{\on}_{k,h+1}| \le H$ a.s. for all $k,h$, $H^2K(2\delta + \delta')$ is a trivial upper bound over $\Ex[\sum_{k=1}^K\sum_{h = 1}^H \xi^{\on}_{k,h+1}|\E_2]\Pr(\E_2)$. Thus
\begin{align}
\Regret(K)
&\le
\min\Biggl\{ \Ex\Biggl[2\sqrt{2H^2K} \Biggl(H \sqrt{2 \log\left(\frac{\det(M^{\all}_{K})^{1/2}}{\det(H^2 d I)^{1/2}\delta}\right)} + \sqrt{H^2 d} B + \Delta\frac{\lambda_{\max}(G_{\off})}{\sqrt{\lambda_{\min}(H^2d I +G_{\off})}}\Biggr) \nonumber
\\
&\qquad\qquad
\times
\sqrt{\log(\det(M^{\all}_{K+1})) - \log(\det(M^{\all}_{1}))} \Bigm| \E_2 \Biggr], \nonumber
\\
&\qquad
\Ex \Biggl[ 2\sqrt{2H^2K} \Biggl(H \sqrt{2 \log\left(\frac{\det(M^{\on}_{K})^{1/2}}{\det(H^2 d I)^{1/2}\delta}\right)} + \sqrt{H^2 d} B \Biggr) \nonumber
\\
&\qquad\qquad
\times
\sqrt{\log(\det(M^{\on}_{K+1})) - \log(\det(H^2 d I))} \Bigm| \E_2 \Biggr]
\Biggr\} 
+ H^2 K (2\delta + \delta') + H K (2\delta + \delta'), \nonumber
\end{align}
where the term \(HK(2\delta + \delta')\) accounts for the contribution of the failure event \(\E_2^c\) to the expected regret, using the facts that \(\Regret(K)\le HK\) almost surely and \(\Pr(\E_2^c)\le 2\delta + \delta'\).

Recall that
\(
M^{\all}_1 = H^2 d I + G_{\off}.
\)
Similar to $G^{\off}$, we define $G^{\on}_{k} = \sum_{k'=1}^{k-1}\sum_{h=1}^{H} x^{\on}_{k',h}\bigl(x^{\on}_{k',h}\bigr)^{\top}$. Thus $M^{\all}_{K+1} = M^{\all}_1 + G^{\on}_K$. Applying Lemma~\ref{lemma_pd} (\cref{sec:useful_facts}) with \(A = M^{\all}_1\) and \(B = G^{\on}_K\), we obtain
\begin{align*}
\log (\det(M^{\all}_{K+1})) - \log (\det(M^{\all}_1)) \le d \log\left(1 + \frac{1}{d} \operatorname{tr}\bigl((M^{\all}_1)^{-1} G^{\on}_K\bigr)\right).
\end{align*}
Since \(M^{\all}_1 \succ 0\) and \(G^{\on}_K \succeq 0\), the trace term can be bounded as
\begin{align*}
\operatorname{tr}\bigl((M^{\all}_1)^{-1} G^{\on}_K\bigr) \le \lambda_{\max}\bigl((M^{\all}_1)^{-1}\bigr) \operatorname{tr}(G^{\on}_K) = \lambda_{\min}(M^{\all}_1)^{-1}\operatorname{tr}(G^{\on}_K),
\end{align*}
where the first inequality is from~\Cref{lem_trace_bound} (\cref{sec:useful_facts}). Moreover, by definition of \(G^{\on}_K\),
\begin{align}
\operatorname{tr}(G^{\on}_K) = \sum_{k=1}^K \sum_{h=1}^H \|x^{\on}_{k,h}\|_2^2 \le H^3 dK,
\label{eq_trace_bound_on_GonK}
\end{align}
where the inequality follows from the boundedness \(\|x^{\on}_{k,h}\|_2^2 \le H^2 d\) and~\cref{eq_bound_of_c2}. Combining the above bounds yields
\[
\log (\det(M^{\all}_{K+1})) - \log (\det(M^{\all}_1)) \le d \log\left(1 + H^3K\lambda_{\min}(H^2 d I + G_{\off})^{-1} \right).
\]
Furthermore, using Lemma~\ref{lemma_pd} (\cref{sec:useful_facts}) again with \(A = H^2dI\) and \(B = G_\off + G^{\on}_K\), and noting that \(\operatorname{tr}(G_{\off}) \le d\lambda_{\max}(G_{\off})\), we obtain that
\begin{align*}
\log\left(\frac{\det\left(M^{\all}_{K}\right)^{1/2}}{\det(H^2 d I)^{1/2}\delta}\right) &\le \log\left(\frac{1}{\delta}\right) + \frac{1}{2}d\log\left(1 + \frac{1}{d}\operatorname{tr}\left(\frac{G_\off + G^{\on}_K}{H^2d}\right)\right) \\
&= \log\left(\frac{1}{\delta}\right) + \frac{1}{2}d\log\left(1 + \frac{1}{H^2d^2}\operatorname{tr}(G_\off + G^{\on}_K)\right) \\
&\le \log\left(\frac{1}{\delta}\right) + \frac{1}{2}d\log\left(1 + (H^2 d)^{-1}\lambda_{\max}(G_{\off}) + \frac{HK}{d}\right),
\end{align*}
where the last inequality follows from the additivity of the trace, \cref{eq_trace_bound_on_GonK}, and the fact that $\operatorname{tr}(G_{\off}) \le d\lambda_{\max}(G_{\off})$. Similarly, we have that
\begin{align*}
\log\left(\frac{\det\left(M^{\on}_{K+1}\right)}{\det(H^2 d I)}\right) \le d\log\left(1 + \frac{HK}{d}\right).
\end{align*}

Thus we obtain that
\begin{align}
\Regret(K)
\le
\min\Biggl\{
&
\Ex \Biggl[2\sqrt{2H^2K} \Biggl[H \sqrt{2 \left(\log \left(\frac{1}{\delta}\right)+ \frac{1}{2} d \log\left(1 + \frac{\lambda_{\max}(G_{\off})}{H^2 d}+ \frac{HK}{d}\right)\right)} + H \sqrt{d} B \nonumber
\\
&\qquad
+ \Delta \frac{\lambda_{\max}(G_{\off})}{\sqrt{\lambda_{\min}(H^2dI + G_{\off})}}
\Biggr] \times
\sqrt{d \log\left(1 + \frac{H^3K}{\lambda_{\min}(H^2 d I + G_{\off})}\right)} \Bigm| \E_2\Biggr], \nonumber
\\
& \Ex \Biggl[2\sqrt{2HT} \left(H \sqrt{2 \left(\log \left(\frac{1}{\delta}\right) + \frac{d}{2} \log\left(1 + \frac{HK}{d}\right)\right)} + H \sqrt{d} B \right) \nonumber
\\
&\qquad\times
\sqrt{d \log\left(1 + \frac{HK}{d}\right)} \Bigm | \E_2 \Biggr]
\Biggr\}
+ H^2 K (2\delta + \delta') + H K (2\delta + \delta') \nonumber\\
\le
\min\Biggl\{
&
\Ex \Biggl[2\sqrt{2H^2K} \Biggl[H \sqrt{2 \left(\log \left(\frac{1}{\delta}\right)+ \frac{1}{2} d \log\left(1 + \frac{\lambda_{\max}(G_{\off})}{H^2 d}+ \frac{HK}{d}\right)\right)} + H \sqrt{d} B \nonumber
\\
&\qquad
+ \Delta \frac{\lambda_{\max}(G_{\off})}{\sqrt{\tau(\delta')M^{\off}}}
\Biggr] \times
\sqrt{d \log\left(1 + \frac{H^3K}{\tau(\delta')M^\off}\right)} \Bigm| \E_2\Biggr], \nonumber
\\
& \Ex \Biggl[2\sqrt{2HT} \left(H \sqrt{2 \left(\log \left(\frac{1}{\delta}\right) + \frac{d}{2} \log\left(1 + \frac{HK}{d}\right)\right)} + H \sqrt{d} B \right) \nonumber
\\
&\qquad\times
\sqrt{d \log\left(1 + \frac{HK}{d}\right)} \Bigm | \E_2 \Biggr]
\Biggr\}
+ H^2 K (2\delta + \delta') + H K (2\delta + \delta'), \label{eq_final}
\end{align}
where the second inequality is from $\E_2 \subset \E_1$ and the definition of $\E_1$.

Finally, we note that $\lambda_{\max}(G_{\off}) \le \operatorname{tr}(G_{\off}) \le H^3 d M^{\off}$, 
where the second inequality follows from~\Cref{eq_bound_of_c2}. We take \(\delta = \frac{1}{2K}\) and \(\delta' = \frac{1}{2KM^{\off}}\). As $\tau = \tau(\frac{1}{2KM^\off})$, substituting these choices into the expression above yields that
\begin{align}
\Regret(K) 
\le
\min\Biggl\{
\tilde{O}\Biggl(
&2\sqrt{2H^2K}
\Biggl[H \sqrt{2 \left(\log (2K)+ \frac{1}{2} d \log\left(1 + \frac{H^3dM^{\off}}{H^2 d} + \frac{HK}{d}\right)\right)} + H \sqrt{d} B + \Delta H^3d\sqrt{\frac{M^{\off}}{\tau}}
\Biggr]\nonumber
\\
&\qquad\times
\sqrt{d \log\left(1 + \frac{H^3K}{\tau M^{\off}}\right)}\Biggr),\nonumber
\\
&
2\sqrt{2H^2K}\Biggl( H \sqrt{2 \left(\log (2K) + \frac{d}{2} \log\left(1 + \frac{HK}{d}\right)\right)} + H \sqrt{d}B \Biggr)\nonumber
\\
&\qquad\times
\sqrt{d \log\left(1 + \frac{HK}{d}\right)}
\Biggr\}
+ H^2(1 + \frac{1}{2M^\off}) +  H(1 + \frac{1}{2M^\off}).
\label{eq_without_tildeo}
\end{align}
Equivalently, up to logarithmic factors, we have
\begin{align*}
\Regret(K) = \tilde O\left(\min\left\{\left[BH^2d \sqrt{K} + H^4d\Delta\sqrt{dK} \sqrt{\frac{M^{\off}}{\tau}}\right] \times \sqrt{\log\left(1 + \frac{H^3K}{\tau M^{\off}}\right)},
BH^2d \sqrt{K}
\right\}
\right),
\end{align*}
where $\tilde O(\cdot)$ hides logarithmic factors in $K$ and $M^{\off}$.
\end{rproofof}

\subsection{Proof in~\Cref{sec_when_offlinedata_informative}}
\label{appendix_proof_2}
\Maincorollary*

\begin{rproof}
\label{proof_observation_upper_bound}
Recall from~\Cref{thm1} that
\begin{align*}
\text{term }(A) = \left[BH^2 d \sqrt{K}+ H^4 d^{3/2}\Delta \sqrt{K}\sqrt{\frac{M^{\off}}{\tau}}\right] \sqrt{\log\left(1 + \frac{H^3K}{\tau M^{\off}}\right)}, 
\qquad
\text{term }(B) = BH^2 d \sqrt{K}.
\end{align*}
By definition, the offline data is informative with respect to $\epsilon$ if $\text{term }(A) \le \tilde{O}(\text{term }(B)\cdot K^{-\epsilon})$. Thus from the definition of $\text{term }(A)$, we directly obtain that
\begin{align*}
\text{term }(A) 
&\le \left[BH^2 d \sqrt{K} + H^4 d^{3/2}\Delta \sqrt{K}\sqrt{\frac{M^{\off}}{\tau}}\right] \sqrt{\frac{H^3K}{\tau M^{\off}}} = \frac{BH^{\frac{7}{2}}dK}{\sqrt{\tau M^{\off}}} + \frac{H^{\frac{11}{2}}d^{\frac{3}{2}}\Delta K}{\tau} \\
&\le \frac{BH^{\frac{7}{2}}dK}{\sqrt{H^3K^{1 + 2\epsilon}}} + H^{\frac{11}{2}}d^{\frac{3}{2}}K\frac{B}{H^{7/2} d^{1/2}K^{\frac{1}{2} + \epsilon}} = BH^2dK^{\frac{1}{2} - \epsilon} + BH^2dK^{\frac{1}{2} - \epsilon} = \tilde{O}(\text{term }(B)\cdot K^{-\epsilon})
\end{align*}
where the first inequality follows from the bound $\log(1+x)\le x$ for all $x>0$, and the second line follows from the conditions in~\Cref{cor_obervation_upper_bound}, with absolute constants omitted.
\end{rproof}

\subsection{Proof in~\Cref{sec_comparison}}
\label{appendix_proof_3}
\PropositionLearnability*
\begin{rproof}
    We prove the contrapositive of the claim. Suppose there exist a behavior policy and a design of $\{\TildeV_{m,h}\}$ such that for some $\delta \in (0,1)$, we have $\tau(\delta) > \frac{\lambda}{M^\off}$. By the definition of $\tau(\delta)$ in~\Cref{assumption_eigen}, this implies that, with probability at least $1-\delta$,
    \begin{align}
        \lambda_{\min}(G_\off) = \lambda_{\min}\left(\sum_{m = 1}^{M^\off}\sum_{h = 1}^H \phi\left(\cdot \mid s^{\off}_{m,h}, a^{\off}_{m,h}\right)\TildeV_{m,h+1} \left(\phi\left(\cdot \mid s^{\off}_{m,h}, a^{\off}_{m,h}\right)\TildeV_{m,h+1}\right)^{\top}\right)> 0. 
        \label{eq_proposition_1}
    \end{align}
    We now show that this implies the feature mapping $\phi$ of the offline MDP is uniformly learnable. 
    If~\cref{eq_proposition_1} holds with probability at least $1-\delta$, then there exists a realization of the offline data, denoted as 
        $\left\{\left(s^{\off,\text{sample}}_{m,h}, a^{\off,\text{sample}}_{m,h}\right)\right\}_{(m,h)}$,
    together with the corresponding realization of the designed vectors $\left\{\TildeV^{\text{sample}}_{m,h}\right\}_{(m,h)}$, such that
    \begin{align*}
        \lambda_{\min}\biggl(&\sum_{m = 1}^{M^\off}\sum_{h = 1}^H \phi\left(\cdot \mid s^{\off,\text{sample}}_{m,h}, a^{\off,\text{sample}}_{m,h}\right)\TildeV^{\text{sample}}_{m,h+1}
        \left(\phi\left(\cdot \mid s^{\off,\text{sample}}_{m,h}, a^{\off,\text{sample}}_{m,h}\right)\TildeV_{m,h+1}^{\text{sample}}\right)^{\top}\biggr)> 0. 
    \end{align*}

    Define $\{(s_i,a_i)\}_{i=1}^I$ as the collection $\left\{\left(s^{\off,\text{sample}}_{m,h}, a^{\off,\text{sample}}_{m,h}\right)\right\}_{(m,h)}$, and let $\{V_1,\ldots,V_L\}$ be the collection $\left\{\TildeV^{\text{sample}}_{m,h+1}\right\}_{(m,h)}$ as in~\Cref{assumption_offlineRL_span_main}. Then
    \begin{align*}
        &\lambda_{\min}\left( \sum_{i=1}^{I}\sum_{l=1}^L \phi(\cdot \mid s_i,a_i) V_l \bigl(\phi(\cdot \mid s_i,a_i) V_l\bigr)^{\top}\right) \ge \\
        &\qquad\lambda_{\min}\biggl(\sum_{m = 1}^{M^\off}\sum_{h = 1}^H \phi\left(\cdot \mid s^{\off,\text{sample}}_{m,h}, a^{\off,\text{sample}}_{m,h}\right)\TildeV_{m,h+1}^{\text{sample}} \left(\phi\left(\cdot \mid s^{\off,\text{sample}}_{m,h}, a^{\off,\text{sample}}_{m,h}\right)\TildeV_{m,h+1}^{\text{sample}}\right)^{\top}\biggr) > 0,
    \end{align*}
    where the first inequality holds because the omitted terms are rank-one symmetric positive semidefinite matrices. Hence, the feature mapping $\phi$ of the offline MDP is uniformly learnable. 
\end{rproof}

\CorollaryTauUniformCoverage*
 \begin{rproof}
Fix any $h'\in[H]$. For each offline trajectory $m\in[M^{\off}]$, define the random PSD matrix
\[
X_m =  \phi(\cdot\mid s^{\off}_{m,h'},a^{\off}_{m,h'})\TildeV_{m,h'}\bigl(\phi(\cdot\mid s^{\off}_{m,h'},a^{\off}_{m,h'})\TildeV_{m,h'}\bigr)^{\top}.
\]
Note that $\{X_m\}_{m=1}^{M^{\off}}$ are independent. Let $V_s=[V_1,\ldots,V_L]\in\mathbb R^{|\mathcal S|\times L}$, where $\{V_1,\ldots,V_L\}$ denote the probe vectors in~\Cref{assumption_offlineRL_span_main}.
Using the definition of $\TildeV_{m,h'}$ and conditioning on 
$(s^{\off}_{m,h'},a^{\off}_{m,h'})$, we obtain
\begin{align*}
\Ex[X_m]
&=
\Ex \left[\Ex \left[\phi(\cdot\mid s^{\off}_{m,h'},a^{\off}_{m,h'})\TildeV_{m,h'}\bigl(\phi(\cdot\mid s^{\off}_{m,h'},a^{\off}_{m,h'})\TildeV_{m,h'}\bigr)^{\top} \,\middle|\,(s^{\off}_{m,h'},a^{\off}_{m,h'})\right] \right] \\
&=
\frac{1}{L^2}\Ex \left[\sum_{l=1}^L\phi(\cdot\mid s^{\off}_{m,h'},a^{\off}_{m,h'}) V_l \left(\phi(\cdot\mid s^{\off}_{m,h'},a^{\off}_{m,h'}) V_l\right)^{\top}\right],
\end{align*}
where the last equality follows from taking the expectation over $\TildeV_{m,h'}$ conditional on the visitation of $(s^{\off}_{m,h'}, a^{\off}_{m,h'})$.

Since the feature mapping of the offline MDP is uniformly learnable, \Cref{assumption_offlineRL_span_main} implies that there exists a set of state--action pairs $\{(s_i,a_i)\}_{i=1}^{I}$ for which~\cref{eq_uniform_learnability} holds. By \Cref{assumption_data_coverage_main}, for each state–action pair $(s_i,a_i)$, $i \in [I]$,
\(
d_{h'}^{\pi^{\off}}(s_i,a_i)\ge p_0.
\)
Restricting the expectation to the set $\{(s_i,a_i)\}_{i=1}^{I}$ yields
\[
\Ex[X_m] \succeq \frac{p_0}{L^2} \sum_{i=1}^I\sum_{l=1}^L \phi(\cdot\mid s_i,a_i)V_l \bigl(\phi(\cdot\mid s_i,a_i)V_l\bigr)^{\top}  \succeq \frac{p_0\kappa}{L^2} I,
\]
where the last inequality is from~\Cref{assumption_offlineRL_span_main}.
Consequently,
\(
\mu_{\min} := \lambda_{\min} \left(\sum_{m=1}^{M^{\off}}\Ex[X_m] \right) \ge M^{\off}\frac{p_0\kappa}{L^2}.
\)

Since $\TildeV_{m,h'}\in[-1,1]^{|\S|}$ for all $m$ and $\sum_{s'}|\phi_j(s'\mid s,a)|\le 1$ for all $j$ and all $(s,a)$, we have that for all $m$, 
\[
\lambda_{\max}(X_m) = \|\phi(\cdot\mid s^{\off}_{m,h'},a^{\off}_{m,h'})\TildeV_{m,h'}\|_2^2 \le d \quad\text{almost surely}.
\]
Applying the matrix Chernoff inequality (\Cref{lem_matrix_concentration}) with $\lambda_{\max}(X_m) \le d$ a.s., we obtain that for any $\delta \in(0,1)$,
\[
\Pr\left(\lambda_{\min}\left(\sum_{m=1}^{M^{\off}} X_m \right) \le (1-\delta)\mu_{\min} \right) \le d\exp\left(-\frac{\delta^2}{2d}\mu_{\min}\right).
\]
By rearranging terms, we obtain that with probability at least $1-\delta$,
\[
\frac{1}{M^{\off}} \lambda_{\min}\left(\sum_{m=1}^{M^{\off}} X_m\right) \ge \frac{p_0\kappa}{L^2} - \sqrt{\frac{2dp_0\kappa}{L^2M^{\off}}\log\left(\frac{d}{\delta}\right)}.
\]
Finally, since
\[
G^{\off} = \sum_{h=1}^H\sum_{m=1}^{M^{\off}} \phi(\cdot\mid s^{\off}_{m,h},a^{\off}_{m,h}) \TildeV_{m,h} \bigl(\phi(\cdot\mid s^{\off}_{m,h},a^{\off}_{m,h}) \TildeV_{m,h}\bigr)^{\top} \succeq \sum_{m=1}^{M^{\off}} X_m,
\]
the same lower bound applies to $\lambda_{\min}(G^{\off})/M^{\off}$, which proves~\cref{eq_bound_stochastic_design_under_delta}. \cref{eq_bound_stochastic_design} follows directly by letting $\delta = \frac{1}{2KM^\off}$ in~\cref{eq_bound_stochastic_design_under_delta} and requiring that
\(
    \sqrt{\frac{2 d p_0 \kappa}{L^2 M^{\off}}\log\left(\frac{d}{\delta}\right)} \le \frac{1}{2}\frac{p_0 \kappa}{L^2}.
\)
\end{rproof}

\CorollaryTauDeterministicUniformCoverage*
\begin{rproof}
Note that under the deterministic design of $\{\TildeV_{m,h}\}$ in~\Cref{cor_uniformdata_coverage2}, the corresponding block of $G_\off$ associated with any fixed $(s,a)$ takes the form
\begin{align}
    \operatorname{diag}\left(\frac{N(s,a)}{|\S|}, \ldots, \frac{N(s,a)}{|\S|}\right),
    \label{eq_proof_tabular_deterministic}
\end{align}
where $N(s,a)$ denotes the number of visits to the state--action pair $(s,a)$ in the offline phase. For simplicity, we assume that $|\S|$ divides $N(s,a)$ for every $(s,a)$.
Fix any $h' \in [H]$. For any fixed $(s,a)$ pair, for any $m \in [M^\off]$, let 
\[
    X_{m,h'}^{(s,a)} = \mathbf{1}\{(s^{\off}_{m,h'}, a^{\off}_{m,h'}) = (s,a)\},
\]
where $\mathbf{1}$ denotes the indicator function.
By assumption, $\Pr(X_{m,h'}^{(s,a)} = 1) \ge p_0$, and the random variables $\{X_{m,h'}^{(s,a)}\}_{m=1}^{M^\off}$ are independent across trajectories. Hence,
\(
    N_{h'}(s,a) := \sum_{m=1}^{M^\off} X_{m,h'}^{(s,a)}
\)
is a sum of independent Bernoulli random variables with mean at least $M^\off p_0$. 

By Hoeffding’s inequality (Chap 2 of \citep{vershynin2018high}), for any $\epsilon>0$,
\[
    \Pr\left(N_{h'}(s,a) \le M^\off p_0 - \epsilon\right) \le \exp\left(-\frac{2\epsilon^2}{M^\off}\right).
\]
Setting
\(
    \epsilon = \sqrt{\frac{M^\off}{2}\log\left(\frac{|\S||\A|}{\delta}\right)}
\)
for any fixed $\delta \in (0,1)$, we obtain
\[
    \Pr\left(N_{h'}(s,a) \le M^\off p_0 - \sqrt{\frac{M^\off}{2}\log\left(\frac{|\S||\A|}{\delta}\right)}\right) \le \frac{\delta}{|\S||\A|}.
\]
Applying a union bound over all $(s,a) \in \S \times \A$, we conclude that with probability at least $1-\delta$,
\begin{align}
    N_{h'}(s,a) \ge M^\off p_0 - \sqrt{\frac{M^\off}{2}\log\left(\frac{|\S||\A|}{\delta}\right)}
    \label{eq_proof_deterministic_design_hoeffding}
\end{align}
holds simultaneously for all $(s,a)$. Note that $N(s,a) \ge N_{h'}(s,a)$ almost surely for all $(s,a)$ pairs. Hence~\cref{eq_bound_det_design_under_delta} follows by combining~\cref{eq_proof_tabular_deterministic} and~\cref{eq_proof_deterministic_design_hoeffding}. \cref{eq_bound_det_design} follows directly by setting $\delta = \frac{1}{2KM^\off}$ in~\cref{eq_bound_det_design_under_delta} and requiring that
\(
    \frac{1}{|\S|}\sqrt{\frac{1}{2M^\off}\log\left(\frac{|\S||\A|}{\delta}\right)} \le \frac{1}{2}\frac{p_0}{|\S|}.
\)
\end{rproof}

\PropositionAdaptivePoilicy*

The proof of~\Cref{instance_adaptive_policy} follows Proposition~3.7 of~\citet{han2024ucb}, which provides a high probability bound on the number of pulls of each arm under \textsf{UCB1}. We restate their result in~\Cref{lemma_bandit_han2024}, replacing $T$ by $M^\off$ to reflect that \textsf{UCB1} is applied in the offline phase. In particular, the number of pulls of arm $i \in [A]$ is denoted by $n_{i;M^\off}$, which is the random quantity to be controlled.

The bound in~\Cref{lemma_bandit_han2024} depends on several absolute constants, including $n_*(\Theta)$, $\{n^*_{i;M^\off}\}_{i \in [A]}$, $D_*$, and $\mathrm{err}(\Theta)$~\citep{han2024ucb}.
We briefly explain their roles.
Let $\Delta^{\off}_i = \mu^{\off, *} - \mu^{\off}(i)$ and $\Delta^\off = [\Delta^{\off}_i]_{i \in [A]}$.
The quantity $n_*(\Theta) \in [0, M^\off]$ serves as a reference scale for the number of pulls, determined by the equation
\begin{align}
\sum_{i \in [A]}  n_*(\Theta)\left(1 + n_*^{1/2}(\Theta)\frac{\Delta^{\off}_i}{\gamma_{M^\off}}\right)^{-2} = M^\off,
\label{eq_han2024_1}
\end{align}
where $\gamma_{M^\off} = \sqrt{2\log (M^\off)}$ corresponds to the exploration rate in \textsf{UCB1}, matching the $\sqrt{2\log T}$ factor in~\citet{auer2002finite}.
Intuitively, $n_*(\Theta)$ captures the overall exploration budget allocated across arms. Based on $n_*(\Theta)$, the constants
\begin{align}
n^*_{i;M^\off}  =  n_*(\Theta) \left(1 + n_*^{1/2}(\Theta)\frac{\Delta^{\off}_i}{\gamma_{M^\off}}\right)^{-2}, 
\qquad 
i \in [A].
\label{eq_han2024_2}
\end{align}
represent the target number of pulls for each arm. Arms with smaller gaps $\Delta^{\off}_i$ receive larger $n^*_{i;M^\off}$, reflecting increased exploration. \Cref{lemma_bandit_han2024} shows that the random variables $n_{i;M^\off}$ concentrates around $n^*_{i;M^\off}$ with high probability.
The quantity
$$
    D_* \coloneqq \frac{1}{M^\off} \sum_{i \in [A]} \left( \frac{n^*_{i;M^\off}}{n_*(\Theta)} \right)^{1/2} n^*_{i;M^\off}
$$
is a normalization factor. It is bounded within $[A^{-1/3}, 1]$ (Lemma~6.1 of~\citet{han2024ucb}), ensuring it does not degenerate.
Finally, $\mathrm{err}(\Theta)$ quantifies the estimation error.
It is given by
\begin{align}
\mathrm{err}(\Theta) = \frac{\sqrt{\log (\gamma_{M^\off})} + \sqrt{\log(\log (M^\off))}}{\gamma_{M^\off}} + \frac{A}{M^\off} + \frac{\|\Delta^\off\|_\infty^2}{\gamma_{M^\off}^2}.
\label{eq_han2024_4}
\end{align}

We now state~\Cref{lemma_bandit_han2024} formally.

\begin{lemma}[Proposition~3.7 of~\citet{han2024ucb}]
\label{lemma_bandit_han2024}
For the offline--online bandit model in~\Cref{sec_comparison}, fixing $\gamma \ge 1$, there exists a universal constant $C>0$ such that if
\begin{align}
\mathrm{err}(\Theta) \vee \left(\frac{\gamma}{\gamma_{M^\off}}\right) \le \frac{1}{C},
\label{eq_han2024_3}
\end{align}
then
\begin{align}
\max_{i \in [A]} \Pr\left( \left| \frac{n_{i;M^\off}}{n^*_{i;M^\off}} - 1 \right| \ge C\left(D_*^{-1}\mathrm{err}(\Theta) + \frac{\gamma}{\gamma_{M^\off}}\right) \right) \le C\left(\gamma_{M^\off}^{-100} + \log (M^\off) \cdot \gamma e^{-\gamma^2/2}\right).
\label{eq_han2024_main}
\end{align}
\end{lemma}

Based on~\Cref{lemma_bandit_han2024}, we now establish~\Cref{instance_adaptive_policy}.

\begin{rproofof}{\cref{instance_adaptive_policy}}
    We begin by showing that in our model, the constant $n_*(\Theta) \ge \frac{M^\off}{A}$.
    By~\cref{eq_han2024_1}, for any $i\in[A]$,
    \(
        \left(1 + n_*^{1/2}(\Theta)\frac{\Delta^{\off}_i}{\gamma_{M^\off}}\right)^{-2} \le 1.
    \)
    Thus
    $$
        M^\off = \sum_{i \in [A]}  n_*(\Theta)\left(1 + n_*^{1/2}(\Theta)\frac{\Delta^{\off}_i}{\gamma_{M^\off}}\right)^{-2} \le An_*(\Theta),
    $$
    which leads to
    \(
        n_*(\Theta) \ge \frac{M^\off}{A}.
    \)
    Under the assumption that there exists $p\in(0,\frac{1}{2}]$ such that
    \(
        \Dgap = O((M^{\off})^{-p}),
    \)
    and noting that the right-hand side of~\cref{eq_han2024_2} is non-increasing in $\Delta_i^{\off}$, we obtain that for each arm $i\in[A]$,
    \(
        n^*_{i;M^\off} = \Omega((M^{\off})^{2p}),
    \)
    up to logarithmic factors arising from $\gamma_{M^\off}=\sqrt{2\log (M^\off)}$.  
    
    For the fixed constant $C$ in~\cref{eq_han2024_3}, the condition~\cref{eq_han2024_3} is satisfied by choosing $\gamma=\gamma_{M^\off}/C$, since all three terms in \cref{eq_han2024_4} converge to $0$ as $M^\off\to\infty$.
    Thus by combining~\cref{eq_han2024_3} and~\cref{eq_han2024_main}, we obtain that when $K$ and $M^\off$ are sufficiently large,
    \begin{align}
        \max_{i \in [A]} &\Pr\left( n_{i;M^\off} \le n^*_{i;M^\off} - 2\right) 
            \le C\left(\gamma_{M^\off}^{-100} + \log (M^\off) \cdot \gamma e^{-\gamma^2/2}\right) \nonumber \\ 
            &= C((2\log (M^\off))^{-50} + \log (M^\off) \cdot \frac{\sqrt{2\log (M^\off)}}{C}\cdot (M^{\off})^{-\frac{1}{C^2}}) \le \frac{1}{K\log (M^\off)}, \label{eq_proof_proposition2_main}
    \end{align}
    where the second line follows by substituting $\gamma=\gamma_{M^\off}/C$, and the third line uses the assumption $K=o((\log (M^\off))^{49})$.

    Define the event
    \(
        \E_{B} = \{\forall i \in [A],  n_{i;M^\off} \ge n^*_{i;M^\off} - 2\}.
    \)
    Then by~\cref{eq_proof_proposition2_main},
    \(
    \Pr(\E_B)\ge1-\frac{A}{K\log (M^\off)}.
    \)
    Conditional on $E_B$, using the deterministic construction of $\{\TildeV_{m,h}\}$ in~\cref{discussion_tabularbandit_coverage}, the offline design matrix satisfies
    \begin{align*}
        G_{\off} = \operatorname{diag}\Bigl(\frac{n_{1;M^\off}}{3}, \frac{n_{1;M^\off}}{3}, \frac{n_{1;M^\off}}{3}, \frac{n_{2;M^\off}}{3}, \ldots, \frac{n_{A;M^\off}}{3} \Bigr).
    \end{align*}
    By the definition of $\tau(\delta)$ in~\Cref{assumption_eigen}, we obtain
    \(
        \tau\left(\frac{A}{K\log (M^\off)}\right) \ge \Omega\left(\frac{(M^{\off})^{2p}}{M^\off}\right) = \Omega((M^{\off})^{2p-1}).
    \)

    Note that $\frac{A}{K\log (M^\off)}$ differs from the conventional choice of $\delta$ used throughout this paper, namely $\frac{1}{2KM^\off}$. 
    Nevertheless, returning to the explicit regret upper bound for arbitrary $K$ and $M^\off$ in~\cref{eq_final} (see the proof of~\Cref{thm1}), and choosing $\delta = \frac{1}{2KM^\off}$ and $\delta' = \frac{A}{K\log (M^\off)}$ yields the same guarantee as in~\Cref{thm1}.
The bound in~\Cref{eq_bandit_adaptive} then follows by the same argument as in \Cref{cor_obervation_upper_bound}, completing the proof.
\end{rproofof}

\subsection{Proofs in~\Cref{sec_lower_bound}}\label{proof_lower_bound}

In this section we provide a detailed proof of~\Cref{thm2}. 
\Lowerbound*

\begin{rproof}
We organize the proof in two parts. First, we establish the regret lower bound for an offline--online linear bandit instance (Step~1). Second, we embed this offline--online linear bandit instance into an offline--online linear mixture MDP and complete the proof (Step~2).

\paragraph{Step 1: Regret Lower Bound for Offline--Online Linear Bandit Instance.}
We begin with the offline--online linear bandit instance. Inspired by~\citet{lattimore2020bandit}, for any given dimension $d$ and $\tilde{\Delta}$, and for any fixed $K$ and $M^\off$, we consider a class of offline--online linear bandit instances in which the environment shift is defined under the $\|\cdot\|_\infty$ norm and bounded by $\tilde{\Delta}$. Let $\A = [-1,1]^d$ and $\Theta = \{-c_{\tilde{\Delta}}, c_{\tilde{\Delta}}\}^d \times \{-c_{\tilde{\Delta}}, c_{\tilde{\Delta}}\}^d$, where $c_{\tilde{\Delta}}$ depends on $\tilde{\Delta}, K,$ and $M^\off$. The value of $c_{\tilde{\Delta}}$ will be specified later.
Concretely, for any instance $\theta = (\theta^{\off, *}, \theta^*) \in \Theta$, assume that for all $i \in [d]$,
\begin{align}
    |\theta^{\off, *}_i - \theta^{*}_i| \le \tilde{\Delta}.
    \label{eq_lowerbound_9}
\end{align}
Note that this defines $\tilde{\Delta}$ in terms of the $\ell_\infty$ norm, but this relates to the $\ell_2$ definition in \Cref{assump_theta_gap} up to a $\sqrt{d}$ factor.
Furthermore, for any action $a \in \A$ and instance $\theta \in \Theta$, the reward of the offline bandit is a Bernoulli random variable with mean $\frac{1}{2} + \langle a, \theta^{\off, *} \rangle$, while the reward of the online bandit is a Bernoulli random variable with mean $\frac{1}{2} + \langle a, \theta^{*} \rangle$.

We next establish the lower bound for this instance class $\Theta$:
\begin{lemma}
\label{lowerbound_lemma}
Under any fixed offline behavior policy and for any offline--online algorithm, for any sufficiently large $K$ and $M^{\off}$, there exists an offline--online linear bandit instance $\theta = (\theta^{\off, *}, \theta^*) \in \Theta$ such that, as long as the induced coverage satisfies $\tau = \Theta(1)$, the expected regret is at least
\begin{align}
    \Ex[\Regret(K)] = \Omega\left(\min\left\{d\sqrt{K}, d\sqrt{K}\left(\sqrt{\log\left(1+\frac{K}{\lambda_{\min}(\Ex[G_\off])}\right)}(1 + \tilde{\Delta}\sqrt{\lambda_{\min}(\Ex[G_\off])})\right)\right\}\right).
    \label{eq_lowerbound_part_1}
\end{align}
\end{lemma}

The proof of~\Cref{lowerbound_lemma} largely follows the lower bound construction for linear bandits in~\citet{lattimore2020bandit}. A key difference in our setting is the need to capture the ``min-of-two'' structure in~\cref{eq_lowerbound_part_1}, which requires constructing different hard instances under different regimes where distinct terms dominate the bound. 
A detailed discussion of each regime, along with the full proof of~\Cref{lowerbound_lemma}, is provided in~\Cref{proof_lemma_in_lower_bound}.

\paragraph{Step 2: Additional $H$ Factor by Embedding in a Linear Mixture MDP.} We next prove~\cref{eq_lowerbound_main_2} by embedding the offline--online linear bandit instances constructed via~\Cref{lowerbound_lemma} into an offline--online linear mixture MDP.
Concretely, for any $H$, $\Delta$, and $d \ge 2$, consider the following offline--online linear mixture MDP instance class. The state space is $\S = \{s_1^*, s_2^*, s_3^*\}$ and the action space is $\A = [-1,1]^{d-1}$. The initial state of every episode is $s_1 = s_1^*$. For any action $a \in \A$ and any parameter $\theta = (\theta^{\off,*}, \theta^*) \in \Theta$, where $\Theta$ is the $(d-1)$-dimensional offline--online linear bandit instance class constructed above, we define the transition probabilities as
\begin{align}
    P^{\off}(s^*_2 \mid s^*_1,a) &= \frac{1}{2} + \langle \theta^{\off,*}, a \rangle, 
    \qquad
    P^{\off}(s^*_3 \mid s^*_1,a) = \frac{1}{2} - \langle \theta^{\off,*}, a \rangle, \label{eq_lowerbound_16}\\
    P(s^*_2 \mid s^*_1,a) &= \frac{1}{2} + \langle \theta^{*}, a \rangle, 
    \qquad
    P(s^*_3 \mid s^*_1,a) = \frac{1}{2} - \langle \theta^{*}, a \rangle.\label{eq_lowerbound_17}
\end{align}
Thus at step $h=2$ in both offline environment and online environment, the system transitions to either $s_2^*$ or $s_3^*$. We further set
\begin{align*}
    P^{\off}(s^*_2 \mid s^*_2,a) = 1, \qquad P^{\off}(s^*_3 \mid s^*_3,a) = 1, \qquad
    P(s^*_2 \mid s^*_2,a) = 1, \qquad P(s^*_3 \mid s^*_3,a) = 1, \qquad \forall a \in \A.
\end{align*}
The reward is $1$ at state $s_2^*$ and $0$ at state $s_3^*$.

By construction, the transition from $s_1^*$ determines whether the process enters $s_2^*$ or $s_3^*$, and the obtained reward is then repeated for the remaining $H-1$ steps. 
Therefore this MDP interaction is equivalent to an offline--online linear bandit interaction whose reward is scaled by a factor $H$.
Let $\mathscr{A}_{\mathrm{MDP}}$ denote the class of offline--online algorithms for the linear mixture MDP problem, and $\mathscr{A}_{\mathrm{B}}$ denote the class of algorithms for the offline--online linear bandit problem.
Also let $\Theta$ denote the $(d-1)$-dimensional offline--online linear bandit instance class constructed in the first part of the proof. Based on $\Theta$, consider any offline--online linear mixture MDP instance class $\Theta_0$ that contains the
subset of instances constructed above following~\cref{eq_lowerbound_16} and~\cref{eq_lowerbound_17}.
Any algorithm $\mathscr a \in \mathscr{A}_{\mathrm{MDP}}$ induces an algorithm in $\mathscr{A}_{\mathrm{B}}$ when restricted to this embedded bandit interaction. Consequently,
\begin{align*}
    \inf_{\mathscr{a} \in \mathscr{A}_{\mathrm{MDP}}} \sup_{\theta \in \Theta_0} &\Ex[\Regret(K)] \ge H\cdot\inf_{\mathscr{a} \in \mathscr{A}_{\mathrm{B}}} \sup_{\theta \in \Theta} \Ex[\Regret(K)] \\
    &= \Omega\Biggl(\min\Biggl\{(d-1)H\sqrt{K}, (d-1)H\sqrt{K}\Biggl(\sqrt{\log\left(1+\frac{K}{\lambda_{\min}(\Ex[G_\off])}\right)}(1 + \tilde{\Delta}\sqrt{\lambda_{\min}(\Ex[G_\off])})\Biggr)\Biggr\}\Biggr) \\
    &= \Omega\Biggl(\min\Biggl\{dH\sqrt{K}, dH\sqrt{K}\Biggl(\sqrt{\log\left(1+\frac{K}{\lambda_{\min}(\Ex[G_\off])}\right)}(1 + \frac{\Delta}{\sqrt{d}}\sqrt{\lambda_{\min}(\Ex[G_\off])})\Biggr)\Biggr\}\Biggr),
\end{align*}
where the second line follows from~\cref{eq_lowerbound_part_1} applied to the $(d-1)$-dimensional linear bandit instance class, and the last line uses the relation between the $\ell_\infty$ and $\ell_2$ environment shift parameters. For any $\Delta$ defined under the $\|\cdot\|_2$ norm, there exists an instance with $\tilde{\Delta}=\Delta/\sqrt{d-1}$ under the $\|\cdot\|_\infty$ norm that satisfies the same shift constraint.
\end{rproof}

\subsubsection{Proof of~\Cref{lowerbound_lemma}}
\label{proof_lemma_in_lower_bound}

\begin{rproof}
Note that for any given $\theta^*$, the optimal action is $a^* = [\operatorname{sign}(\theta^*_i)]_{i \in [d]}$, since $\A = [-1,1]^d$.  Now, fix $i \in [d]$ and $\theta = (\theta^{\off,*}, \theta^*) \in \Theta$, and define
\begin{align}
    p_{\theta i} = \Pr_{\theta}\left(\sum_{k = 1}^{K} \mathbf{1}\{\operatorname{sign}(a_{ki}) \neq \operatorname{sign}(\theta^*_{i})\} \ge \frac{K}{2}\right),
    \label{eq_lowerbound_6}
\end{align}
where $a_{ki}$ denotes the $i$-th coordinate of the action $a_k$. 
Thus, for any $\theta \in \Theta$,
\begin{align}
\Ex[\Regret(K)]
&= \mathbb{E}_\theta \left[\sum_{k=1}^{K} \sum_{i=1}^{d} (\operatorname{sign}(\theta^*_i) - a_{ki})\theta^*_i \right] \nonumber \\
&\ge c_{\tilde{\Delta}}\sum_{i=1}^{d} \Ex_\theta \left[ \sum_{k=1}^{K} \mathbf{1}\{\operatorname{sign}(a_{ki}) \neq \operatorname{sign}(\theta^*_i)\}\right] \nonumber \\
&\ge \frac{Kc_{\tilde{\Delta}}}{2}\sum_{i=1}^{d}\Pr_\theta \left(
\sum_{k=1}^{K} \mathbf{1}\{\operatorname{sign}(a_{ki}) \neq \operatorname{sign}(\theta^*_i)\} \ge \frac{K}{2} \right) \nonumber \\
&= \frac{Kc_{\tilde{\Delta}}}{2}\sum_{i=1}^{d} p_{\theta i} \label{eq_lowerbound_critical2},
\end{align}
where the first inequality is from the definition of $\Theta$ and $\A$, the third line follows from Markov's inequality~\citep{lattimore2020bandit}, and the fourth line follows from substituting the definition of $p_{\theta i}$.

Next, we lower bound $\sum_{i=1}^{d} p_{\theta i}$ via a pairing argument.  For each $i \in [d]$, define $\theta'$ by flipping the $i$-th coordinate of $\theta^*$ while keeping all other coordinates fixed, and choose $\theta'^{\off,*} \in \{-c_{\tilde{\Delta}}, c_{\tilde{\Delta}}\}^d$ so that $\theta' \in \Theta$. 
(We suppress the dependence of $\theta'$ on $i$ in the notation.)
Restricting attention to a subset of $\Theta$ that is closed under this mapping, we obtain a one-to-one pairing between $\theta$ and $\theta'$. 
For each such pair, we bound $p_{\theta i} + p_{\theta' i}$, and summing over $i \in [d]$ yields a lower bound on $f(\theta) + f(\theta')$, where $f(\theta) := \sum_{i=1}^d p_{\theta i}$. 
Averaging over all $\theta$ in this subset and using the one-to-one pairing, we conclude that there exists $\theta \in \Theta$ such that $\sum_{i=1}^d p_{\theta i}$ satisfies the desired lower bound.

To bound $p_{\theta i} + p_{\theta' i}$, let $P_{\theta}$ and $P_{\theta'}$ denote the probability measures over outcomes under $\theta$ and $\theta'$, respectively.
Concretely, under any fixed algorithm and environment $\theta = (\theta^{\off, *}, \theta^*)$, let $P_{\theta}$ denote the joint measure over
\[
\Bigl(
\underbrace{(\ldots, a_{M^\off}, r_{M^\off})}_{\text{governed by } \theta^{\off,*}},
\underbrace{(\ldots, a_{K}, r_{K})}_{\text{governed by } \theta^{*}}
\Bigr),
\]
where we slightly abuse notation and use $a_m$ and $a_k$ to denote the actions taken at episode $m$ in the offline phase and episode $k$ in the online phase, respectively. $P_{\theta'}$ is defined analogously.
Let $D_{KL}(P\|Q)$ denote the Kullback--Leibler divergence between two probability measures $P$ and $Q$. 
Suppose that for all $m \in [M^\off]$ and $k \in [K]$,
\begin{align}
    \frac{1}{4} < \frac{1}{2} - |\langle a_m, \theta^{\off, *}\rangle| \le \frac{1}{2} + |\langle a_m, \theta^{\off, *}\rangle| < \frac{3}{4},
    \qquad 
    \frac{1}{4} < \frac{1}{2} - |\langle a_k, \theta^*\rangle| \le \frac{1}{2} + |\langle a_k, \theta^*\rangle| < \frac{3}{4}.\label{eq_lowerbound_15}
\end{align}
Then
\begin{align}
    D_{KL}(P_{\theta}\|P_{\theta'}) &= \Ex_{\theta} \Biggl[\sum_{m = 1}^{M^\off} D_{KL}(B(\frac{1}{2} + \langle a_m, \theta^{\off, *} \rangle) \bigm \| B(\frac{1}{2} + \langle a_m, \theta'^{\off, *} \rangle)) \nonumber \\
    &\qquad + \sum_{k = 1}^{K} D_{KL}(B(\frac{1}{2} + \langle a_k, \theta^{*} \rangle) \bigm \| B(\frac{1}{2} + \langle a_k, \theta'^* \rangle))\Biggr] \nonumber \\
    &= \Ex_{\theta} \Biggl[\sum_{m = 1}^{M^\off} \left(\frac{1}{2} + \langle a_m, \theta^{\off, *} \rangle\right) \log \left(\frac{\frac{1}{2} + \langle a_m, \theta^{\off, *} \rangle}{\frac{1}{2} + \langle a_m, \theta'^{\off, *} \rangle}\right) + \left(\frac{1}{2} - \langle a_m, \theta^{\off, *} \rangle\right) \log \left(\frac{\frac{1}{2} - \langle a_m, \theta^{\off, *} \rangle}{\frac{1}{2} - \langle a_m, \theta'^{\off, *} \rangle}\right) \nonumber \\
    &\qquad + \sum_{k = 1}^{K} \left(\frac{1}{2} + \langle a_k, \theta^* \rangle\right) \log \left(\frac{\frac{1}{2} + \langle a_k, \theta^* \rangle}{\frac{1}{2} + \langle a_k, \theta'^* \rangle}\right) + \left(\frac{1}{2} - \langle a_k, \theta^* \rangle\right) \log \left(\frac{\frac{1}{2} - \langle a_k, \theta^* \rangle}{\frac{1}{2} - \langle a_k, \theta'^* \rangle}\right) \Biggr] \label{eq_lowerbound_5} \\
    &\le \frac{8}{3} \Ex_{\theta} \left[\sum_{m = 1}^{M^\off} \langle a_m, (\theta^{\off, *} - \theta'^{\off, *})\rangle^2 + \sum_{k = 1}^{K} \langle a_k, (\theta^* - \theta'^*)\rangle^2\right] \label{eq_lowerbound_1},
\end{align}
where the first equality follows from the standard decomposition of Kullback--Leibler divergence (see Section~4.7 of~\citet{lattimore2020bandit}), and the second equality follows from the definition of the Kullback--Leibler divergence. The inequality is from the following bound on the Kullback--Leibler divergence between Bernoulli distributions: for two Bernoulli random variables with means $\mu_1$ and $\mu_2$, if
\(
    \frac{1}{4} \le \mu_1 \le \frac{3}{4}
\)
and
\(
    \frac{1}{4} \le \mu_2 \le \frac{3}{4},
\)
then $D_{KL}(B(\mu_1) \| B(\mu_2)) \le \frac{8}{3}(\mu_1 - \mu_2)^2$. We omit the detailed proof.

Recall that $\theta'^*_{j} = \theta^*_{j}$ for $j \neq i$, $\theta'^*_{i} = -\theta^*_{i}$, and $\theta'^{\off,*} \in \{-c_{\tilde{\Delta}}, c_{\tilde{\Delta}}\}^d$. By the Bretagnolle--Huber inequality (Theorem~14.2 of~\citet{lattimore2020bandit}) and~\cref{eq_lowerbound_1}, 
\begin{align}
    p_{\theta i} + p_{\theta' i} \ge \frac{1}{2} \exp\left(-\frac{8}{3} \Ex_{\theta} \left[\sum_{m = 1}^{M^\off} \langle a_m, (\theta^{\off, *} - \theta'^{\off, *})\rangle^2 + \sum_{k = 1}^{K} \langle a_k, (\theta^* - \theta'^*)\rangle^2\right]\right).
    \label{eq_lowerbound_critical1}
\end{align}

To complete the proof of~\Cref{lowerbound_lemma}, we consider the following four cases.
\begin{itemize}
    \item {\bf Case 1}: $d\sqrt{K} \ge d\sqrt{K}\left(\sqrt{\log\left(1+\frac{K}{\lambda_{\min}(\Ex[G_\off])}\right)}\left(1 + \tilde{\Delta}\sqrt{\lambda_{\min}(\Ex[G_\off])}\right)\right)$ and $\tilde{\Delta}\sqrt{\lambda_{\min}(\Ex[G_\off])} < 1$.

    \item {\bf Case 2}: $d\sqrt{K} \ge d\sqrt{K}\left(\sqrt{\log\left(1+\frac{K}{\lambda_{\min}(\Ex[G_\off])}\right)}\left(1 + \tilde{\Delta}\sqrt{\lambda_{\min}(\Ex[G_\off])}\right)\right)$ and $\tilde{\Delta}\sqrt{\lambda_{\min}(\Ex[G_\off])} \ge 1$. 

    \item {\bf Case 3}: $d\sqrt{K} < d\sqrt{K}\left(\sqrt{\log\left(1+\frac{K}{\lambda_{\min}(\Ex[G_\off])}\right)}\left(1 + \tilde{\Delta}\sqrt{\lambda_{\min}(\Ex[G_\off])}\right)\right)$ and $\sqrt{\log\left(1+\frac{K}{\lambda_{\min}(\Ex[G_\off])}\right)} \ge \frac{1}{2}$.

    \item {\bf Case 4}: $d\sqrt{K} < d\sqrt{K}\left(\sqrt{\log\left(1+\frac{K}{\lambda_{\min}(\Ex[G_\off])}\right)}\left(1 + \tilde{\Delta}\sqrt{\lambda_{\min}(\Ex[G_\off])}\right)\right)$ and $\sqrt{\log\left(1+\frac{K}{\lambda_{\min}(\Ex[G_\off])}\right)} < \frac{1}{2}$.
\end{itemize}
Cases~1 and~2 correspond to regimes where the offline--online term dominates in~\cref{eq_lowerbound_part_1}. Cases 3 and 4 correspond to regimes where the online-only term dominates.
For each case, we verify that (i) the condition in~\cref{eq_lowerbound_15} holds, and (ii) a lower bound on $\sum_{i=1}^{d} p_{\theta i}$ can be established via~\cref{eq_lowerbound_critical1}. The desired lower bound on the expected regret then follows from~\cref{eq_lowerbound_critical2}.

{\bf Case 1}: $d\sqrt{K} \ge d\sqrt{K}\left(\sqrt{\log\left(1+\frac{K}{\lambda_{\min}(\Ex[G_\off])}\right)}\left(1 + \tilde{\Delta}\sqrt{\lambda_{\min}(\Ex[G_\off])}\right)\right)$ and $\tilde{\Delta}\sqrt{\lambda_{\min}(\Ex[G_\off])} < 1$. 
In this case, we set $c_{\tilde{\Delta}} = \frac{1}{\sqrt{K + M^\off}}$.
We focus on a subset of $\Theta$ where the offline environment is identical to the online environment.
Denote
\begin{align}
    \Theta_1 = \{\theta \in \Theta\ \mid \theta^{\off, *} = \theta^*\}.
    \label{eq_lowerbound_10}
\end{align}
As a consequence, the environment shift condition~\cref{eq_lowerbound_9} is automatically satisfied.

We first verify the condition in~\cref{eq_lowerbound_15}.
In our setting, for any actions $a_m$ and $a_k$ and any $\theta \in \Theta_1$,
$$
    |\langle a_m, \theta^{\off, *}\rangle| \le \frac{d}{\sqrt{K + M^\off}},
    \qquad 
    |\langle a_k, \theta^*\rangle| \le \frac{d}{\sqrt{K + M^\off}}.
$$
Therefore, for sufficiently large $M^\off$ and $K$ satisfying $K+M^\off \ge 16d^2$,
\begin{align}
    \frac{1}{4} < \frac{1}{2} - |\langle a_m, \theta^{\off, *}\rangle| \le \frac{1}{2} + |\langle a_m, \theta^{\off, *}\rangle| < \frac{3}{4},
    \qquad 
    \frac{1}{4} < \frac{1}{2} - |\langle a_k, \theta^*\rangle| \le \frac{1}{2} + |\langle a_k, \theta^*\rangle| < \frac{3}{4}.
\end{align}
Thus~\cref{eq_lowerbound_15} holds.

For any $i \in [d]$, by~\cref{eq_lowerbound_critical1},
\begin{align}
    p_{\theta i} + p_{\theta' i} \ge \frac{1}{2} \exp\left(-\frac{8}{3} \Ex_{\theta} \left[\sum_{m = 1}^{M^\off} \langle a_m, (\theta^* - \theta'^*)\rangle^2 + \sum_{k = 1}^{K} \langle a_k, (\theta^* - \theta'^*)\rangle^2\right]\right) \ge \frac{1}{2} \exp\left(-\frac{32}{3}\right),
    \label{eq_lowerbound_7}
\end{align}
where the second inequality follows from $c_{\tilde{\Delta}} = \frac{1}{\sqrt{K + M^\off}}$ and $\A = [-1,1]^d$.
Summing over $\theta \in \Theta_1$ and $i \in [d]$, we obtain
\begin{align}
\sum_{\theta \in \Theta_1} \frac{1}{|\Theta_1|} \sum_{i=1}^{d} p_{\theta i} = \frac{1}{|\Theta_1|} \sum_{i=1}^{d} \sum_{\theta \in \Theta_1} p_{\theta i} \ge \frac{d}{4} \exp\left(-\frac{32}{3}\right).
\label{eq_lowerbound_8}
\end{align}
Therefore, there exists $\theta \in \Theta_1$ such that $\sum_{i = 1}^d p_{\theta i} \ge \frac{d}{4} \exp\left(-\frac{32}{3}\right)$.
For this specific choice of $\theta$, following~\cref{eq_lowerbound_critical2},
\begin{align}
\Ex[\Regret(K)]
\ge \frac{Kc_{\tilde{\Delta}}}{2}\sum_{i=1}^{d} p_{\theta i} = \frac{K}{2\sqrt{K + M^\off}}\sum_{i=1}^{d} p_{\theta i} \ge \frac{d}{8} \exp\left(-\frac{32}{3}\right)\frac{K}{\sqrt{K + M^\off}}\label{eq_lowerbound_2},
\end{align}
where the equality follows from substituting the definition of $c_{\tilde{\Delta}}$.

Under the condition $\tau = \Theta(1)$, for any constant $H$, there exists a constant $c>0$ such that for sufficiently large $M^\off$,
$$
    \Pr(\lambda_{\min}(G_{\off}) \ge cM^\off - H^2d) \ge 1 - \frac{1}{2KM^\off} \Rightarrow \Pr(\lambda_{\min}(G_{\off}) \ge \frac{c}{2}M^\off) \ge 1 - \frac{1}{2KM^\off}.
$$
Note that $G_\off$ is positive semidefinite, and therefore $\lambda_{\min}(G_\off) \ge 0$ a.s.. Furthermore,
\begin{align}
    \lambda_{\min}(\Ex[G_\off]) \ge \Ex[\lambda_{\min}(G_\off)] \ge \frac{c}{2}M^\off\left(1 - \frac{1}{2KM^\off}\right)
    \label{eq_lowerbound_3},
\end{align}
where the first inequality follows from~\cref{lemma_lambda_min_concavitiy}, and the second inequality follows from the above high-probability bound together with $\lambda_{\min}(G_\off) \ge 0$ a.s..

Note that the assumption in Case~1, $d\sqrt{K} \ge d\sqrt{K}\left(\sqrt{\log\left(1+\frac{K}{\lambda_{\min}(\Ex[G_\off])}\right)}\left(1 + \tilde{\Delta}\sqrt{\lambda_{\min}(\Ex[G_\off])}\right)\right)$ and $\tilde{\Delta}\sqrt{\lambda_{\min}(\Ex[G_\off])} < 1$, implies that 
\begin{align}
    \sqrt{\log\left(1+\frac{K}{\lambda_{\min}(\Ex[G_\off])}\right)} \le 1 \Rightarrow K \le (e - 1)\lambda_{\min}(\Ex[G_\off]),
    \label{eq_lowerbound_4}
\end{align}
since $1 + \tilde{\Delta}\sqrt{\lambda_{\min}(\Ex[G_\off])} \ge 1$ holds trivially.
Combining~\cref{eq_lowerbound_2} and~\cref{eq_lowerbound_3}, 
\begin{align}
    \Ex[\Regret(K)] 
    &\ge \frac{d}{8} \exp\left(-\frac{32}{3}\right)\frac{K}{\sqrt{K + \frac{2\lambda_{\min}(\Ex[G_\off])}{c\left(1 - \frac{1}{2KM^\off}\right)}}} = \Omega\left(d\frac{K}{\sqrt{K + \lambda_{\min}(\Ex[G_\off])}}\right) \label{eq_lowerbound_11}\\
    &= \Omega\left(d\sqrt{K}\sqrt{\frac{K}{\lambda_{\min}(\Ex[G_\off])}}\right) = \Omega\left(d\sqrt{K}\sqrt{\log\left(1 + \frac{K}{\lambda_{\min}(\Ex[G_\off])}\right)}\right) \nonumber\\
    &= \Omega\left(\min\left\{d\sqrt{K}, d\sqrt{K}\left(\sqrt{\log\left(1+\frac{K}{\lambda_{\min}(\Ex[G_\off])}\right)}\left(1 + \tilde{\Delta}\sqrt{\lambda_{\min}(\Ex[G_\off])}\right)\right)\right\}\right),\nonumber
\end{align}
where the second line uses~\cref{eq_lowerbound_4}, which holds by the condition defining Case~1.
The third equality uses the inequality $x \ge \log(1+x)$ for all $x > -1$, and the last line follows from the assumption $\tilde{\Delta}\sqrt{\lambda_{\min}(\Ex[G_\off])} < 1$.

{\bf Case 2}: $d\sqrt{K} \ge d\sqrt{K}\left(\sqrt{\log\left(1+\frac{K}{\lambda_{\min}(\Ex[G_\off])}\right)}\left(1 + \tilde{\Delta}\sqrt{\lambda_{\min}(\Ex[G_\off])}\right)\right)$ and $\tilde{\Delta}\sqrt{\lambda_{\min}(\Ex[G_\off])} \ge 1$. 
The instance class setting is similar to that in Case~1, with the only difference being the choice of $c_{\tilde{\Delta}}$ and the environment shift. In this case, we set
\(
    c_{\tilde{\Delta}} = \frac{\tilde{\Delta}}{2}.
\)
Thus for any $\theta = (\theta^{\off, *}, \theta^*) \in \Theta = \{-c_{\tilde{\Delta}}, c_{\tilde{\Delta}}\}^d \times \{-c_{\tilde{\Delta}}, c_{\tilde{\Delta}}\}^d$, 
\(
    \|\theta^{\off, *} - \theta^*\|_\infty \le \tilde{\Delta}.
\)

Denote 
\begin{align}
    \Theta_2 = \{\theta = (\theta^{\off, *}, \theta^*) \in \Theta \mid \theta^{\off, *}_i = c_{\tilde{\Delta}}, \forall i \in [d]\},
    \label{eq_lowerbound_12}
\end{align}
that is, the offline environment is identical for all $\theta \in \Theta_2$. 

The condition defining Case~2,
\(
    d\sqrt{K} \ge d\sqrt{K}\left(\sqrt{\log\left(1+\frac{K}{\lambda_{\min}(\Ex[G_\off])}\right)}\left(1 + \tilde{\Delta}\sqrt{\lambda_{\min}(\Ex[G_\off])}\right)\right),
\)
implies
\begin{align}
    \sqrt{\log\left(1+\frac{K}{\lambda_{\min}(\Ex[G_\off])}\right)}\tilde{\Delta}\sqrt{\lambda_{\min}(\Ex[G_\off])} \le 1 \Rightarrow \tilde{\Delta} \le \frac{1}{\sqrt{\lambda_{\min}(\Ex[G_\off]) \log \left(1 + \frac{K}{\lambda_{\min}(\Ex[G_\off])}\right)}}. \label{eq_lowerbound_18}
\end{align}

In turn,~\cref{eq_lowerbound_18} ensures that the condition in~\cref{eq_lowerbound_15} is satisfied when $K \ge \frac{32}{\log (3)}d^2$, as
\begin{align}
    \max_{m,k} \{|\langle a_m, \theta^{\off, *}\rangle|, |\langle a_k, \theta^*\rangle|\} \le c_{\tilde{\Delta}}d = \frac{\tilde{\Delta}}{2}d \le \frac{d}{2}\frac{1}{\sqrt{\lambda_{\min}(\Ex[G_\off]) \log \left(1 + \frac{K}{\lambda_{\min}(\Ex[G_\off])}\right)}} \le \frac{d}{\sqrt{\frac{\log (3)}{2}K}} \le \frac{1}{4},
    \label{eq_lowerbound_19}
\end{align}
where the first inequality is from the definition of $\Theta$ and $\A$, the equality is from $c_{\tilde{\Delta}} = \frac{\tilde{\Delta}}{2}$, the second inequality is from~\cref{eq_lowerbound_18}, and the third inequality is from the bound $x/\log(1+x) \le 2/\log (3)$ for all $x \le 2$, together with $\sqrt{\log\left(1+\frac{K}{\lambda_{\min}(\Ex[G_\off])}\right)} \le 1$ (similar to~\cref{eq_lowerbound_4}), which implies $\frac{K}{\lambda_{\min}(\Ex[G_\off])} \le e-1 \le 2$.

For any $i \in [d]$, by~\cref{eq_lowerbound_critical1}, 
\begin{align*}
    p_{\theta i} + p_{\theta' i} 
    &\ge \frac{1}{2} \exp\left(-\frac{8}{3} \Ex_{\theta} \left[\sum_{k = 1}^{K} \langle a_k, (\theta^* - \theta'^*)\rangle^2\right]\right) \ge \frac{1}{2} \exp\left(-\frac{8}{3}\tilde{\Delta}^2K\right) \\
    &\ge \frac{1}{2} \exp\left(-\frac{8}{3}\frac{K}{\lambda_{\min}(\Ex[G_\off])\log\left(1 + \frac{K}{\lambda_{\min}(\Ex[G_\off])}\right)}\right) \\
    &\ge \frac{1}{2} \exp\left(-\frac{8}{3}\frac{2}{\log(3)}\right) = \frac{1}{2} \exp\left(-\frac{16}{3\log(3)}\right),
\end{align*}
where the first line follows from the definition of $\Theta_2$ together with $\A=[-1,1]^d$. 
The second line follows from~\cref{eq_lowerbound_18}.
The third line again follows from $\sqrt{\log\left(1+\frac{K}{\lambda_{\min}(\Ex[G_\off])}\right)} \le 1$, as in~\cref{eq_lowerbound_19}.

Thus, similar to~\cref{eq_lowerbound_8}, there exists $\theta \in \Theta_2$ such that $\sum_{i = 1}^d p_{\theta i} \ge \frac{d}{4} \exp\left(-\frac{16}{3\log(3)}\right)$. For this choice of $\theta$, following~\cref{eq_lowerbound_critical2}, we obtain
\begin{align}
\Ex[\Regret(K)]
&\ge \frac{Kc_{\tilde{\Delta}}}{2}\sum_{i=1}^{d} p_{\theta i} = \frac{K\tilde{\Delta}}{4}\sum_{i=1}^{d} p_{\theta i} \ge \exp\left(-\frac{16}{3\log(3)}\right) \cdot \frac{K\tilde{\Delta} d}{16} \nonumber\\
&= \Omega\left(d\sqrt{K}\sqrt{\frac{K}{\lambda_{\min}(\Ex[G_\off])}}(1 + \tilde{\Delta}\sqrt{\lambda_{\min}(\Ex[G_\off])})\right) \nonumber\\
&= \Omega\left(\min\left\{d\sqrt{K}, d\sqrt{K}\left(\sqrt{\log\left(1+\frac{K}{\lambda_{\min}(\Ex[G_\off])}\right)}\left(1 + \tilde{\Delta}\sqrt{\lambda_{\min}(\Ex[G_\off])}\right)\right)\right\}\right)
\label{eq_lowerbound_13},
\end{align}
where the second line follows from the assumption in Case~2 that $\tilde{\Delta}\sqrt{\lambda_{\min}(\Ex[G_\off])} \ge 1$, and the third line also follows from the assumption in Case~2 that
$d\sqrt{K} \ge d\sqrt{K}\left(\sqrt{\log\left(1+\frac{K}{\lambda_{\min}(\Ex[G_\off])}\right)}\left(1 + \tilde{\Delta}\sqrt{\lambda_{\min}(\Ex[G_\off])}\right)\right)$, together with the inequality $x \ge \log(1+x)$ for all $x>0$.

{\bf Case 3}: $d\sqrt{K} < d\sqrt{K}\left(\sqrt{\log\left(1+\frac{K}{\lambda_{\min}(\Ex[G_\off])}\right)}\left(1 + \tilde{\Delta}\sqrt{\lambda_{\min}(\Ex[G_\off])}\right)\right)$ and $\sqrt{\log\left(1+\frac{K}{\lambda_{\min}(\Ex[G_\off])}\right)} \ge \frac{1}{2}$.
We consider the same setting as in Case~1. In particular, we set $c_{\tilde{\Delta}} = \frac{1}{\sqrt{K + M^\off}}$ and focus on the instance class $\Theta_1$ defined in~\cref{eq_lowerbound_10}. Following the argument in Case~1,~\cref{eq_lowerbound_15} holds when $K+M^\off \ge 16d^2$, and there exists $\theta \in \Theta_1$ such that
\begin{align*}
\Ex[\Regret(K)] = \Omega\left(d\frac{K}{\sqrt{K + \lambda_{\min}(\Ex[G_\off])}}\right),
\end{align*}
which follows from~\cref{eq_lowerbound_11}.

Under the condition defining Case~3,
\(
    \sqrt{\log\left(1+\frac{K}{\lambda_{\min}(\Ex[G_\off])}\right)} \ge \frac{1}{2},
\)
which implies
\(
    K \ge (e^{\frac{1}{4}} -1 )\lambda_{\min}(\Ex[G_\off]).
\)
Consequently,
\begin{align*}
\Ex[\Regret(K)] &= \Omega\left(d\frac{K}{\sqrt{K + \lambda_{\min}(\Ex[G_\off])}}\right) = \Omega\left(d\sqrt{K}\right) \\
&= \Omega\left(\min\left\{d\sqrt{K}, d\sqrt{K}\left(\sqrt{\log\left(1+\frac{K}{\lambda_{\min}(\Ex[G_\off])}\right)}\left(1 + \tilde{\Delta}\sqrt{\lambda_{\min}(\Ex[G_\off])}\right)\right)\right\}\right),
\end{align*}
where the second line follows from the condition defining Case~3 that 
\(
d\sqrt{K} < d\sqrt{K}\left(\sqrt{\log\left(1+\frac{K}{\lambda_{\min}(\Ex[G_\off])}\right)}\left(1 + \tilde{\Delta}\sqrt{\lambda_{\min}(\Ex[G_\off])}\right)\right).
\)

{\bf Case 4}: $d\sqrt{K} < d\sqrt{K}\left(\sqrt{\log\left(1+\frac{K}{\lambda_{\min}(\Ex[G_\off])}\right)}\left(1 + \tilde{\Delta}\sqrt{\lambda_{\min}(\Ex[G_\off])}\right)\right)$ and $\sqrt{\log\left(1+\frac{K}{\lambda_{\min}(\Ex[G_\off])}\right)} < \frac{1}{2}$.
We consider the same setting as in Case~2 with a different choice of $c_{\tilde{\Delta}}$, since the condition defining Case~4 implicitly restricts the choice of $\tilde{\Delta}$ in a different way.
Note that the condition defining Case~4,
$$
    d\sqrt{K} < d\sqrt{K}\left(\sqrt{\log\left(1+\frac{K}{\lambda_{\min}(\Ex[G_\off])}\right)}\left(1 + \tilde{\Delta}\sqrt{\lambda_{\min}(\Ex[G_\off])}\right)\right)
$$
implies that
$$
    \sqrt{\log\left(1+\frac{K}{\lambda_{\min}(\Ex[G_\off])}\right)} + \sqrt{\log\left(1+\frac{K}{\lambda_{\min}(\Ex[G_\off])}\right)}\tilde{\Delta}\sqrt{\lambda_{\min}(\Ex[G_\off])} > 1.
$$
Since $\sqrt{\log\left(1+\frac{K}{\lambda_{\min}(\Ex[G_\off])}\right)} < \frac{1}{2}$ (also see the condition defining Case~4), we obtain 
$$
    \tilde{\Delta}\sqrt{\log\left(1+\frac{K}{\lambda_{\min}(\Ex[G_\off])}\right)}\sqrt{\lambda_{\min}(\Ex[G_\off])} > \frac{1}{2}.
$$
Thus
$$
    \tilde{\Delta}\sqrt{K} = \tilde{\Delta}\sqrt{\lambda_{\min}(\Ex[G_\off])}\sqrt{\frac{K}{\lambda_{\min}(\Ex[G_\off])}} \ge \tilde{\Delta}\sqrt{\lambda_{\min}(\Ex[G_\off])}\sqrt{\log\left(1 + \frac{K}{\lambda_{\min}(\Ex[G_\off])}\right)} > \frac{1}{2},
$$
where the first inequality follows from $x \ge \log(1+x)$ for all $x>0$.
Therefore, by choosing
\begin{align}
    c_{\tilde{\Delta}} = \frac{1}{4\sqrt{K}},
    \label{eq_lowerbound_14}
\end{align}
for any $\theta = (\theta^{\off, *}, \theta^*) \in \Theta = \{-c_{\tilde{\Delta}}, c_{\tilde{\Delta}}\}^d \times \{-c_{\tilde{\Delta}}, c_{\tilde{\Delta}}\}^d$, 
\(
    \|\theta^{\off, *} - \theta^*\|_\infty \le \tilde{\Delta}.
\)
This allows us to consider $\Theta_2$ as defined in~\cref{eq_lowerbound_12}.

For any $i \in [d]$, following~\cref{eq_lowerbound_critical1}, and noting that~\cref{eq_lowerbound_14} guarantees that the condition~\cref{eq_lowerbound_15} holds when $K \ge d^2$, we obtain
\begin{align*}
    p_{\theta i} + p_{\theta' i} 
    &\ge \frac{1}{2} \exp\left(-\frac{8}{3} \Ex_{\theta} \left[\sum_{k = 1}^{K} \langle a_k, (\theta^* - \theta'^*)\rangle^2\right]\right) \ge \frac{1}{2} \exp\left(-\frac{2}{3}\right). 
\end{align*}
Similar to~\cref{eq_lowerbound_8}, there exists $\theta \in \Theta_2$ such that $\sum_{i = 1}^d p_{\theta i} \ge \frac{d}{4} \exp\left(-\frac{2}{3}\right)$. For this choice of $\theta$, following~\cref{eq_lowerbound_critical2}, 
\begin{align*}
\Ex[\Regret(K)]
&\ge \frac{Kc_{\tilde{\Delta}}}{2}\sum_{i=1}^{d} p_{\theta i} = \Omega(d\sqrt{K}) \\
&= \Omega\left(\min\left\{d\sqrt{K}, d\sqrt{K}\left(\sqrt{\log\left(1+\frac{K}{\lambda_{\min}(\Ex[G_\off])}\right)}\left(1 + \tilde{\Delta}\sqrt{\lambda_{\min}(\Ex[G_\off])}\right)\right)\right\}\right),
\end{align*}
where the second line follows from~\cref{eq_lowerbound_14}, and the last line follows from the condition defining Case~4 that
\(
    d\sqrt{K} < d\sqrt{K}\left(\sqrt{\log\left(1+\frac{K}{\lambda_{\min}(\Ex[G_\off])}\right)}\left(1 + \tilde{\Delta}\sqrt{\lambda_{\min}(\Ex[G_\off])}\right)\right).
\)
\end{rproof}

\section{Useful Facts}
\label{sec:useful_facts}
In this section we collect several auxiliary results that are used throughout the proofs in~\Cref{sec_appendix_proof}. These include self-normalized concentration inequalities adopted from the literature, a matrix Chernoff bound, as well as basic algebraic inequalities that are invoked repeatedly. We begin with the self-normalized concentration inequality of~\citet{abbasi2011improved}.

\begin{lemma}[Theorem 1 of~\citet{abbasi2011improved}]\label{lemma_concentration}
Let $\{F_t\}_{t=0}^\infty$ be a filtration. Let $\{\eta_t\}_{t=1}^\infty$ be a real-valued stochastic process such that $\eta_t$ is $F_t$-measurable and $\eta_t$ is conditionally $R$-sub-Gaussian for some $R \ge 0$. Let $\{x_t\}_{t=1}^\infty$ be an $\mathbb{R}^d$-valued stochastic process such that $x_t$ is $F_{t-1}$-measurable. Assume that $W$ is a $d \times d$ positive definite matrix. For any $t \ge 0$, define
\[
\bar W_t = W + \sum_{s=1}^t x_s x_s^\top,
\qquad
S_t = \sum_{s=1}^t \eta_s x_s .
\]
Then, for any $\delta > 0$, with probability at least $1-\delta$, for all $t \ge 0$,
\[
\|S_t\|_{\bar W_t^{-1}}^2 \le 2 R^2 \log\left( \frac{\det(\bar W_t)^{1/2} \det(W)^{-1/2}}{\delta}\right).
\]
\end{lemma}

We refer the reader to~\citet{abbasi2011improved} for a detailed proof of~\Cref{lemma_concentration}. Our analysis also relies on a matrix concentration inequality, for which we invoke standard results from~\citet{tropp2012user}.

\begin{lemma}[Corollay 5.2 of~\citet{tropp2012user}]\label{lem_matrix_concentration}
Consider a finite sequence $\{X_k\}$ of independent, random, self-adjoint matrices that satisfy $ X_k \succeq 0$ and $\lambda_{\max}(X_k) \le R$ a.s.. Compute the minimum eigenvalue of the sum of expectations $\mu_{\min} = \lambda_{\min}\left(\sum_k \mathbb{E} X_k \right)$. Then for any $\delta \in (0,1)$,
\begin{align}
\Pr\left(\lambda_{\min}\left(\sum_k X_k\right) \le (1-\delta)\mu_{\min}\right) \le d \cdot
\left[\frac{e^{-\delta}}{(1-\delta)^{1-\delta}}\right]^{\mu_{\min}/R}. \label{eq_matrix_concentration_1}
\end{align}
where $X_k \in \mathbb{R}^{d \times d}$ for any $k$. Furthermore, the bound above admits the following simplified form:
\begin{align}
\Pr\left(\lambda_{\min}\left(\sum_k X_k\right) \le (1-\delta)\mu_{\min}\right) \le de^{-\frac{\delta^2}{2}\frac{\mu_{\min}}{R}}. \label{eq_matrix_concentration_2}
\end{align}
\end{lemma}

\begin{rproof}
We refer the reader to~\citet{tropp2012user} for the proof of the first part of~\Cref{lem_matrix_concentration}, namely~\cref{eq_matrix_concentration_1}. Building on this result, we now prove~\cref{eq_matrix_concentration_2}. Note that from~\cref{eq_matrix_concentration_1}, it suffices to show that for every $\delta\in(0,1)$,
\(
\frac{e^{-\delta}}{(1-\delta)^{1-\delta}} \le \exp\left(-\frac{\delta^2}{2}\right).
\)
Taking logarithms, this is equivalent to 
$
-\delta-(1-\delta)\log(1-\delta)\le -\frac{\delta^2}{2},
$
or, equivalently,
\(
g(\delta):=-\delta-(1-\delta)\log(1-\delta)+\frac{\delta^2}{2}\le 0.
\)
We verify that $g(\delta)\le 0$ on $(0,1)$ using concavity. A direct calculation gives that for any $\delta \in (0,1)$,
\(
g'(\delta)=\log(1-\delta)+\delta
\)
and
\(
g''(\delta)=-\frac{\delta}{1-\delta}\le 0.
\)
Hence $g$ is concave on $(0,1)$, so its derivative $g'$ is nonincreasing. Moreover, since $g'(0) = 0$, we have that for every $\delta\in(0,1)$, $g'(\delta)\le g'(0)=0$, which implies that $g$ is nonincreasing on $(0,1)$. Since $g(0)=0$, we conclude that
\(
g(\delta)\le g(0)=0 
\)
for all $\delta\in(0,1)$.
So far the proof is complete.
\end{rproof}

Next, we present several elementary lemmas concerning algebraic properties of the weighted norm $\|\cdot\|_A$, where $A$ is a symmetric and positive definite matrix.

\begin{lemma}\label{lem_monotone_norm}
Let $A \in \mathbb{R}^{d\times d}$ be symmetric and positive definite, and let
\(
B = A + \sum_{i} x_i x_i^\top
\)
for some vectors $\{x_i\} \subset \mathbb{R}^d$. Then for any $y \in \mathbb{R}^d$,
\(
\|y\|_{A} \le \|y\|_{B}.
\)
\end{lemma}
\begin{rproof}
Note that
\(
y^\top B y = y^\top A y + \sum_i (x_i^\top y)^2 \ge y^\top A y.
\)
Taking square roots on both sides yields $\|y\|_B \ge \|y\|_A$, as claimed.
\end{rproof}

\begin{lemma}\label{lem_inverse_order}
Let $A,B \in \mathbb{R}^{d\times d}$ be symmetric and positive definite matrices. If
\(
A \succeq B \succ 0,
\)
that is, $A-B$ is positive semidefinite, then
\(
B^{-1} \succeq A^{-1}.
\)
\end{lemma}

\begin{rproof}
Since $A \succeq B \succ 0$, we can write
\(
A = B^{1/2}\left(I + C\right)B^{1/2},
\)
where $C \succeq 0$. Taking inverses gives
\(
A^{-1} = B^{-1/2}\bigl(I + C\bigr)^{-1}B^{-1/2}.
\)
Because $C \succeq 0$, all eigenvalues of $I+C$ are at least $1$, hence
\(
(I+C)^{-1} \preceq I.
\)
Therefore,
\(
A^{-1} \preceq B^{-1/2} I B^{-1/2} = B^{-1},
\)
which proves the claim.
\end{rproof}

\begin{lemma}\label{lem_weighted_to_l2}
Let $A\in\mathbb{R}^{d\times d}$ be symmetric and positive definite. Then for any $x\in\mathbb{R}^d$,
\(
\|x\|_{A} \le \sqrt{\lambda_{\max}(A)}\|x\|_2.
\)
\end{lemma}

\begin{rproof}
Since $A$ is symmetric positive definite, it admits an eigendecomposition $A = U \Lambda U^\top$, where $U$ is orthogonal and $\Lambda=\mathrm{diag}(\lambda_1,\ldots,\lambda_d)$ with $0<\lambda_i\le \lambda_{\max}(A)$ for all $i$. Let $z = U^\top x$. Then $\|z\|_2=\|x\|_2$ and
\[
\|x\|_A^2 = x^\top A x = x^\top U\Lambda U^\top x = z^\top \Lambda z = \sum_{i=1}^d \lambda_i z_i^2
\le \lambda_{\max}(A)\sum_{i=1}^d z_i^2 = \lambda_{\max}(A)\|z\|_2^2 = \lambda_{\max}(A)\|x\|_2^2.
\]
Taking square roots on both sides yields the claim.
\end{rproof}

\begin{lemma}\label{lemma_pd}
Let $A\in\mathbb{R}^{d\times d}$ be symmetric positive definite and $B\in\mathbb{R}^{d\times d}$ be symmetric positive definite. Then
\[
\log(\det(A+B))-\log(\det(A)) \le d\log\left(1+\frac{1}{d}\operatorname{tr}(A^{-1}B)\right).
\]
\end{lemma}

\begin{rproof}
We write
\(
\log(\det(A+B))-\log(\det(A)) = \log\det\left(I+A^{-1/2}BA^{-1/2}\right).
\)
Let $\widetilde B:=A^{-1/2}BA^{-1/2}$. Since $B\succeq 0$ and $A^{-1/2}$ is symmetric, $\widetilde B$ is symmetric PSD. Let $\nu_1,\dots,\nu_d\ge 0$ be the eigenvalues of $\widetilde B$. Then
\(
\log\det(I+\widetilde B)=\sum_{i=1}^d \log(1+\nu_i).
\)
Thus,
\[
\prod_{i=1}^d (1+\nu_i) \le \left(\frac{1}{d}\sum_{i=1}^d (1+\nu_i)\right)^d,
\]
hence
\(
\log(\det(I+\widetilde B)) \le d\log\left(1+\frac{1}{d}\sum_{i=1}^d \nu_i\right) = d\log\left(1+\frac{1}{d}\operatorname{tr}(\widetilde B)\right).
\)
Finally, the proof is complete with
\(
\operatorname{tr}(\widetilde B)=\operatorname{tr}(A^{-1/2}BA^{-1/2})=\operatorname{tr}(A^{-1}B).
\)
\end{rproof}

\begin{lemma}\label{lem_trace_bound}
Let $A \in \mathbb R^{d\times d}$ be symmetric positive definite and let $B \in \mathbb R^{d\times d}$ be symmetric positive semi-definite. Then
\(
\operatorname{tr}(AB)\le \lambda_{\max}(A)\operatorname{tr}(B).
\)
\end{lemma}

\begin{rproof}
Since $A\succ 0$ is symmetric, it admits an eigendecomposition $A=U\Lambda U^\top$, where $U$ is orthogonal and
\(
\Lambda=\diag(\lambda_1,\ldots,\lambda_d)
\)
with $0<\lambda_i\le \lambda_{\max}(A)$ for all $i\in[d]$. By cyclicity of the trace,
\(
\operatorname{tr}(AB) = \operatorname{tr}(U\Lambda U^\top B) = \operatorname{tr}(\Lambda U^\top B U).
\)
Let $\widetilde B \coloneqq U^\top B U$. Since $B\succeq 0$ and $U$ is orthogonal, we have $\widetilde B\succeq 0$, which implies $\widetilde B_{ii}\ge 0$ for all $i$. Therefore,
\[
\operatorname{tr}(AB) = \sum_{i=1}^d \lambda_i \widetilde B_{ii} \le \lambda_{\max}(A)\sum_{i=1}^d \widetilde B_{ii} = \lambda_{\max}(A)\operatorname{tr}(\widetilde B) = \lambda_{\max}(A)\operatorname{tr}(B),
\]
where the last equality again follows from cyclicity of the trace.
\end{rproof}

\begin{lemma}\label{lemma_lambda_min_concavitiy}
Let $A \in \mathbb{R}^{d \times d}$ be a random positive semidefinite matrix. Then
\(
\Ex\left[\lambda_{\min}(A)\right] \le \lambda_{\min}\left(\Ex[A]\right).
\)
\end{lemma}

\begin{rproof}
Recall that for any symmetric matrix $A$,
\(
\lambda_{\min}(A) = \min_{\|x\|_2=1} x^\top A x.
\)
Hence, for any deterministic unit vector $x \in \mathbb{R}^d$,
\(
\lambda_{\min}(A) \le x^\top A x.
\)
Taking expectation on both sides yields
\(
\Ex\left[\lambda_{\min}(A)\right] \le x^\top \Ex[A] x.
\)
Since the above inequality holds for every unit vector $x$, taking the minimum over all $\|x\|_2=1$ gives
\(
\Ex\left[\lambda_{\min}(A)\right] \le \min_{\|x\|_2=1} x^\top \Ex[A] x = \lambda_{\min}\left(\Ex[A]\right).
\)
\end{rproof}



\newpage
\section{Omitted Details in~\Cref{sec_simulations}}
\label{appendix_experiment}

\subsection{Implementation of~\ALG and Baseline Algorithms}

\paragraph{Computational Simplification of~\ALG.}
In line 13 of~\Cref{alg_main}, the algorithm requires solving
\[
\Qhat_{k,h}(s,a) = r(s,a) + \max_{\theta\in\B_k\cap\C_k} \sum_{j=1}^d \theta_j \phi_j(\cdot\mid s,a)^\top \Vhat_{k,h+1},
\]
where $\B_k$ and $\C_k$ are ellipsoids. Since the objective is linear in $\theta$, this problem is a quadratically constrained quadratic program (QCQP)~\citep{bao2011semidefinite} and is tractable.

For computational efficiency in simulations, we adopt the following relaxation:
\[
\Qhat_{k,h}(s,a) = r(s,a) + \min\Bigl\{
\max_{\theta\in\B_k} \sum_{j=1}^d \theta_j \phi_j(\cdot\mid s,a)^\top \Vhat_{k,h+1},
\max_{\theta\in\C_k} \sum_{j=1}^d \theta_j \phi_j(\cdot\mid s,a)^\top \Vhat_{k,h+1}
\Bigr\}.
\]
Following~\citet{jia2020model}, this admits the closed-form expression
\begin{align}
\hat{Q}_{k,h}(s,a) = r(s,a) + (x^{\on}_{k,h})^{\top}\hat{\theta}_{k}^{\on} +
\min\Bigl\{ \beta_k\|x^{\on}_{k,h}\|_{(M^{\on}_k)^{-1}},
\gamma_k\|x^{\on}_{k,h}\|_{(M^{\all}_k)^{-1}}
\Bigr\}.
\label{eq_simplification}
\end{align}
In our implementation of~\ALG, we use~\cref{eq_simplification} for computational simplicity.

\paragraph{Baseline Algorithms.}
In addition to the baselines introduced in the main text, in this section we include comparisons between~\ALG and two tabular MDP algorithms, \DPLSVI and \UCBVI.
\begin{itemize}
    \item \DPLSVI (Algorithm~2 in~\citet{chen2022data}): The offline--online tabular MDP algorithm proposed by~\citet{chen2022data}. In contrast to~\Cref{alg_main}, \DPLSVI can perform worse than a purely online method when environment shift $\Delta$ is large.
    \item \UCBVI~\citep{azar2017minimax}: A standard purely online algorithm for tabular MDPs.
\end{itemize}
The comparison between~\ALG and the two tabular MDP algorithms is presented in~\Cref{fig_appendix}.

The implementations of \UCRL, \DPLSVI, \UCBVI, \ALG(\textsf{Optimal}), and \ALG(\textsf{Pessimistic}) are straightforward. For implementation details of \UCRL, \DPLSVI, \UCBVI, and the offline phase of \ALG(\textsf{Pessimistic}), we refer the reader to \citet{jia2020model}, \citet{chen2022data}, \citet{azar2017minimax}, and \citet{nguyen2023instance}, respectively. Following the spirit of the \COMPLETE algorithm proposed in~\citet{chen2022data}, we present the full description of our \COMPLETE baseline in~\Cref{alg_complete}. \Cref{alg_complete} closely mirrors~\Cref{alg_main}, with only two differences:
(i) it does not employ the ``min-of-two'' safeguard (since assuming no environment shift), and
(ii) the confidence set is defined as
\(
\mathcal B_k' =  \left\{ \theta \in \mathbb R^d : \|\theta - \theta^{\all}_k\|_{M^{\all}_k} \le \beta_k \right\},
\)
also due to the fact that we assume no environment shift. Recall that the confidence radius $\beta_k$ is defined in~\Cref{thm1} as
\(
\beta_k = H \sqrt{ 2 \log\left(\frac{2\det(M^{\on}_{k})^{1/2}K}{\det(\lambda I)^{1/2}}\right)} + \sqrt{\lambda}B.
\)
Here we adopt the same deterministic design of $\{\TildeV_{m,h}\}$ as in~\Cref{sec_comparison}, and choose $\lambda = H^2d$. Conceptually, \COMPLETE leverages offline data under the assumption of no environment shift, in the same spirit as classical offline--online approaches for bandits (e.g.,~\citet{shivaswamy2012multi}).

\paragraph{Hyperparameter Tuning.}
For \UCRL, \ALG(\textsf{Optimal}), and the online phase of \ALG(\textsf{Pessimistic}), we set $\lambda$, $\beta_k$, and $\gamma_k$ according to~\Cref{thm1}.  The offline phase of \ALG(\textsf{Pessimistic}) involves an additional uncertainty parameter, which is chosen following Lemma D.1 of~\citet{nguyen2023instance}.  All hyperparameters of \DPLSVI follow their original specification in~\citet{chen2022data}; in particular, we set $\xi_t = 1$ (as noted to be valid in~\citet{chen2022data}).  The bonus term of \UCBVI follows Algorithm 3 of~\citet{azar2017minimax}.

Since the convergence behaviors of different algorithms vary, we introduce an additional bonus scaling parameter $b_s$ to ensure a fair comparison~\citep{burda2018exploration}. For example, in~\ALG, the estimate in~\cref{eq_simplification} becomes
\[
\Qhat_{k,h}(s,a) = r(s,a) + (x^{\on}_{k,h})^{\top}\thetaon_k + b_s \min\left\{ \beta_k\|x^{\on}_{k,h}\|_{(M^{\on}_k)^{-1}}, \gamma_k\|x^{\on}_{k,h}\|_{(M^{\all}_k)^{-1}} \right\}.
\]
The bonus scales for \UCRL, \ALG(\textsf{Optimal}), \ALG(\textsf{Pessimistic}), \DPLSVI, \UCBVI, and \COMPLETE are selected via grid search over $[0,1]$. We use the same $b_s$ for \UCRL, \ALG(\textsf{Optimal}), and \ALG(\textsf{Pessimistic}). 

\begin{algorithm}[h!]
\caption{\COMPLETE}
\label{alg_complete}
\begin{algorithmic}[1]
\REQUIRE Online MDP, dimension $d$, number of online episodes $K$, offline dataset $\mathcal{D}_{\off}$, auxiliary value functions $\{\TildeV_{m,h}\}$, confidence level $\delta$, radius $\{\beta_k\}_{k\le K}$, ridge regression coefficient $\lambda$.
\STATE Initialize online-only design matrix and offline--online design matrix:
$
M^{\on}_1 \leftarrow \lambda I, w^{\on}_1 \leftarrow 0\in\R^d, M^{\all}_1 \leftarrow \lambda I, w^{\all}_1 \leftarrow 0\in\R^d.
$ 
\FOR{$m=1 : M_{\off}$} 
    \FOR{$h=1 : H$}
        \STATE Form the regression pair 
        $
        x^{\off}_{m,h} \leftarrow \phi(\cdot \mid  s^{\off}_{m,h},a^{\off}_{m,h})\TildeV_{m,h+1}\in\R^d,
        y^{\off}_{m,h} \leftarrow \TildeV_{m,h+1}\bigl(s^{\off}_{m,h+1}\bigr)
        $. 
        \STATE Update statistics
        $
        M^{\all}_1 \leftarrow M^{\all}_1 + x^{\off}_{m,h}(x^{\off}_{m,h})^\top,
        w^{\all}_1 \leftarrow w^{\all}_1 + x^{\off}_{m,h} y^{\off}_{m,h}.
        $
    \ENDFOR
\ENDFOR
\FOR{$k=1:K$}
    \STATE Compute estimators 
    $
    \thetaall_k \leftarrow (M^{\all}_k)^{-1} w^{\all}_k.
    $
    \STATE Define confidence set
    $
    \B_k' \leftarrow \Bigl\{\theta\in\R^d:\|\theta-\thetaall_k\|_{M^{\all}_k}\le \beta_k\Bigr\}.
    $
    \STATE Set $\Qhat_{k, H+1}(\cdot,\cdot)\leftarrow 0$, and for $h=H,\dots,1$ define \hfill 
    \[
    \Vhat_{k,h}(s)\leftarrow \max_{a\in\A} \Qhat_{k,h}(s,a),\qquad
    \Qhat_{k,h}(s,a)\leftarrow r(s,a)+\max_{\theta\in\B_k'}\sum_{j=1}^d \theta_j \phi_j(\cdot\mid s,a)^\top \Vhat_{k,h+1}.
    \]
    \STATE Set $s^{\on}_{k,1} =  s_1$.
    \FOR{$h=1:H$}
        \STATE Choose $a^{\on}_{k,h}\in\arg\max_{a\in\A} \Qhat_{k,h}(s^{\on}_{k,h},a)$ and observe $s^{\on}_{k,h+1}\sim P(\cdot\mid s^{\on}_{k,h},a^{\on}_{k,h})$.
        \STATE Form the online regression pair
        $
        x^{\on}_{k,h} \leftarrow \phi(\cdot \mid s^{\on}_{k,h},a^{\on}_{k,h})\Vhat_{k,h+1},
        y^{\on}_{k,h} \leftarrow \Vhat_{k,h+1}\bigl(s^{\on}_{k,h+1}\bigr).
        $
        \STATE Update offline--online statistics
        $
        M^{\all}_{k} \leftarrow M^{\all}_k + x^{\on}_{k,h}(x^{\on}_{k,h})^\top,
        w^{\all}_{k} \leftarrow w^{\all}_k + x^{\on}_{k,h} y^{\on}_{k,h}.
        $
        \ENDFOR
        \STATE $
        M^{\all}_{k+1} \leftarrow M^{\all}_{k},
        w^{\all}_{k+1} \leftarrow w^{\all}_{k}.
        $
\ENDFOR
\end{algorithmic}
\end{algorithm}

\subsection{MDP Setting and Behavior Policy}

We consider a synthetic episodic MDP with $|\mathcal S|=5$, $|\mathcal A|=10,H=3$, and $B = 2$. We follow the setting $\lambda=H^2 d$ as in the main text. Under the standard embedding of tabular MDPs into linear mixture MDPs (\Cref{remark_tabularMDP_equivalence}), the feature dimension is $d=|\mathcal S|^2|\mathcal A|=250$. For each $(s,a)$, the reward $r(s,a)$ is sampled independently from the uniform distribution $\mathrm{Unif}(0,1)$.

\paragraph{Online and Offline Transition Kernel.} In the online transition kernel $P$, each $P(\cdot\mid s,a)$ is sampled from a Dirichlet distribution with parameter $(1,1,\ldots,1)$ over the probability simplex, yielding uniform density over transitions. The offline transition kernel $P^{\off}$ is constructed to maximize the discrepancy between the offline and online environments subject to the budget $\Delta$. Given a fixed online transition kernel $P$ and reward function $r$, we first compute, for each state $s \in \S$, the value $V(s)$ induced by the worst policy under the online transition kernel $P$. For each state--action pair $(s,a)$, we then solve the following constrained optimization problem:
\begin{align}
    P^{\off}(\cdot \mid s,a) = \arg \min_{P^{\off}(\cdot \mid s,a)} \Ex_{s' \sim P^{\off}(\cdot \mid s,a)}[V(s')]
    \quad
    \text{s.t.}
    \quad
    \|P^{\off}(s,a) - P(s,a)\|_1 \le \Delta.
    \label{eq_poffdesign}
\end{align}
Intuitively, states with smaller values of $V(s)$ are regarded as worse under the online kernel $P$. Under constraint $\Delta$, the construction~\cref{eq_poffdesign} shifts probability mass of $P^{\off}(\cdot \mid s,a)$ toward such states for every $(s,a)$. Note that, although this procedure yields a substantially shifted offline environment, it does not necessarily correspond to the most adversarial design of $P^\off$. 

\paragraph{Behavior Policy and Control of $\tau$.} In all experiments, the offline behavior policy is the random policy. In~\cref{fig2a},~\ref{fig2b}, and~\ref{fig2d}, where $\tau=\Theta(1)$,
we adopt the designs of $\{\TildeV_{m,h}\}$ described in~\Cref{sec_simulations} for \ALG(\textsf{Optimal}) and \ALG(\textsf{Pessimistic}).
In~\cref{fig2c} and~\cref{fig_3}, where we vary $\tau$, the behavior policy is fixed and
$\tau$ is controlled through the design of $\{\TildeV_{m,h}\}$. Specifically, let $\S=\{s_1,s_2,s_3,s_4,s_5\}$.
For any target value of $\tau$, whenever the state at $(m,h)$ is $s_1$ or $s_2$, we set
\(
\Pr(\TildeV_{m,h}=0)=1-\tau,
\)
and with probability $\tau$, we apply the designs of $\{\TildeV_{m,h}\}$ described in~\Cref{sec_simulations}. 

\paragraph{Additional Simulation Results Including \DPLSVI and \UCBVI.}
We conclude this section by presenting additional simulation results that include comparisons with \DPLSVI and \UCBVI.

\begin{figure}[t!]
    \centering
    \begin{subfigure}[t]{0.48\textwidth}
        \centering
        \includegraphics[width=\linewidth]{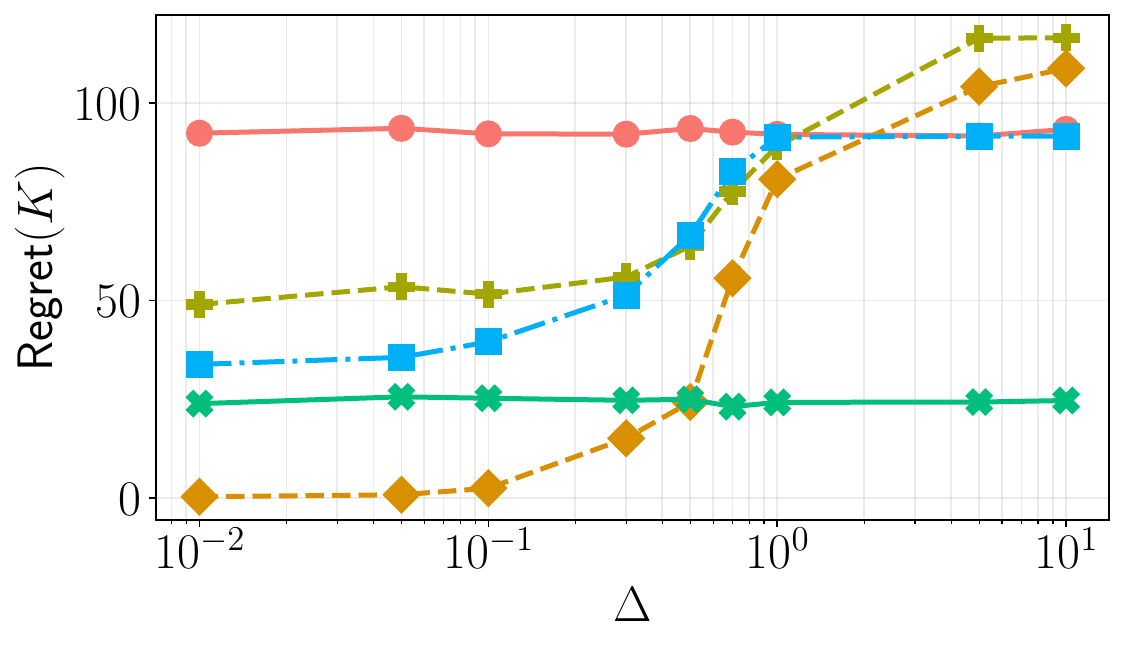}
        \caption{$\Regret(K)$ versus $\Delta$.}
        \label{fig4a}
    \end{subfigure}
    \hfill
    \begin{subfigure}[t]{0.48\textwidth}
        \centering
        \includegraphics[width=\linewidth]{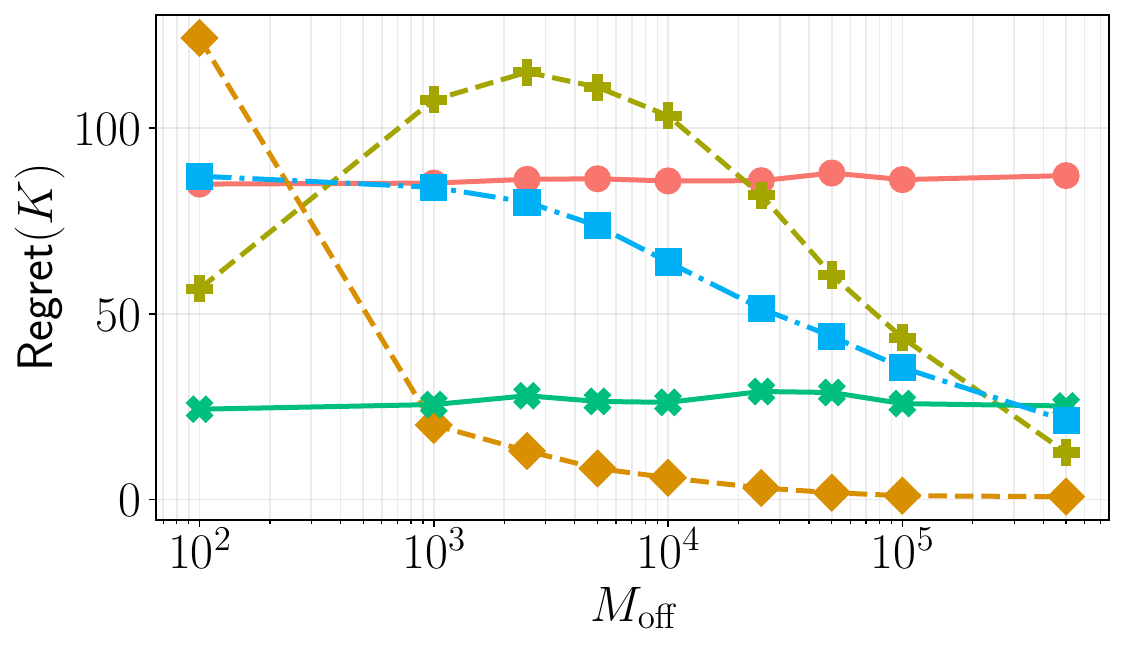}
        \caption{$\Regret(K)$ versus $M^\off$.}
        \label{fig4b}
    \end{subfigure}
    \hfill
    \vspace{0.4em}
    \begin{subfigure}[t]{0.65\textwidth}
        \centering
        \includegraphics[width=\linewidth]{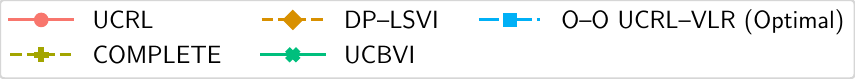}
    \end{subfigure}
    \caption{
        Supplementary results to~\Cref{fig_main}, including \DPLSVI and \UCBVI. The setting in~\Cref{fig4a} matches that of~\Cref{fig2a}, and the setting in~\Cref{fig4b} matches that of~\Cref{fig2b}.
    }
    \label{fig_appendix}
\end{figure}

As shown in \cref{fig_appendix}, \UCRL and \UCBVI remain insensitive to $\Delta$ since it does not use any offline data. \DPLSVI and \UCBVI outperform \UCRL, \ALG, and \COMPLETE across many regimes. This is expected, as \DPLSVI and \UCBVI are designed for tabular settings, whereas \ALG is developed for the more general linear mixture MDP framework.

However, in certain regimes of $M^\off$ and $\Delta$ (e.g., large $\Delta$ in~\Cref{fig4a} and large $M^\off$ in~\Cref{fig4b}), \ALG can outperform \UCBVI and \DPLSVI, as \DPLSVI lacks a safeguard mechanism and \UCBVI does not utilize the offline data.  This suggests that a well-designed integration of offline data can compensate for, and in some regimes even overcome, the dimensional disadvantage inherent in the linear mixture formulation, thereby validating the effectiveness of \ALG.

\end{APPENDICES}







\end{document}